\documentclass[a4paper]{llncs}

\usepackage{alltt}
\usepackage{amssymb}
\usepackage{caption}
\usepackage{paralist}
\usepackage{nicefrac}
\usepackage{booktabs}
\usepackage{xspace}
\usepackage{amsmath}
\usepackage{marvosym}
\usepackage{wasysym}
\usepackage{verbatim}
\usepackage{float}
\usepackage{hyperref}
\usepackage{multirow}
\usepackage{cite}
\usepackage[capitalize]{cleveref}
\usepackage[per-mode=symbol,detect-all]{siunitx}
\usepackage{graphics}
\usepackage{amssymb}
\usepackage{enumitem}
\usepackage[table]{xcolor}
\usepackage{booktabs}
\usepackage{colortbl}
\usepackage{ifthen}
\usepackage{mathdots}
\usepackage{arydshln}
\usepackage[outline]{contour}
\usepackage{graphicx}
\usepackage{wrapfig}
\usepackage{algorithm}
\usepackage{algpseudocode}
\usepackage{algorithmicx}
\usepackage{subcaption}
\usepackage{tabularx, stackengine,collcell}

\definecolor{navy}{RGB}{0,0,128}
\definecolor{forestGreen}{RGB}{0,110,51}

\usepackage{tikz,ifthen,pgfplots}
\usetikzlibrary{arrows,trees,backgrounds,automata,shapes,decorations,plotmarks,fit,calc,positioning,shadows,chains}
\usetikzlibrary{quotes,arrows.meta}
\tikzstyle{every pin edge}=[<-,shorten <=1pt]
\tikzstyle{neuron}=[circle,fill=black!25,minimum size=17pt,inner sep=0pt]
\tikzstyle{input neuron}=[neuron, fill=green!50]
\tikzstyle{output neuron}=[neuron, fill=red!50]
\tikzstyle{hidden neuron}=[neuron, fill=blue!50]
\tikzstyle{small neuron}        =[hidden neuron, draw, minimum size=15pt]
\tikzstyle{small input neuron}  =[input neuron , draw, minimum size=15pt]
\tikzstyle{small output neuron} =[output neuron, draw, minimum size=15pt]
\tikzstyle{annot} = [text width=4em, text centered]
\tikzstyle{nnedge} = [-{stealth},shorten >=0.1cm, shorten <=0.05cm,line width=0.8pt,black]
\tikzstyle{edge} = [->,line width = 0.3pt, shorten >=0.2cm]
\tikzstyle{edgeWide} = [->,line width = 2pt, , shorten >=0.2cm]
\usetikzlibrary{calc}

\tikzset{every picture/.style={line width=0.75pt}} 
\tikzstyle{BadSquare}=[rectangle,fill=red!30!white,minimum size=25pt,inner 
sep=0pt]
\tikzstyle{InitSquare}=[rectangle,fill=green!30!white,minimum size=25pt,inner 
sep=0pt]

\newcommand{\mysubsection}[1]{\medskip\noindent\textbf{#1}}

\newcommand{\relu}{\text{ReLU}\xspace}
\newcommand{\pdt}{\text{PDT}\xspace}

\newcommand{\sat}{\texttt{SAT}\xspace}
\newcommand{\unsat}{\texttt{UNSAT}\xspace}
\newcommand{\init}{\texttt{INIT}\xspace}
\newcommand{\project}{\texttt{PROJECT}\xspace}

\newcommand{\conditionMax}{\texttt{MAX}\xspace}
\newcommand{\conditionPercentile}{\texttt{PERCENTILE}\xspace}
\newcommand{\conditionCombined}{\texttt{COMBINED}\xspace}

\newcommand{\smtsolver}{\texttt{SMT SOLVER }}
\newcommand{\query}{\texttt{query }}
\newcommand{\maxAgree}{\texttt{low}}
\newcommand{\minDisagree}{\texttt{high}}
\newcommand{\disagreementUB}{\texttt{M}}

\newcommand{\marabou}{\textit{Marabou}\xspace}

\newcommand{\modelsSet}{\mathcal{N}}
\newcommand{\modelsSubset}{\mathcal{N'}}
\newcommand{\actionSpace}{\mathcal{A}}
\newcommand{\statesSpace}{\mathcal{S}}
\newcommand{\outputSpace}{\mathcal{O}}
\newcommand{\distanceFn}{d}

\newcommand{\feasibleStatesSpace}{\Psi}
\newcommand{\maxAgg}{\texttt{MAX}\xspace}

\newcommand{\percentileAgg}{\texttt{PERCENTILE}\xspace}
\DeclareMathOperator*{\argmax}{arg\,max}

\usepackage{bussproofs}
\usepackage{gensymb}

\newif\ifcomments
\commentstrue

\newif\ifoutline
\outlinefalse

\newif\iflong
\longtrue

\renewcommand{\paragraph}[1]{\vspace{1mm}\noindent{\bf #1}\ }

\newcounter{experimentCounter}
\newcommand{\experiment}[1]{\noindent%
    \refstepcounter{experimentCounter}\textbf{Experiment (\theexperimentCounter): #1}
   }

\usepackage{array}
\newcolumntype{P}[1]{>{\centering\arraybackslash}p{#1}}

 \usepackage{placeins}

\begin{document}
	
	\title{Verifying Generalization in Deep Learning}
	
	\author{
		Guy Amir$^{*}$,
		Osher Maayan$^{*}$,
		Tom Zelazny,
		Guy Katz and
		Michael Schapira
	}

	\institute{
		The Hebrew University of Jerusalem, Jerusalem, Israel \\
		 \email{ \{guyam, osherm, tomz, guykatz, schapiram\}@cs.huji.ac.il} }

\maketitle

\let\svthefootnote\thefootnote
\let\thefootnote\relax\footnotetext{[*] Both authors contributed equally.}
\let\thefootnote\svthefootnote
\addtocounter{footnote}{0}

\begin{abstract} 

Deep neural networks (DNNs) are the workhorses of deep learning, which
constitutes the state of the art in numerous application domains. However,
DNN-based decision rules are notoriously prone to poor
\emph{generalization}, i.e., may prove inadequate 
on inputs not encountered during training. This limitation poses a significant obstacle to employing deep learning for mission-critical tasks, and also in real-world environments that exhibit high variability. We propose a novel, verification-driven methodology for identifying
DNN-based decision rules that generalize well to new input domains. Our approach quantifies
generalization to an input domain by the extent to which decisions reached by
\textit{independently trained} DNNs are in agreement for inputs in this domain. We show how, by harnessing the power of DNN verification, our approach can be efficiently and effectively realized.
We evaluate our verification-based approach on three deep
reinforcement learning (DRL) benchmarks, including a system for
Internet congestion control. Our results establish the usefulness of our 
approach.  More broadly, our work puts
forth a novel objective for formal verification, with the potential for
mitigating the risks associated with deploying DNN-based systems in the wild. 

		
\end{abstract}

\section{Introduction}
\label{sec:introduction}

Over the past decade, deep learning~\cite{GoBeCo16} has achieved
state-of-the-art results in natural
language processing, image recognition, game playing, computational biology,
and many additional fields~\cite{SiZi14, SiHuMaGuSiVaScAnPaLaDi16, Al19,
  CoWeBoKaKaKu11, KrSuHi12, BoDeDwFiFlGoJaMoMuZhZhZhZi16,
  JuLoBrOwKo16}. However, despite its impressive success, deep learning still
suffers from severe drawbacks that limit its applicability in domains
that involve mission-critical tasks or highly variable inputs.

One such crucial limitation is the notorious difficulty of deep neural
networks (DNNs) to \emph{generalize} to new input domains, i.e., their
tendency to perform poorly on inputs that significantly differ from those
encountered while training. During training, a DNN is presented with
input data sampled from a specific distribution over some input domain (``\textit{in-distribution}''
inputs). The induced DNN-based rules may fail in generalizing to inputs not encountered during training due to (1) the DNN being invoked
``out-of-distribution'' (OOD), i.e., when there is a mismatch between the
distribution over inputs in the training data and in the DNN's
operational data; (2) some inputs not being sufficiently represented in the finite training data (e.g., various
low-probability corner cases); and (3) ``overfitting'' the decision rule to the training data.

A notable example of the importance of establishing the
generalizability of DNN-based decisions lies in recently proposed
applications of deep reinforcement learning (DRL)~\cite{Li17} to real-world
systems. Under DRL, an \textit{agent}, realized as a DNN, is trained by repeatedly interacting with its environment to learn a decision-making \textit{policy} that attains high performance with respect to a certain objective (``\emph{reward}''). DRL has recently been applied to many real-world challenges~\cite{MnKaSi13,ZhLiZhLiXiChXiWaChLi19, MaJaWo20, LiZhChMe18,
JaRoGoScTa19, VaScShTa17, ChXuWu17, MaAlMeKa16, LeMoSuSa20, MaNeAl17}. In many application domains, the learned policy is expected to perform well
across a daunting breadth of operational environments, whose diversity
cannot possibly be captured in the training data. Further, the cost of
erroneous decisions can be dire.  Our discussion
of DRL-based Internet congestion control (see Sec.~\ref{subsec:aurora}) illustrates this point.

 Here, we present a methodology for identifying DNN-based decision rules that 
 generalize well to \emph{all possible distributions} over an input
domain of interest. Our approach hinges on the following key observation. DNN
training in general, and DRL policy training in particular,
incorporate multiple stochastic aspects, such as the initialization of
the DNN's weights and the order in which inputs are
observed during training. Consequently, even when DNNs with \emph{the same}
architecture are trained to perform an \emph{identical} task on \emph{the same}
data, somewhat different decision rules will typically be
learned. Paraphrasing Tolstoy's Anna Karenina~\cite{To77}, we argue
that ``successful decision rules are all alike; but every unsuccessful
decision rule is unsuccessful in its own way''. Differently put, when
examining the decisions by several \textit{independently trained} DNNs
on a certain input, these are likely to agree only when their
(similar) decisions yield high performance.

In light of the above, we propose the following heuristic for
generating DNN-based decision rules that generalize well to \textit{an entire} 
given
domain of inputs: independently train multiple DNNs, and then seek a
subset of these DNNs that are in strong agreement across \emph{all}
possible inputs in the considered input domain (implying, by our hypothesis, that these DNNs' learned decision rules
generalize well to all probability distributions over this domain). Our 
evaluation demonstrates (see Sec.~\ref{sec:Evaluation})
that this methodology is extremely powerful and enables distilling
from a collection of decision rules the few that indeed generalize
better to inputs within this domain. Since our heuristic seeks DNNs whose 
decisions are in
agreement for \emph{each and every} input in a specific domain, the decision 
rules reached this way achieve robustly
high generalization across different possible distributions over inputs in this 
domain.

Since our methodology involves contrasting the outputs of different
DNNs over possibly \textit{infinite} input domains, using formal verification is natural. To this end, we build on recent advances in formal verification of DNNs~\cite{LuScHe21, AlAvHeLu20,
  AvBlChHeKoPr19, BaShShMeSa19, PrAf20, AnPaDiCh19, SiGePuVe19,
  XiTrJo18, Eh17}.
DNN verification literature has focused
 on establishing the local adversarial robustness of DNNs, i.e., seeking small input perturbations
that result in misclassification by the DNN~\cite{GeMiDrTsChVe18,
  LyKoKoWoLiDa20, GoKaPaBa18}. Our approach broadens the applicability of
DNN verification by demonstrating, for the first time (to the best of our knowledge), how
it can also be used to identify DNN-based decision rules that
generalize well. More specifically, we show how, for a given input
domain, a DNN verifier can be utilized to assign a score to a DNN reflecting its level of agreement with other DNNs across the
entire input domain. This enables iteratively pruning the set of candidate DNNs, eventually
keeping only those in strong agreement, which tend to generalize well.





To evaluate our methodology, we focus on three popular DRL benchmarks:
\begin{inparaenum}[(i)]
\item \emph{Cartpole}, which involves controlling a cart while
  balancing a pendulum;
\item \emph{Mountain Car}, which involves controlling a car that needs to escape a valley; and
\item \emph{Aurora}, an Internet congestion controller.
\end{inparaenum}

Aurora is a particularly compelling example for our approach. While Aurora is 
intended to tame network congestion across a vast
diversity of real-world Internet environments, Aurora is trained only
on synthetically generated data. Thus, to deploy Aurora in the real world, it 
is critical to ensure that its policy is 
sound for numerous scenarios not captured by its training inputs.

Our evaluation results show that, in all three settings, our
verification-driven approach is successful at ranking DNN-based DRL
policies according to their ability to generalize well to
out-of-distribution inputs. Our experiments also demonstrate that
formal verification is superior to gradient-based methods and predictive 
uncertainty methods. These results 
showcase the potential of our approach. 
Our code and benchmarks are publicly available as an artifact accompanying this 
work~\cite{ArtifactRepository}. 

The rest of the paper is organized as follows.
Sec.~\ref{sec:Background} contains background on DNNs, DRLs, and
DNN verification. In Sec.~\ref{sec:Approach} we present
our verification-based methodology for identifying DNNs that
successfully generalize to OOD inputs. We present our evaluation
in Sec.~\ref{sec:Evaluation}. Related work is covered in
Sec.~\ref{sec:RelatedWork}, and we
\begin{wrapfigure}[7]{r}{0.45\textwidth}
  \vspace{-0.8cm}
  \begin{center}
    \scalebox{0.75} {
      \def\layersep{2.0cm}
      \begin{tikzpicture}[shorten >=1pt,->,draw=black!50, node 
        distance=\layersep,font=\footnotesize]
        
        \node[input neuron] (I-1) at (0,-1) {$v^1_1$};
        \node[input neuron] (I-2) at (0,-2.5) {$v^2_1$};
        
        \node[hidden neuron] (H-1) at (\layersep,-1) {$v^1_2$};
        \node[hidden neuron] (H-2) at (\layersep,-2.5) {$v^2_2$};
        
        \node[hidden neuron] (H-3) at (2*\layersep,-1) {$v^1_3$};
        \node[hidden neuron] (H-4) at (2*\layersep,-2.5) {$v^2_3$};
        
        \node[output neuron] at (3*\layersep, -1.75) (O-1) 
        {$v^1_4$};
        
        \draw[nnedge] (I-1) --node[above] {$1$} (H-1);
        \draw[nnedge] (I-1) --node[above, pos=0.3] {$\ -3$} (H-2);
        \draw[nnedge] (I-2) --node[below, pos=0.3] {$4$} (H-1);
        \draw[nnedge] (I-2) --node[below] {$2$} (H-2);
        
        \draw[nnedge] (H-1) --node[above] {$\relu$} (H-3);
        \draw[nnedge] (H-2) --node[below] {$\relu$} (H-4);https://www.overleaf.com/project/63a03b9d4341f61c76dcdcf0
        
        \draw[nnedge] (H-3) --node[above] {$2$} (O-1);
        \draw[nnedge] (H-4) --node[below] {$-1$} (O-1);

        \node[below=0.05cm of H-1] (b1) {$+1$};
        \node[below=0.05cm of H-2] (b2) {$-2$};
        
        \node[annot,above of=H-1, node distance=0.8cm] (hl1) 
        {Weighted 
          sum};
        \node[annot,above of=H-3, node distance=0.8cm] (hl2) {ReLU 
        };
        \node[annot,left of=hl1] {Input };
        \node[annot,right of=hl2] {Output };
      \end{tikzpicture}
    }	
 \end{center}
	\vspace{-4mm}
	\caption{A toy DNN.}
	\label{fig:toyDnn}
      \end{wrapfigure}
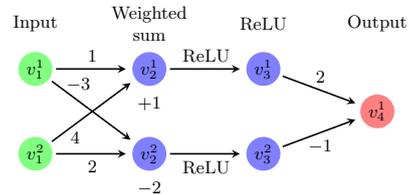
      conclude in Sec.~\ref{sec:Conclusion}.

\section{Background}
\label{sec:Background}
\noindent{\bf Deep neural networks (DNNs)}~\cite{GoBeCo16} are directed graphs that comprise several layers. Upon receiving an assignment of values to the nodes of its
first (input) layer, the DNN propagates these values, layer by layer, until ultimately
reaching the assignment of the final (output) layer.  Computing the
value for each node is performed according to the type of that node's layer. For example, in weighted-sum layers, the node's value is an
affine combination of the values of the nodes in the preceding layer to which it is connected. In
\emph{rectified linear unit} (\emph{ReLU}) layers, each node $y$
computes the value $y=\relu{}(x)=\max(x,0)$, where $x$ is a single
node from the preceding layer. For additional details on DNNs and
their training see~\cite{GoBeCo16}.
Fig.~\ref{fig:toyDnn} depicts a toy DNN. For
 input $V_1=[1, 2]^T$, the second layer computes the (weighted sum)
$V_2=[10,-1]^T$. The \relu{} functions are subsequently applied
in the third layer, and the result is $V_3=[10,0]^T$. Finally, the
network's single output is $V_4=[20]$.

\medskip
\noindent{\bf Deep reinforcement learning (DRL)}~\cite{Li17} is a machine learning paradigm, in which
a DRL agent, implemented as a DNN, interacts with an
\textit{environment} across discrete time-steps $t\in{0,1,2...}$. At each time-step, the
agent is presented with the environment's \textit{state}
$s_{t} \in \statesSpace$, and selects an \textit{action}
$N(s_t)=a_{t} \in \actionSpace$. The environment then transitions to its next
state $s_{t+1}$, and presents the agent with the \textit{reward} $r_t$ for its previous action. The agent is trained through repeated interactions with its environment to maximize the \textit{expected cumulative
  discounted reward}
$R_t=\mathbb{E}\big[\sum_{t}\gamma^{t}\cdot r_t\big]$ (where
$\gamma \in \big[0,1\big]$ is termed the \textit{discount factor})~\cite{SuBa18, ZhJoBr20, ShWoDh17, vHaGuSi16, SuMcSi99,
  HaZhAbLe18}.

\mysubsection{DNN and DRL Verification.}  A sound DNN
verifier~\cite{KaBaDiJuKo17} receives as input 
\begin{inparaenum}[(i)]
\item a \textit{trained} DNN $N$;
\item a precondition $P$ on the DNN's inputs, limiting the possible assignments to a domain of interest; and	
\item a postcondition $Q$ on the DNN's outputs, limiting the
  possible outputs of the DNN.
\end{inparaenum}
  The verifier can reply in one of two ways:
\begin{inparaenum}[(i)]
\item \sat, with a concrete input $x'$ for which
  $P(x') \wedge Q(N(x'))$ is satisfied; or
\item \unsat, indicating there does not exist such an $x'$.
\end{inparaenum}
Typically, $Q$ encodes the \textit{negation} of $N$'s desirable
behavior for inputs that satisfy $P$. Thus, a \sat result indicates
that the DNN errs, and that $x'$  triggers a bug;
whereas an \unsat result indicates that the DNN performs as
intended. An example of this process appears in
subsection~\ref{sec:appendix:VerificationQueries:toyExample} of the
Appendix.  To date, a plethora of verification approaches have been
proposed for general, feed-forward DNNs~\cite{Al21,KaBaDiJuKo17,
  GeMiDrTsChVe18, WaPeWhYaJa18, LyKoKoWoLiDa20, HuKwWaWu17}, as well
as DRL-based agents that operate within reactive
environments~\cite{CoMaFa21, BaGiPa21, ElKaKaSc21, AmScKa21,
  AmCoYeMaHaFaKa23}.

\section{Quantifying Generalizability via Verification}
\label{sec:Approach}

Our approach for assessing how well a DNN is expected to generalize on out-of-distribution
inputs relies on the ``Karenina hypothesis'': while there are many
(possibly infinite) ways to produce \emph{incorrect
  results}, correct outputs are likely to be fairly similar. Hence, to identify DNN-based decision rules that generalize well to new input domains, we advocate training multiple DNNs and scoring the learned decision models according to how well their outputs are
aligned with those of the other models for the considered input domain. These
scores can be computed using a backend DNN verifier. We show how, by iteratively filtering out models that tend to disagree with the rest, DNNs that generalize well can be effectively distilled.

We begin by introducing the following definitions for reasoning about the extent to which two DNN-based decision rules are in agreement over an input domain.

\begin{definition}[\textbf{Distance Function}]
 Let $\outputSpace$ be the space of possible outputs for a DNN. A \emph{distance function} for
   $\outputSpace$ is a function $\distanceFn: \outputSpace \times \outputSpace \mapsto \mathbb{R^+}$.
 \end{definition}

Intuitively, a distance function (e.g., the $L_{1}$ norm) allows us to quantify 
the level
 of (dis)agreement between the decisions of two DNNs on the same input.
We elaborate on some choices of distance functions that may be appropriate
 in various domains in
 Appendix~\ref{sec:appendix:VerificationQueries}.





\begin{definition}[\textbf{Pairwise Disagreement Threshold}]\label{def:disagreementThreshold}
  Let $N_{1}, N_{2}$ be DNNs with the same output space $\outputSpace$, let $\distanceFn$ be a distance function, and let
  $\feasibleStatesSpace$ be an input domain. We
  define the \emph{pairwise disagreement threshold} (PDT) of $N_1$ and
  $N_2$ as:
  \[
     \alpha =
     \pdt{}_{\distanceFn, \feasibleStatesSpace}(N_1, N_2) \triangleq \min \left\{\alpha' \in \mathbb{R}^{+} \mid \forall x \in \feasibleStatesSpace \colon d(N_1(x),N_2(x)) \leq \alpha' \right\}
    \]
\end{definition}

The definition captures the notion that for \textit{any} input in
$\feasibleStatesSpace$, $N_{1}$ and $N_{2}$ produce
outputs that are at most $\alpha$-distance apart. A small $\alpha$ value indicates
that the outputs of $N_1$ and $N_2$ are close for all inputs in
$\feasibleStatesSpace$, whereas a high value indicates that there
exists an input in $\feasibleStatesSpace$ for which the decision models diverge
significantly.  


To compute $\pdt$ values, our approach employs verification to conduct a binary search for the maximum distance between
the outputs of two DNNs; see 
Alg.~\ref{alg:algorithmPairDisagreementScores}. 



\begin{algorithm}[ht]\caption{Pairwise Disagreement Threshold}
    \textbf{Input:} 
    DNNs ($N_{i}$, $N_{j}$), distance func. $\distanceFn$, input domain $\feasibleStatesSpace$, max. disagreement $\disagreementUB > 0$ \\
    %
    \textbf{Output:} $\pdt(N_{i}, N_{j})$

\begin{algorithmic}[1]  
        \State $\maxAgree \gets 0$, $\minDisagree \gets \disagreementUB$
         
        \While {$\left( \maxAgree < \minDisagree  \right)$} 

        \State $\alpha \gets \frac{1}{2} \cdot (\maxAgree+\minDisagree)$
        \State \query $\gets$ \smtsolver $\langle P \gets 
        \feasibleStatesSpace, 
        [N_i;N_j], Q \gets d(N_i,N_j)\geq\alpha \rangle$
        \label{line:SMTsolverForPdt}
        \State \textbf{if} \query is \sat \textbf{then}:  $\maxAgree \gets \alpha$
        
        \State \textbf{else if} \query is \unsat \textbf{then}:   $\minDisagree \gets \alpha$ 


        \EndWhile 
        
   

    \State \Return $\alpha$
    \end{algorithmic}
    \label{alg:algorithmPairDisagreementScores}
\end{algorithm}

Pairwise disagreement thresholds can be aggregated to
measure the disagreement between a decision model and a \textit{set} of other decision models, as defined next.

\begin{definition}[\textbf{Disagreement Score}]
  \label{def:disagreementScores}
  Let 
  $\modelsSet=\{N_{1}, N_{2},\ldots,N_{k}\}$
be a set of $k$ DNN-induced decision models, 
  let $\distanceFn$ be a distance function, and let
  $\feasibleStatesSpace$ be an input domain.
   A model's \emph{disagreement score} (DS) with respect to $\modelsSet$
   is defined as:
   \[
    DS_{\modelsSet,\distanceFn, \feasibleStatesSpace}(N_i) = \frac{1}{|\modelsSet|-1}\sum_{j \in [k], j\neq i}\pdt_{\distanceFn, \feasibleStatesSpace}(N_{i}, N_{j})
    \]
\end{definition}
Intuitively, the disagreement score measures how much a single
  decision model tends to disagree with the remaining models, on average. 


Using disagreement scores, our heuristic employs an iterative scheme for selecting a subset of
models that generalize to OOD scenarios --- as encoded by inputs in $\feasibleStatesSpace$ (see Alg.~\ref{alg:modelSelection}).
First, a set of $k$
DNNs $\{N_1, N_2,\ldots,N_k\}$ are \emph{independently} trained on the training data. Next, a backend verifier is invoked to calculate, 
for each of the ${k \choose 2} $ DNN-based model pairs, their
respective pairwise-disagreement threshold (up to some
$\epsilon$ accuracy).
Next, our algorithm iteratively: \begin{inparaenum}[(i)]
    \item calculates the disagreement score for each model in the remaining subset of models;
    \item identifies the models with the (relative) highest $DS$
      scores; and 
    \item removes them (Line~\ref{lst:line:removeModels} in Alg.~\ref{alg:modelSelection}).
\end{inparaenum}
The algorithm terminates after exceeding a user-defined number of
iterations (Line~\ref{lst:line:mainLoopStart} in Alg.~\ref{alg:modelSelection}), or when the remaining
models ``agree'' across the input domain, as indicated by nearly identical disagreement scores
(Line~\ref{lst:line:checkDifferenceBetweenModels} in Alg.~\ref{alg:modelSelection}). We note that the algorithm is also given an upper bound (\disagreementUB) on the maximum difference, informed by the user's domain-specific knowledge.




\begin{algorithm}[ht]\caption{Model Selection}\label{alg:modelSelection}
 \textbf{Input:} Set of models $\modelsSet=\{N_{1},\ldots,N_{k}\}$,
 max disagreement $\disagreementUB$,  number of \texttt{ITERATIONS}  \\
 \textbf{Output:} $\modelsSubset \subseteq \modelsSet$
  \begin{algorithmic}[1]
  \State \texttt{PDT} $\gets $ \Call{Pairwise Disagreement Thresholds}{$\modelsSet, d, \feasibleStatesSpace, \disagreementUB$} 
  \Comment{table with all PDTs} 
\label{def:disagreementThreshold}
  \State $\modelsSubset \gets \modelsSet$
   \For {$l=1 \ldots $\texttt{ITERATIONS} } \label{lst:line:mainLoopStart} 
        \For{$N_{i} \in \modelsSubset$}
        
            \State \texttt{currentDS}[$N_{i}$] $\gets DS_{\modelsSubset}(N_{i}, \texttt{PDT})$ 
            \Comment{based on definition~\ref{def:disagreementScores}} 
     \label{line:getDS}
        \EndFor
        \State \textbf{if}  \texttt{modelScoresAreSimilar(\texttt{currentDS})} \textbf{then}: \texttt{break}
 \label{lst:line:checkDifferenceBetweenModels}
       \State $\texttt{modelsToRemove} \gets \texttt{findModelsWithHighestDS(\texttt{currentDS})} $
\label{lst:line:modelsToRemove}
        \State $\modelsSubset \gets \modelsSubset
  \setminus \texttt{modelsToRemove}$
        \Comment{remove models that tend to disagree} \label{lst:line:removeModels}
    \EndFor
    \State \Return $\modelsSubset$
\end{algorithmic}
\end{algorithm}

\medskip
\noindent{\textbf{DS Removal Threshold.}}
Different criteria are possible for determining the DS threshold above for 
which models are removed, and how many models to remove in each iteration 
(Line~\ref{lst:line:modelsToRemove} in Alg.~\ref{alg:modelSelection}). A 
natural and simple approach, used in our evaluation, is to remove the $p\%$ 
models with the \emph{highest} disagreement scores, for some choice of $p$ 
($25\%$ in our evaluation). Due to space constraints, a thorough discussion of 
additional filtering criteria (all of which proved successful) is relegated to
Appendix~\ref{sec:appendix:AlgorithmAdditionalInformation}.


\section{Evaluation}
\label{sec:Evaluation}

We extensively evaluated our method using three DRL benchmarks. As discussed in 
the introduction, verifying the generalizability of DRL-based systems is 
important since such systems are often expected to provide robustly high 
performance across a broad range of environments, whose diversity is not 
captured by the training data. Our evaluation spans two classic DRL settings, 
Cartpole~\cite{BaSuAn83} and Mountain Car~\cite{Mo90}, as well as the recently 
proposed Aurora congestion controller for Internet traffic~\cite{JaRoGoScTa19}. 
Aurora is a particularly compelling example for a fairly complex DRL-based 
system that addresses a crucial real-world challenge and must generalize to 
real-world conditions not represented in its training data. 

\medskip
\noindent{\textbf{Setup.}} For each of the three DRL benchmarks, we first trained multiple DNNs with the same architecture,
where the training process differed only in the random seed used. We then 
removed from this set of DNNs all but the ones that achieved high reward values 
in-distribution (to eliminate the possibility that a decision model generalizes 
poorly simply due to poor training). Next, we defined out-of-distribution input 
domains of interest for each specific benchmark, and used
Alg.~\ref{alg:modelSelection} to select the models most likely to
generalize well on those domains according to our framework. To establish the 
ground truth for how well different models actually generalize in practice, we 
then applied the models to OOD inputs drawn from the considered domain and 
ranked them based on their empirical performance (average reward). To 
investigate the robustness of our results, the last step was conducted for 
varying choices of probability distributions over the inputs in the domain. All 
DNNs used have a feed-forward architecture comprised of two hidden layers of 
\relu activations, and include $32$-$64$ neurons in the first hidden layer, and 
$16$ neurons in the second hidden layer.

The results indicate that models selected by our approach are likely to perform \emph{significantly better}
than the rest. Below we describe the gist of our evaluation;
extensive additional information is available in the appendices.

\subsection{Cartpole}\label{subsec:cartpole}
\begin{wrapfigure}{r}{0.5\textwidth}
  \vspace{-2.0cm}
  \centering
  \begin{center}
    \includegraphics[width=0.45\textwidth]{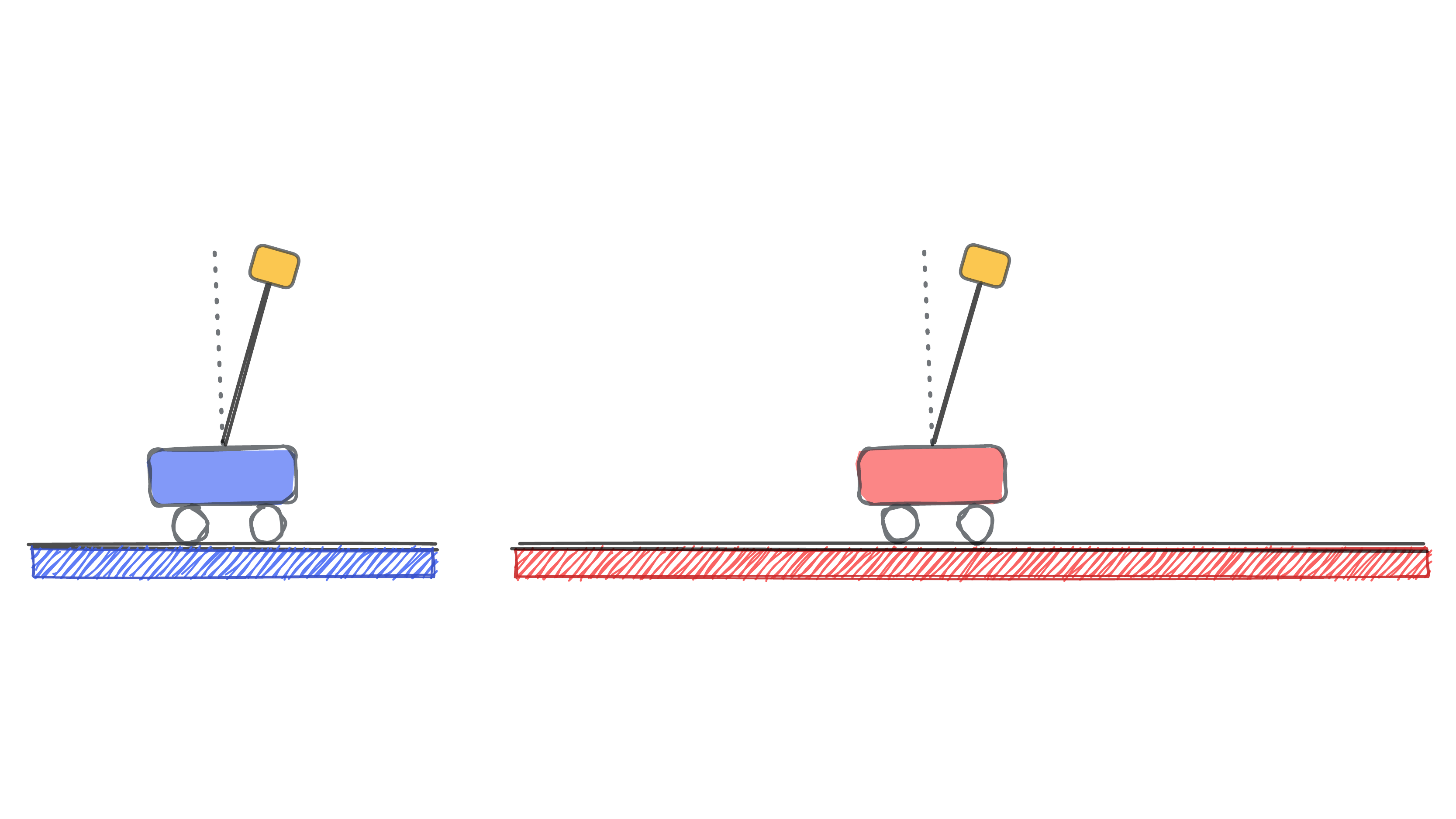}
  \end{center}
  \vspace{-1.0cm}
  \caption{Cartpole: in-distribution setting (blue) and OOD setting (red).}
  \vspace{-0.5cm}
  \label{fig:CartpoleInDistributionAndOod}
\end{wrapfigure}
Cartpole~\cite{GeSi93} is a well-known RL benchmark in which an agent
controls the movement of a cart with an upside-down pendulum
(``pole'') attached to its top. The cart moves on a platform and the
agent's goal is to keep the pole balanced for as long as possible
(see
Fig.~\ref{fig:CartpoleInDistributionAndOod}).

\medskip
\noindent{\textbf{Agent and Environment.}}
The agent's inputs are $s=(x, v_{x}, \theta, v_{\theta})$, where $x$
represents the cart's location on the platform, $\theta$ represents the
pole's angle (i.e., $|\theta| \approx 0$ for a balanced pole,
$|\theta| \approx 90\degree$ for an unbalanced pole), $v_{x}$
represents the cart's horizontal velocity and $v_{\theta}$ represents
the pole's angular velocity.

\medskip
\noindent{\textbf{In-Distribution Inputs.}} During
training, the agent is incentivized to
balance the pole, while  
staying within the platform's boundaries.
In each iteration, the agent's single output indicates the cart's acceleration (sign and magnitude) for the next step. 
During training, we defined the platform's bounds to be $[-2.4,
2.4]$, and the cart's initial position as near-static, 
and close to the center of the platform (left-hand side of
Fig.~\ref{fig:CartpoleInDistributionAndOod}). This was achieved by
drawing the cart's initial state vector values uniformly from the range
$[-0.05, 0.05]$.



\medskip
\noindent{\textbf{(OOD) Input Domain.}}
We consider an input domain with larger platforms than the ones used
in training. To wit, we now allow the $x$ coordinate of the input vectors to cover a wider
range of $[-10,10]$.
For the other inputs, we used the same bounds as during the training.
See Appendices~\ref{sec:appendix:trainingAndEvaluation}
and~\ref{sec:appendix:VerificationQueries} for additional details.

\begin{wrapfigure}{r}{0.6\textwidth}
            \centering
		\begin{center}
			\includegraphics[width=0.65\textwidth]{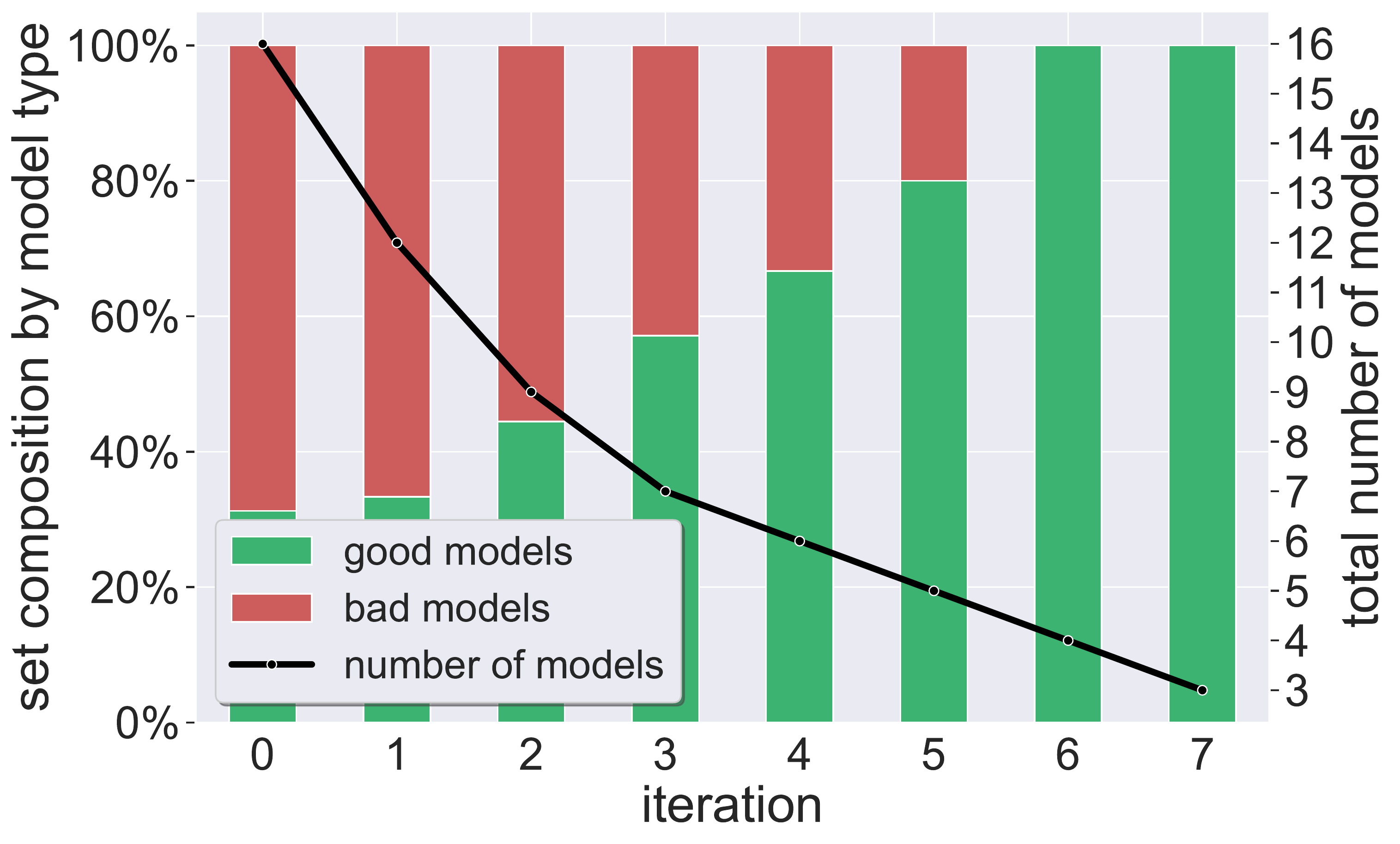}
		\end{center}
            \vspace{-0.5cm}
            \caption{Cartpole: Alg.~\ref{alg:modelSelection}'s results, per 
            iteration: the bars reflect the ratio between the good/bad models 
            (left y-axis) in the surviving set of models, and the curve 
            indicates the number of surviving models (right y-axis).}
        \vspace{-0.5cm}
		\label{fig:cartpolePercentileGoodBadResults}
\end{wrapfigure}
\medskip
\noindent{\textbf{Evaluation.}}
We trained $k=16$ models, all of which achieved high rewards
during training on the short platform. Next, 
we ran Alg.~\ref{alg:modelSelection} until convergence ($7$
iterations, in our experiments) on the
aforementioned input domain, resulting in a set of $3$ models. 
We then tested all $16$ original models using (OOD) inputs drawn from the new domain, such that the generated distribution encodes a novel setting: the cart is now placed at the center of a much longer, shifted platform (see the red cart in
Fig.~\ref{fig:CartpoleInDistributionAndOod}).

All other parameters in the OOD environment
were identical to those used for the original training. 
Fig.~\ref{fig:cartpoleRewards} (in the Appendix)
depicts the results of evaluating the models using $20,000$ OOD
instances.
Of the original $16$ models, $11$ scored a low-to-mediocre
average reward, indicating their poor ability to generalize to this
new distribution. Only $5$
models obtained high reward values, including the $3$ models identified
by Alg.~\ref{alg:modelSelection};
thus implying that our method was able to effectively remove all $11$ models that
would have otherwise performed poorly in this OOD setting
(see Fig.~\ref{fig:cartpolePercentileGoodBadResults}).  For additional
information, see
Appendix~\ref{sec:appendix:CartPoleSupplementaryResults}.

\subsection{Mountain Car}
\label{subsec:mountaincar} 
For our second experiment, we evaluated our method on the
Mountain Car~\cite{Ri05} benchmark, in which an agent controls a car
that needs to learn how to escape a valley and reach a target. As in the
Cartpole experiment, we selected a set of models that performed well in-distribution and applied our method to identify a subset of models that make similar decisions in a predefined input domain. We again generated OOD inputs (relative to the training) from within this domain, and observed that the
models selected by our algorithm indeed generalize significantly better than 
their peers that were iteratively removed. Detailed information about this
benchmark can be found in
Appendix~\ref{sec:appendix:MountainCarSupplementaryResults}.

\subsection{Aurora Congestion Controller}\label{subsec:aurora}

In our third benchmark, we applied our method to a complex, real-world
system that implements a policy for Internet congestion control. The goal of congestion control is to determine, for each traffic source in a communication
network, the pace at which data packets should be sent into the network. Congestion
control is a notoriously difficult and fundamental challenge in computer
networking~\cite{LoPaDo02, Na84}; sending packets too fast might cause network
congestion, leading to data loss and delays. Conversely, low sending
rates might under-utilize available network bandwidth.
\emph{Aurora}~\cite{JaRoGoScTa19} is a DRL-based congestion
controller that is the subject of recent work on DRL verification~\cite{ElKaKaSc21,
  AmScKa21}. In each time-step, an Aurora agent observes statistics
regarding the network and decides the packet sending rate for the
following time-step. For example, if the agent observes excellent network
conditions (e.g., no packet loss), we expect it to increase the
packet sending rate to better utilize the network.
We note that Aurora handles a much harder task than classical RL benchmarks 
(e.g., Cartpole and Mountain Car): congestion controllers must react 
gracefully to various possible events based on nuanced signals, as reflected by 
Aurora’s inputs. Here, unlike in the previous benchmarks, it
is not straightforward to characterize the optimal policy.

\medskip
\noindent{\textbf{Agent and Environment.}}
Aurora's inputs are $t$ vectors $v_{1}, \ldots,v_{t}$, representing
observations from the $t$ previous time-steps.  The agent's single
output value indicates the change in the packet sending rate over the
next time-step.  Each vector $v_{i}\in\mathbb{R}^3$ includes three
distinct values, representing statistics that reflect the network's condition (see details in Appendix~\ref{sec:appendix:AuroraSupplementaryResults}).  In line with
previous work~\cite{ElKaKaSc21,JaRoGoScTa19, AmScKa21}, we set $t=10$
time-steps, making Aurora's inputs of size $3t=30$.
The reward function is a linear combination of
the data sender's throughput, latency, and packet loss, as observed by
the agent (see~\cite{JaRoGoScTa19} for additional details).

\medskip
\noindent{\textbf{In-Distribution Inputs.}} Aurora's training applies the congestion controller to simple network scenarios where a \emph{single} sender sends traffic towards a \emph{single} receiver across a \emph{single} network link. Aurora is trained across varying choices of initial sending rate, link bandwidth, link packet-loss rate, link latency, and size of the link's packet buffer. During training, packets are initially sent by Aurora at a rate corresponding to $0.3-1.5$ times the link's bandwidth.

\medskip
\noindent{\textbf{(OOD) Input Domain.}}
In our experiments, the input domain encoded a link with a
\emph{shallow packet buffer}, implying that only a few packets can accumulate in the network (while most excess traffic is discarded), causing the link to exhibit a volatile behavior. This is captured by the initial sending rate being up to $8$ times the link's bandwidth, to model the possibility of a dramatic decrease in available bandwidth (e.g., due to competition, traffic shifts, etc.). See Appendix~\ref{sec:appendix:AuroraSupplementaryResults} for additional details.

\medskip
\noindent{\textbf{Evaluation.}} We ran our algorithm and scored the models based on
their disagreement upon this large domain, which includes inputs they
had not encountered during training, representing the aforementioned novel link conditions.

\medskip
\noindent
\experiment{High Packet Loss.}\label{exp:auroraShort} In this experiment, we 
trained over
$100$ Aurora
agents in the original (in-distribution) environment.
Out of these, 
we selected $k=16$ agents that achieved a high
average reward in-distribution (see
Fig.~\ref{subfig:auroraRewards:inDist}). 
Next, we evaluated these agents on OOD inputs that are included in the previously described domain. The main difference
between the training distribution and the new (OOD) ones is the possibility of extreme packet loss rates upon initialization.

\begin{figure}[htbp]
    \vspace{-0.6cm}
    \centering
    \captionsetup[subfigure]{justification=centering}
    \captionsetup{justification=centering}
    \begin{subfigure}[t]{0.49\linewidth}
         \includegraphics[width=\textwidth, height=0.67\textwidth]{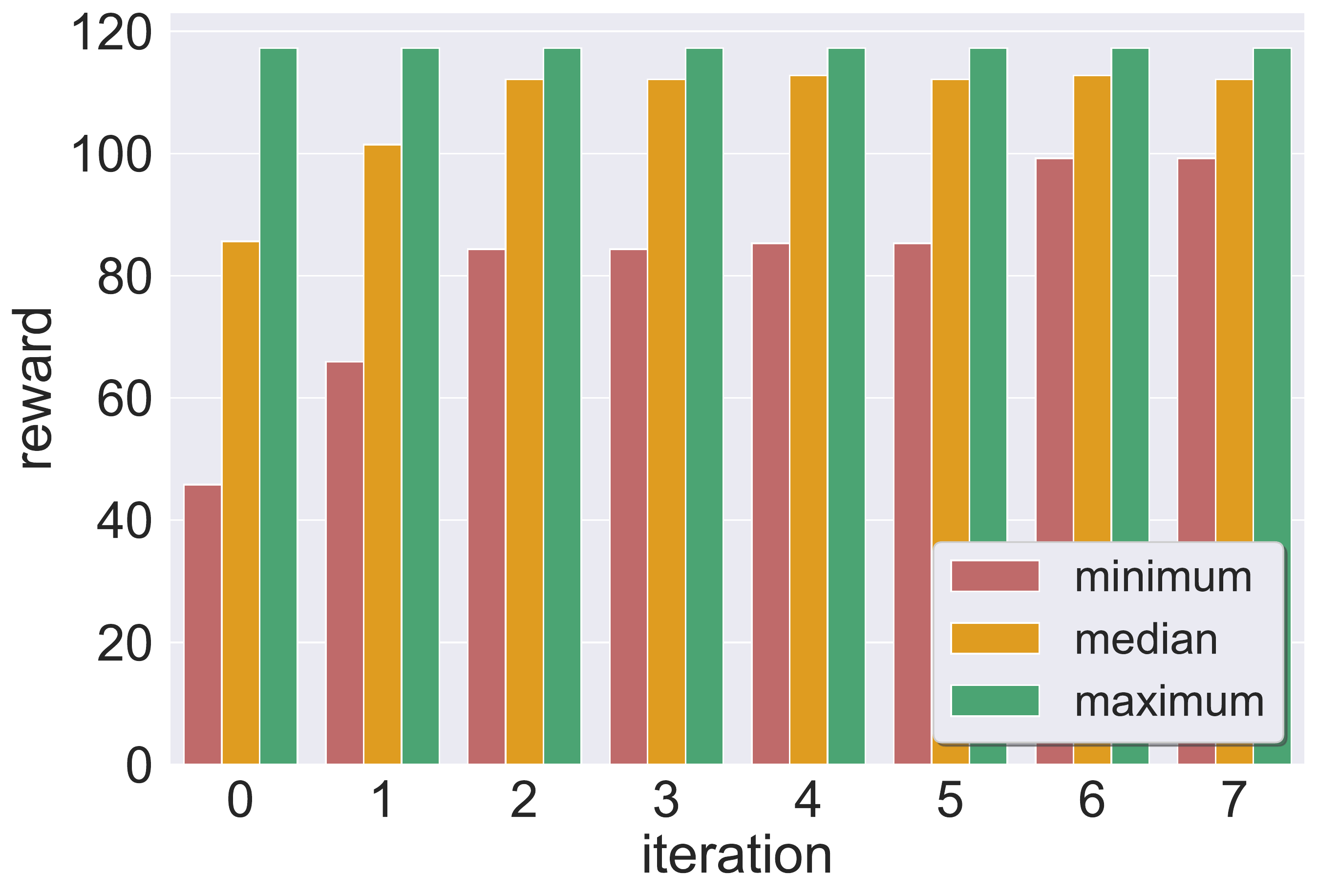}
         
         \caption{Reward statistics of remaining models}
         \label{fig:auroraShortMinMaxRewardsAlongIterations}
     \end{subfigure}
     \hfill
     \begin{subfigure}[t]{0.49\linewidth}
        \includegraphics[width=\textwidth, height=0.67\textwidth]{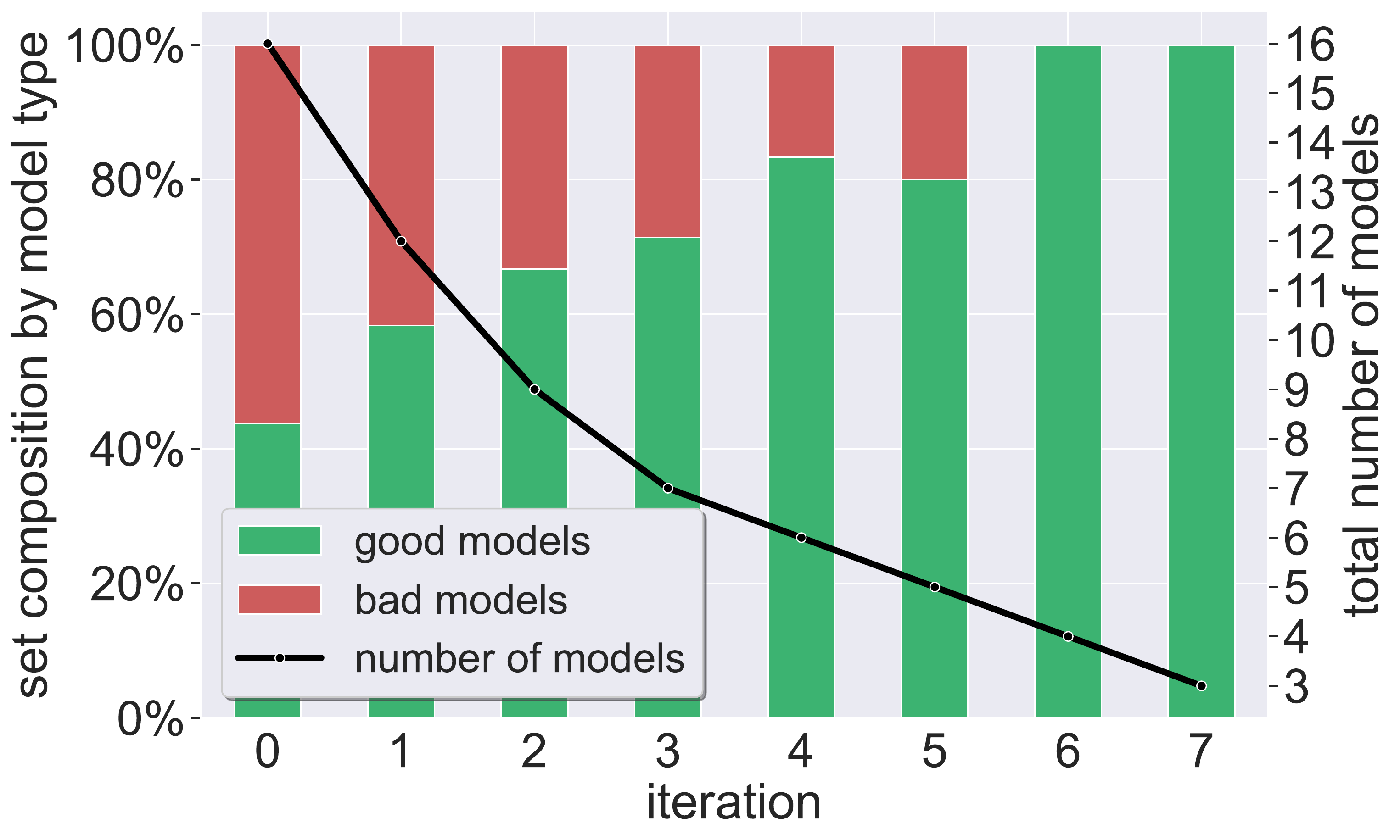}
         
         \caption{Ratio between good/bad models}
         \label{fig:auroraShortGoodBadModelsPercentages}
     \end{subfigure}
    
    \caption{Aurora: Alg.~\ref{alg:modelSelection}'s results, per iteration.} 
    \label{fig:AuroraShortTrainingMinMaxRewardAndGoodBadRatioPerIteration}

    \vspace{-0.3cm}
\end{figure}

Our evaluation over the OOD inputs, within the domain, indicates that although all $16$
models performed well in-distribution, only $7$ agents could
successfully handle such OOD inputs (see
Fig.~\ref{subfig:auroraRewards:OOD}).
When we ran Alg.~\ref{alg:modelSelection} on the 16 models, it was able to
filter out \emph{all} $9$ models that generalized poorly on the OOD inputs (see
Fig.~\ref{fig:AuroraShortTrainingMinMaxRewardAndGoodBadRatioPerIteration}).
In particular, our method returned model $\{16\}$, which is the
best-performing model according to our simulations. 
We note that in the first iterations, the four models
to be filtered out were models $\{1, 2, 6, 13\}$, which
are indeed the four worst-performing models on the OOD inputs  (see 
Appendix~\ref{sec:appendix:AuroraSupplementaryResults}).


\experiment{Additional Distributions over OOD Inputs.}
To further demonstrate that, in the specified input domain, our method is indeed likely to keep better-performing models while removing bad models, we reran the previous Aurora experiments for additional distributions (probability density functions) over the OOD inputs.
Our evaluation reveals that all models removed by
Alg.~\ref{alg:modelSelection} achieved low reward values also for these additional
distributions (see Appendix~\ref{sec:appendix:AuroraSupplementaryResults}).  
These
results highlight an important advantage of our approach: it applies
to all inputs within the considered domain, and so it applies to \emph{all
distributions over these inputs}. 


\medskip
\noindent{\textbf{Additional Experiments.}}
We also generated a new set of Aurora models by altering the training
process to include significantly longer interactions. 
We then repeated the aforementioned experiments. The results 
(summarized in Appendix~\ref{sec:appendix:AuroraSupplementaryResults})
demonstrate that our approach (again) successfully selected a
subset of models that generalizes well to
distributions over the OOD input domain.

\subsection{Comparison to Additional Methods}
\label{subsec:gradientAttackComparison}

\emph{Gradient-based methods}~\cite{HuPaGoDuAb17,MaMaScTsVl17,MaDiMe20, 
KuGoBe16} are optimization
algorithms capable of finding DNN inputs that satisfy prescribed
constraints, similarly to verification methods.  These algorithms
are extremely popular due to their simplicity and
scalability. However, this comes at the cost of being inherently incomplete and 
not as precise as DNN verification~\cite{WuZeKaBa22, AmZeKaSc22}. Indeed, when 
modifying our algorithm to calculate PDT scores with gradient-based methods, 
the results (summarized in
Appendix~\ref{sec:appendix:gradientAttacks}) reveal that, in our context, 
the verification-based approach
is superior to the gradient-based ones. Due to the
incompleteness of gradient-based approaches~\cite{WuZeKaBa22}, they
often computed sub-optimal PDT values, resulting in models that generalize 
poorly being retained.


\medskip
\noindent
\emph{Predictive uncertainty methods}~\cite{OvFeReNaScNoDiLaSn19, 
	AbPoHuReLiGhFiCaKhAcMaNa21} are \emph{online} methods for assessing uncertainty with respect 
	to observed inputs, to determine whether an encountered input is drawn from 
	the training distribution. We ran an experiment comparing our approach  to 
uncertainty-prediction-based model selection:  
we generated ensembles~\cite{KrVe94, Di00, GaHuMaTaSu22} of our original 
models, and used a variance-based metric 
(motivated by~\cite{LoSeSc20}) to identify subsets of models with low output 
variance on 
OOD-sampled inputs. Similar to gradient-based methods, predictive-uncertainty 
techniques proved fast and scalable, but lacked the precision afforded by 
verification-driven model selection and were unable to discard poorly 
generalizing models. For example, when ranking Cartpole models by their 
uncertainty on OOD inputs, the three 
models with the lowest uncertainty included also ``bad” models, which had been 
filtered out by our approach.

\section{Related Work}
\label{sec:RelatedWork}
Recently, a plethora of approaches and tools have been put forth for
ensuring DNN correctness~\cite{Eh17, KaBaDiJuKo17, KaBaDiJuKo21,
	KaHuIbJuLaLiShThWuZeDiKoBa19, HuKwWaWu17, KuKaGoJuBaKo18,
	GoKaPaBa18, SiGePuVe19, GeMiDrTsChVe18, TjXiTe17, LoMa17,
	KoLoJaBl20, WuOzZeIrJuGoFoKaPaBa20, SeDeDrFrGhKiShVaYu18, PoAbKr20,
	OkWaSeHa20, GoPaPuRuSa21, LyKoKoWoLiDa20, XiTrJo18, BuTuToKoMu18,
	VaPeWaNiSiKh22, BaGiPa21, DuChSa19, DuJhSaTi18b, SuKhSh19, FuPl18,
	GeLeXuWaGuSi22,RuHuKw18, IsBaZhKa22, UrChWuZh20, SoTh19,
	YaYaTrHoJoPo21, GoAdKeKa20, DoSuWaWaDa20, UsGoSuNoPa21, ZhShGuGuLeNa20, 
	JaBaKa20,AlAvHeLu20}, including techniques for DNN 
	shielding~\cite{LuScHe21},
optimization~\cite{AvBlChHeKoPr19, StWuZeJuKaBaKo21}, quantitative
verification~\cite{BaShShMeSa19}, abstraction~\cite{PrAf20, AnPaDiCh19, 
SiGePuVe19, ZeWuBaKa22, OsBaKa22,
	AsHaKrMo20}, size reduction~\cite{Pr22}, and more. 
Non-verification techniques, including runtime-monitoring~\cite{HaKrRiSc22}, 
ensembles~\cite{YaZeZhWu13, OsAsCa18, RoScTa20, OrCaMa22}  and additional 
methods~\cite{PaGaKoKrKoSo18} have been utilized for OOD input detection.

In contrast to the above approaches, we aim to establish 
\textit{generalization
	guarantees} with respect to an \textit{entire input domain} (spanning all 
distributions across this domain).
In addition, to the best of our knowledge, ours is the first
attempt to exploit variability across models for distilling a subset thereof, 
with improved \emph{generalization} capabilities. In particular, it is also the 
first approach to apply formal verification for this purpose.

\section{Conclusion}
\label{sec:Conclusion}

This work describes a novel, verification-driven
approach for identifying DNN models that generalize well to an input domain
of interest. We presented an iterative scheme that employs a backend DNN 
verifier, allowing us to score models based on their ability to
produce similar outputs on the given domain. We demonstrated
extensively that this approach indeed distills models capable of good 
generalization. As DNN verification technology matures, our
approach will become increasingly scalable, and also applicable to a wider
variety of DNNs. 


\medskip
\noindent
\textbf{Acknowledgements.}  
The work of Amir, Zelazny, and Katz was
partially supported by the Israel Science Foundation (grant number
683/18). 
The work of Amir was supported by a scholarship from the Clore Israel 
Foundation.
The work of Maayan and Schapira was partially supported by
funding from Huawei.
	
{
\bibliographystyle{abbrv}
\bibliography{references}
}

\newpage
\appendix
\renewcommand{\thesection}{\Alph{section}}
	
\section*{\huge Appendices}
	
\section{Training and Evaluation}
	\label{sec:appendix:trainingAndEvaluation}

    In this appendix, we elaborate on the hyperparameters and the training 
    procedure, for reproducing all models and environments of all three 
    benchmarks. We also provide a thorough overview of various implementation 
    details. The code is based on the
    \textit{Stable-Baselines 3}~\cite{RaHiGlKaErDo21} and \textit{OpenAI 
    Gym}~\cite{BrChPeScScTaZa16} packages. Unless stated otherwise, the values 
    of the various parameters used during training and evaluation are the 
    default values (per training algorithm, environment, etc.).
    Our original code and models are publicly available 
    online~\cite{ArtifactRepository}.

    \subsection{Training Algorithm}
    We trained our models with \emph{Actor-Critic} algorithms. These are state-of-the-art RL training algorithms that iteratively optimize two
    neural networks: 
    
    \begin{itemize}
    \item a \textit{critic} network, that learns a value
    function~\cite{MnKaSi13} (also known as a \emph{Q-function}), that assigns a
    value to each $\langle$state,action$\rangle$ pair; and 
    \item an \textit{actor} network, which is the DRL-based agent trained by the algorithm. This network iteratively maximizes the value function
    learned by the critic, thus improving the learned policy.
    \end{itemize}
    
    Specifically, we used two implementations of Actor-Critic algorithms: \textit{Proximal Policy Optimization} (PPO)~\cite{ShWoDh17} and
    \emph{Soft Actor-Critic} (SAC)~\cite{HaZhAbLe18}. 
    
    Actor-Critic algorithms are considered very advantageous, due to their typical requirement of relatively few samples to learn from, and also due to their ability to allow the agent to learn policies for continuous spaces of
    $\langle$state,action$\rangle$ pairs. 
    
    In each training process, all models were trained using the same 
    hyperparameters, with the exception of the \textit{Pseudo Random Number 
    Generator's (PRNG) seed}. Each training phase consisted of $10$ 
    checkpoints, while each checkpoint included a constant number of 
    environment steps, as described below. For model evaluation, we used the 
    last checkpoint of each training process (per benchmark).

    \subsection{Architecture}
    In all benchmarks, we used DNNs with a feed-forward architecture. We 
    refer the reader to Table~\ref{table:benchmarksTrainingInfo} for a summary 
    of the chosen architecture per each benchmark.

\begin{table}[ht]
    \centering
    \begin{tabular}{| P{0.2\linewidth} | P{0.15\linewidth} | P{0.2\linewidth} | P{0.3\linewidth} | P{0.15\linewidth} |}
        \hline
        \textbf{benchmark} & \textbf{hidden layers} & \textbf{layer size} & \textbf{activation function} & \textbf{training algorithm} \\ \hline
        Cartpole	& 2	& [32, 16]	& \relu	& PPO \\ \hline
        Mountain Car & 2 & [64, 16] & \relu & SAC \\ \hline
        Aurora & 2 & [32, 16] & \relu & PPO \\ \hline
    \end{tabular}
     \newline\newline
    \caption{DNN architectures and training algorithms, per benchmark.}
    \label{table:benchmarksTrainingInfo}
\end{table}

\newpage

\subsection{Cartpole Parameters}
    \label{subsec:appendix:trainingAndEvaluation:Cartpole}

    \hfill

    \subsubsection{Architecture and Training}
    
    \begin{enumerate}
         
    \item
    \textbf{Architecture}
    \begin{itemize}
	\item \textit{hidden layers}: 2
	\item \textit{size of hidden layers}: 32 and 16, respectively
        \item \textit{activation function}: \relu
	\end{itemize}
    
    \item
    \textbf{Training}
    \begin{itemize}
    \item \textit{algorithm}: Proximal Policy Optimization (PPO)
    \item \textit{gamma ($\gamma$)}: 0.95
	\item \textit{batch size}: 128
 
	\item \textit{number of checkpoints}: $10$
        \item \textit{total time-steps} (number of training steps for each checkpoint): $50,000$
        \item \textit{PRNG seeds} (each one used to train a different model):\\
        $\{1, 2, 3, 4, 5, 6, 7, 8, 9, 10, 11, 12, 13, 14, 15, 16\}$
    \end{itemize}
    
\end{enumerate}

    \begin{figure}[ht]
    \centering
    \captionsetup{justification=centering}
        \includegraphics[width=0.6\textwidth]{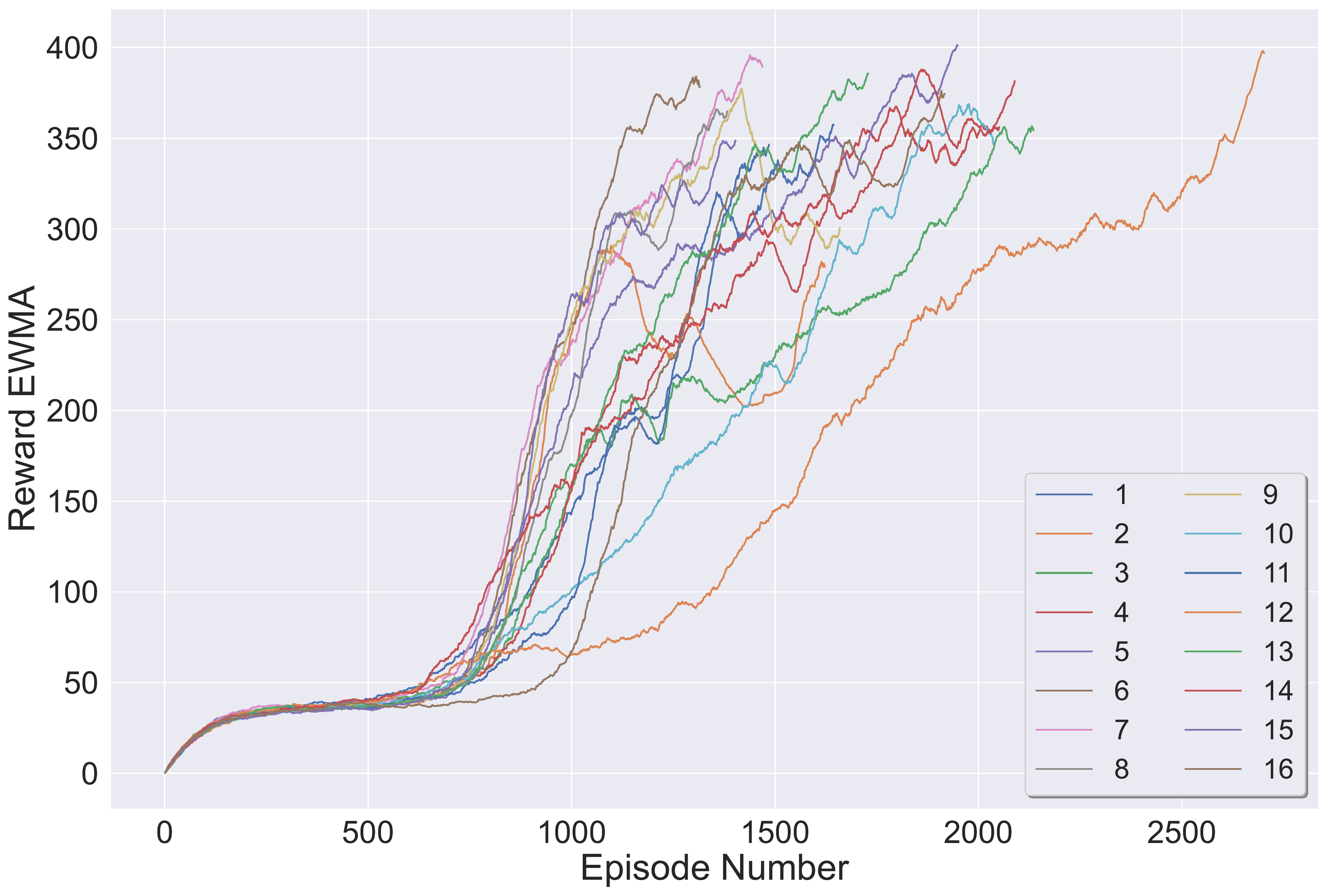}
    \caption{Cartpole: models' reward exponential weighted moving average 
    (EWMA). All models achieved a high reward (at the end of their training).}
    \label{fig:cartpole:trainingRewards}
\end{figure}

    \subsubsection{Environment}
    \hfill
    
    \noindent We used the configurable \textit{CartPoleContinuous-v0} environment.
    Given lower and upper bounds for the x-axis location, denoted as $[low,high]$, and $mid=\frac{high+low}{2}$, the initial x position is randomly, uniformly drawn from the interval $[mid-0.05, mid+0.05]$.
    
    An \emph{episode} is a sequence of agent interactions with the  environment, which ends when a terminal state is reached. In the Cartpole environment, an episode terminates after the first of the following occurs: 
        \begin{enumerate}
            \item The cart's location exceeds the platform's boundaries (as expressed via the $x$-axis location); or
            
            \item The cart was unable to balance the pole, which fell (as expressed via the $\theta$-value); or
        
            \item $500$ time-steps have passed
            
        \end{enumerate}

    \hfill

   \subsubsection{Domains}
    \begin{enumerate}
   
    \item \textbf{(Training) In-Distribution}
    
    \begin{itemize}
        \item \textit{action min magnitude}: True
        \item \textit{x-axis lower bound} (x\_threshold\_low): $-2.4$
        \item \textit{x-axis upper bound} (x\_threshold\_high): $2.4$
    \end{itemize}
    
    \item  \textbf{(OOD) Input Domain}
    
    Two symmetric OOD scenarios were evaluated: the cart's $x$ position represented significantly extended platforms in a single direction, hence, including areas previously unseen during training. Specifically, we
    generated a domain of input points characterized by $x$-axis
    boundaries that were selected, with an equal probability, either from 
    $[-10, -2.4]$ or from $[2.4, 10]$ (instead of the in-distribution range of 
    $[-2.4,2.4])$. The cart's initial location was uniformly drawn from the 
    range's \emph{center} $\pm0.05$: $[-6.4-0.05, -6.4+0.05]$ and $[6.4-0.05, 
    6.4+0.05]$, respectively. 
    
    \noindent All other parameters were the same as the ones used in-distribution.
    
    \ \\
    
    \noindent \underline{OOD scenario 1}
    \begin{itemize}
        \item \textit{x-axis lower bound} (x\_threshold\_low): $-10.0$
        \item \textit{x-axis upper bound} (x\_threshold\_high): $-2.4$
    \end{itemize}
    \hfill
    
    \noindent \underline{OOD scenario 2}
     \begin{itemize}
        \item \textit{x-axis lower bound} (x\_threshold\_low): $2.4$
        \item \textit{x-axis upper bound} (x\_threshold\_high): $10.0$
    \end{itemize}
     
\end{enumerate}

    \subsection{Mountain Car Parameters}
    \label{subsec:appendix:trainingAndEvaluation:mountaincar}
    
    \subsubsection{Architecture and Training}

    \hfill

    \begin{enumerate}
         
    \item
    \textbf{Architecture}

    \begin{itemize}
	\item \textit{hidden layers}: 2
	\item \textit{size of hidden layers}: 64 and 16, respectively
        \item \textit{activation function}: \relu
	\item \textit{clip mean} parameter: 5.0
        \item \textit{log stdinit} parameter: -3.6
	\end{itemize}

    \item
    \textbf{Training}

    \begin{itemize}
    \item \textit{algorithm:} Soft Actor Critic (SAC)
    \item \textit{gamma ($\gamma$)}: 0.9999	
     \item \textit{batch size}: 512
        \item \textit{buffer size}: 50,000
        
        \item \textit{gradient steps}: 32
        \item \textit{learning rate}: $\num{3e-4}$
        \item \textit{learning starts}: 0
        \item \textit{tau ($\tau$)}: 0.01
        \item \textit{train freq}: 32
        \item \textit{use sde}: True

	\item \textit{number of checkpoints}: $10$
        \item \textit{total time-steps} (number of training steps for each checkpoint): $5,000$
        \item \textit{PRNG seeds} (each one used to train a different model):\\
        $\{1, 2, 3, 4, 5, 6, 7, 8, 9, 10, 11, 12, 13, 14, 15, 16\}$
    \end{itemize}

\end{enumerate}

\begin{figure}[ht]
    \centering
    \captionsetup{justification=centering}
        \includegraphics[width=0.6\textwidth]{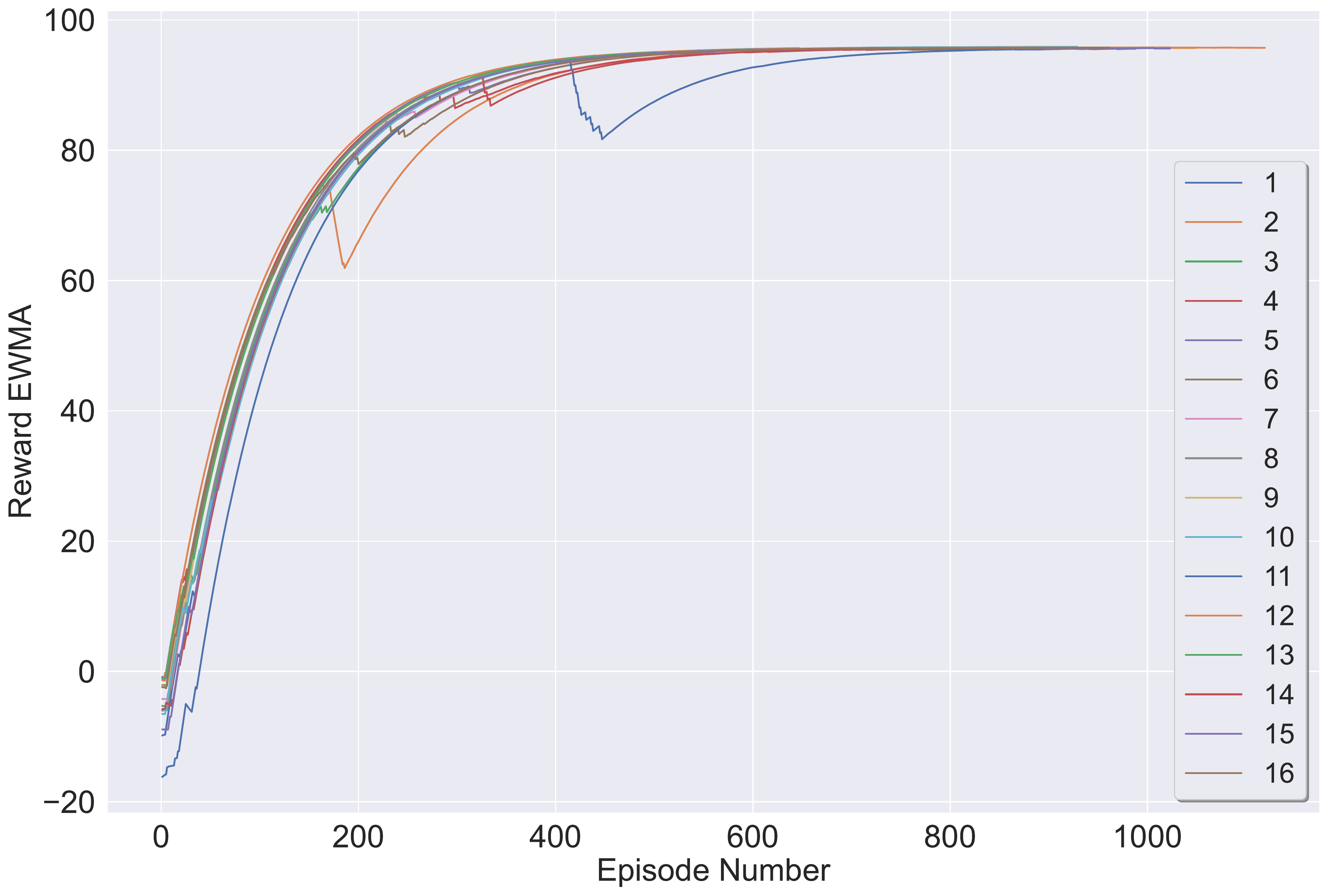}
    \caption{Mountain Car: models' reward exponential weighted moving average 
    (EWMA). All models achieved a high reward (at the end of their training).}
    \label{fig:mountaincar:trainingRewards}
\end{figure}

    \subsubsection{Environment}
    \hfill
    
    \noindent We used the \textit{MountainCarContinuous-v1} environment.
    
    \hfill

     \subsubsection{Domains}
    \begin{enumerate}
   
    \item \textbf{(Training) In-Distribution}
    
    \begin{itemize}
        \item \textit{min position}: $-1.2$
        \item \textit{max position}: $-0.6$
        \item \textit{goal position}: $0.45$
        \item \textit{min action} (if the agent's action is negative and under this value, this value is used): $-2$
        \item \textit{max action} (if the agent's action is positive and above this value, this value is used): $2$
        \item \textit{max speed}: $0.4$
        \item \textit{initial location range} (from which the initial location is uniformly drawn): $[-0.9, -0.6]$
        \item \textit{initial velocity range} (from which the initial velocity is uniformly drawn): $[0, 0]$ (i.e., the initial velocity in this scenario is always $0$) 
        \item \textit{x scale factor} (used for scaling the x-axis): $1.5$
    \end{itemize}
    
    \item  \textbf{(OOD) Input Domain}
    
    The inputs are the same as the ones used in-distribution, except for the 
    following:
    \begin{itemize}
        \item \textit{min position}: $-2.4$
        \item \textit{max position}: $1.2$
        \item \textit{goal position}: $0.9$
        \item \textit{initial location range}: $[0.4, 0.5]$
        \item \textit{initial location velocity}: $[-0.4, -0.3]$
    \end{itemize}

\end{enumerate}

\subsection{Aurora Parameters}
    \label{subsec:appendix:trainingAndEvaluation:aurora}


 \hfill
 
\subsubsection{Architecture and Training}

\begin{enumerate}
     
\item
\textbf{Architecture}
   
    \begin{itemize}
	\item \textit{hidden layers}: 2
	\item \textit{size of hidden layers}: 32 and 16, respectively
        \item \textit{activation function}: \relu
    \end{itemize}

\item
    \textbf{Training}
    \begin{itemize}
    \item \textit{algorithm}: Proximal Policy Optimization (PPO)
    
    \item \textit{gamma ($\gamma$)}: 0.99
        \item \textit{number of steps to run for each environment, per update} 
        (n\_steps): $8,192$
        \item \textit{Number of epochs when optimizing the surrogate loss} 
        (n\_epochs): $4$
        \item \textit{learning rate}: $\num{1e-3}$
        \item \textit{Value function loss coefficient} (vf\_coef): $1$
        \item \textit{Entropy function loss coefficient} (ent\_coef): 
        $\num{1e-2}$

	\item \textit{number of checkpoints}: $6$
        \item \textit{total time-steps} (number of training steps for each checkpoint): $656,000$ (as used in the original paper~\cite{JaRoGoScTa19})
        \item \textit{PRNG seeds} (each one used to train a different model): 
        \\$\{4, 52, 105, 666, 850, 854, 857, 858, 885, 897, 901, 906, 907, 929, 944, 945\}$ \\
        We note that for simplicity, these were mapped to indices $\{1 \ldots 16\}$, accordingly (e.g., $\{4\} \rightarrow \{1\}$, $\{52\} \rightarrow \{2\}$, etc.).
    
    \end{itemize}

\end{enumerate}

    \begin{figure}[ht]
    \centering
    \captionsetup{justification=centering}
        \includegraphics[width=0.6\textwidth]{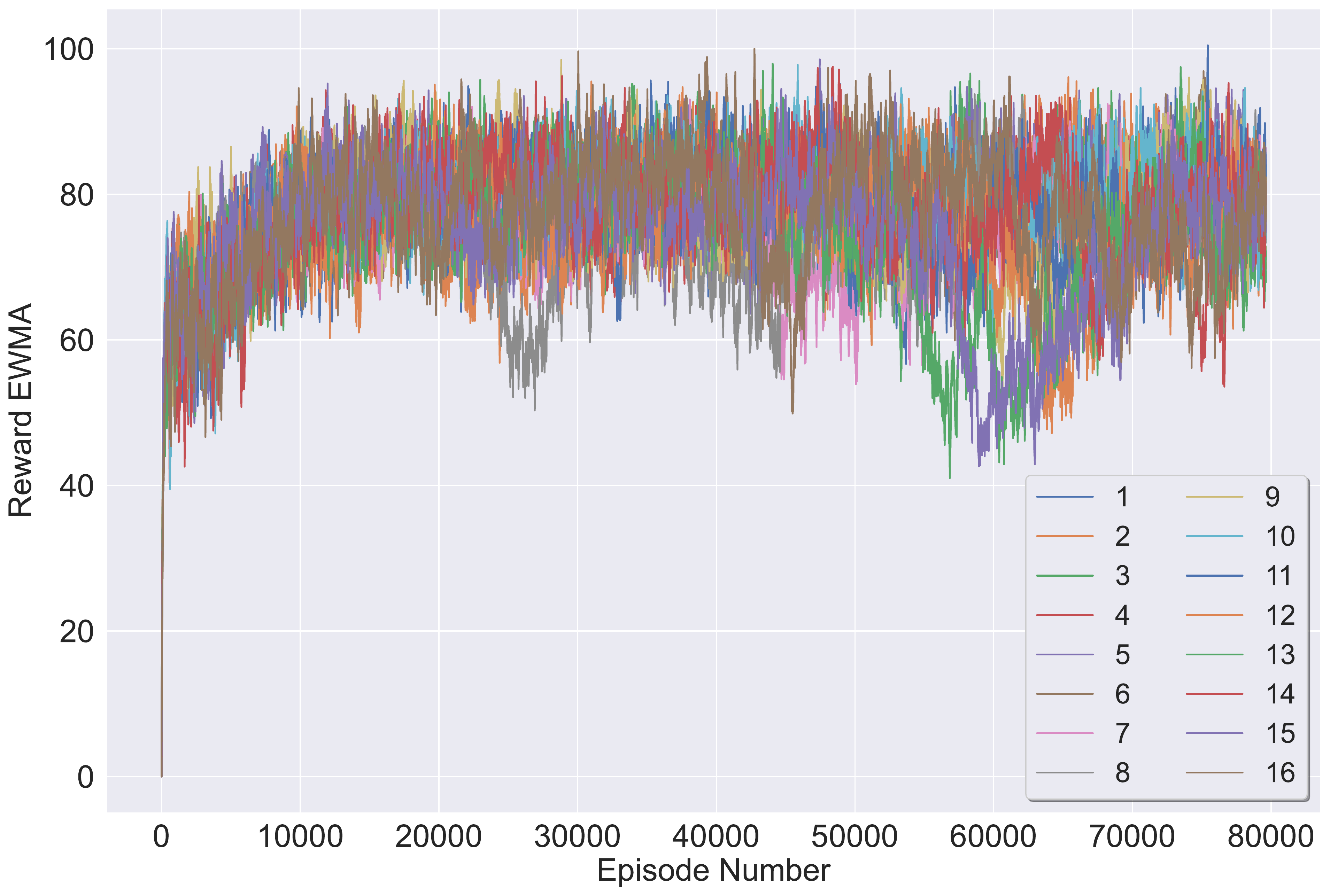}
    \caption{Aurora (short training): models' reward exponential weighted 
    moving average (EWMA) during training. All models achieved a high reward 
    (at the end of their training).}
    \label{fig:auroraShort:trainingRewards}
    \end{figure}

    \noindent \subsubsection{Environment}
    \hfill
    
    \noindent We used a configurable version of the \textit{PccNs-v0} environment. 
    For models in Exp.~\ref{exp:auroraShort}(with the \emph{short} training), each episode consisted of $50$ steps. For models in Exp.~\ref{exp:auroraLong} (with the \emph{long} training), each episode consisted of $400$ steps.

    \hfill

   \subsubsection{Domains}
    \begin{enumerate}
   
    \item \textbf{(Training) In-Distribution}
    
    \begin{itemize}
        \item \textit{minimal initial Sending Rate ratio (to the link's 
        bandwidth)} (min\_initial\_send\_rate\_bw\_ratio): $0.3$
        \item \textit{maximal initial Sending Rate ratio (to the link's 
        bandwidth)} (max\_initial\_send\_rate\_bw\_ratio): $1.5$
    \end{itemize}
    
    \item  \textbf{(OOD) Input Domain}

     To bound the \textit{latency gradient} and \textit{latency ratio} elements of the input, we used a shallow buffer setup, with a bounding parameter $\delta>0$ such that \textit{latency gradient} $\in [-\delta, \delta]$ and \textit{latency ratio} $\in [1.0, 1.0 +\delta]$.

    \begin{itemize}
        \item \textit{minimal initial sending rate ratio (to the link's 
        bandwidth)} (min\_initial\_send\_rate\_bw\_ratio): $2.0$
        \item \textit{maximal initial sending rate ratio (to the link's 
        bandwidth)} (max\_initial\_send\_rate\_bw\_ratio): $8.0$
        \item \textit{Use shallow buffer}: True
        \item \textit{Shallow buffer $\delta$ bound parameter}: $\num{1e-2}$
        
    \end{itemize}
    
\end{enumerate}

\clearpage

    \section{Verification Queries}
    \label{sec:appendix:VerificationQueries}

    \subsection{A DNN Verification Query: Toy Example}
\label{sec:appendix:VerificationQueries:toyExample}

Let us revisit the DNN in Fig.~\ref{fig:toyDnn}. Suppose that we wish to verify that for all nonnegative
inputs the DNN outputs a
value strictly smaller than $25$, i.e., for all inputs
$x=\langle v_1^1,v_1^2\rangle \in \mathbb{R}^2_{\geq 0}$, it holds
that $N(x)=v_4^1 < 25$. This is encoded as a verification query by
choosing a precondition restricting the inputs to be non-negative, i.e., $P= ( v^1_1\geq 0 \wedge v_1^2\geq 0)$, and
by setting $Q=(v_4^1\geq 25)$, which is the \textit{negation} of our
desired property. For this specific verification query, a
sound verifier will return \sat, alongside a feasible counterexample
such as $x=\langle 0, 4\rangle$, which produces $v_4^1=28 \geq
25$. Hence, this property does not hold for the DNN described in
Fig.~\ref{fig:toyDnn}. All queries were dispatched to 
\marabou~\cite{KaHuIbJuLaLiShThWuZeDiKoBa19} --- a sound and complete 
verification engine, previously used in other DNN-verification-related 
work~\cite{AmZeKaSc22, AmWuBaKa21, AmCoYeMaHaFaKa23, CaKoDaKoKaAmRe22, 
JaBaKa20, 
WuZeKaBa22, WaPeWhYaJa18, SiGePuVe19, ElGoKa20, OsBaKa22, CoYeAmFaHaKa22, 
BaKa22, ElCoKa22,AmFrKaMaRe23, CoElBaKa22}.

    \subsection{Input Domain Queries}    
    Next, we elaborate on how we encoded the queries, which we later fed to our backend verification engine, in order to compute the PDT scores for a DNN pair. Given a DNN pair, $N_1$ and $N_2$, we execute the following stages:

    \begin{enumerate}

        \item We concatenate $N_1$ and $N_2$ to a new DNN $N_3=[N_1; N_2]$, which is roughly twice the size of each of the original DNNs (as both $N_1$ and $N_2$ have the same architecture). The input of $N_3$ is of the same original size as each single DNN and is connected to the second layer of each DNN, consequently allowing the same input to flow throughout the network to the output layer of each original DNN ($N_1, N_2$). Thus, the output layer of $N_3$ is a concatenation of the outputs of both $N_1$ and $N_2$. A scheme depicting the construction of a concatenated DNN appears in Fig.~\ref{fig:concatenatedDnns}.

 \begin{figure}[ht]
    \centering
    \captionsetup{justification=centering}
        \includegraphics[width=0.6\textwidth]{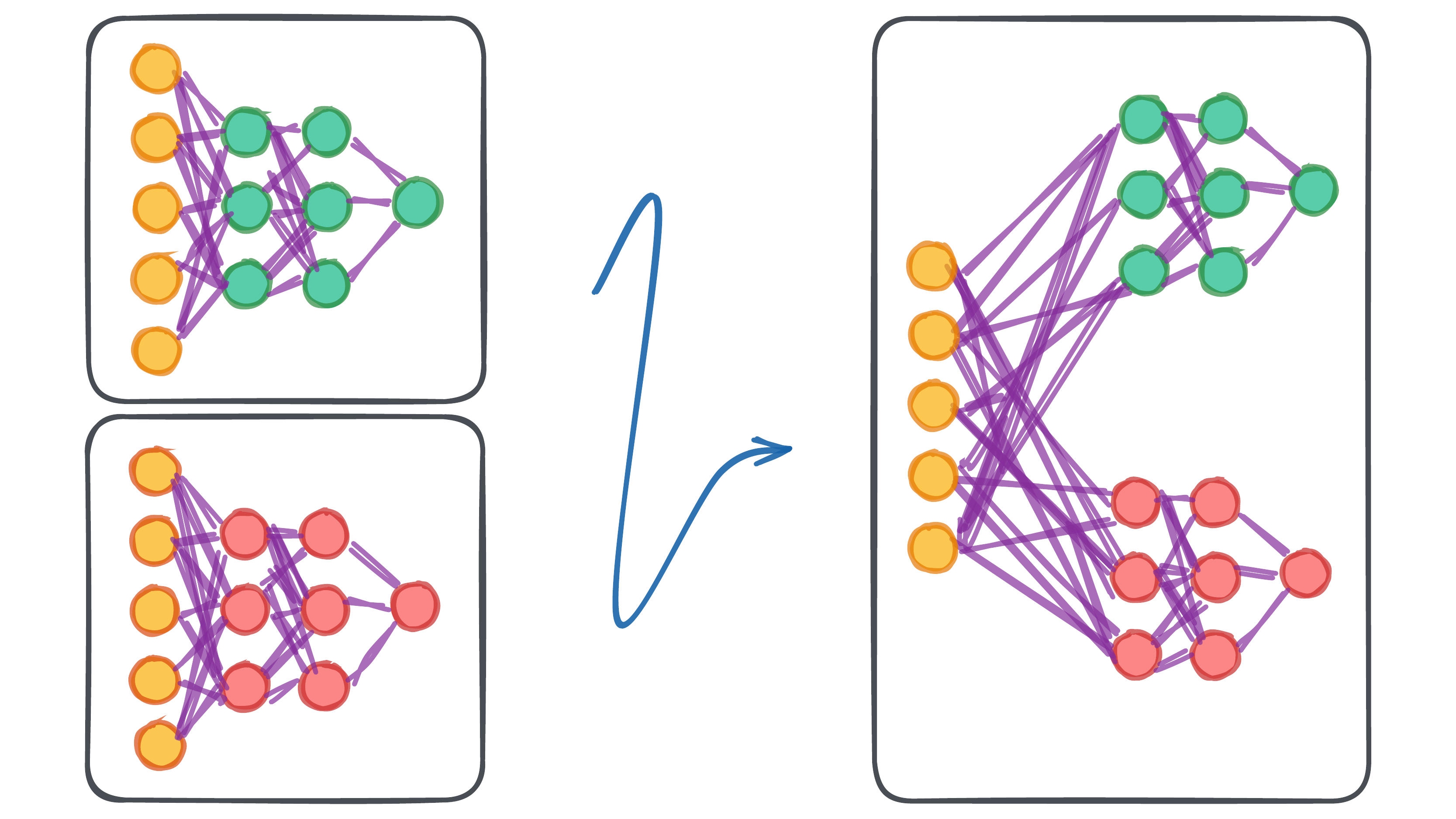}
    \caption{In order to calculate the PDT scores, we generated a new DNN which 
    is the concatenation of each pair of DNNs, sharing the same input.} 
    \label{fig:concatenatedDnns}
    \end{figure}

        \item Encoding a \emph{precondition} P which represents the ranges of the input variables. We used these inputs to include values in the input domains of interest. In some cases, these values were predefined to match the OOD setting evaluated, and in others, we extracted these values based on empirical simulations of the models post-training.
        As we mentioned before, the bounds are supplied by the system designer, based on prior knowledge of the input domain.
        In our experiments, we used the following bounds of the input domain:
        
        \begin{enumerate}
            \item \textbf{Cartpole:}
            \begin{itemize}
            
                \item x position: $x \in [-10,-2.4]$ or $x \in [2.4,10]$
                The PDT was set to be the maximum PDT score of each of these two scenarios.
                
                \item $x$ velocity: $v_{x} \in [-2.18,2.66]$
                
                \item angle: $\theta \in [-0.23, 0.23]$ 
                
                \item angular velocity: $v_{\theta} \in [-1.3, 1.22]$
                
            \end{itemize}
            
            \item \textbf{Mountain Car:}
             \begin{itemize}
             \item x position: $x \in [-2.4,0.9]$

                \item $x$ velocity: $v_{x} \in [-0.4,0.134]$
            \end{itemize}
        
            \item \textbf{Aurora:}
            \begin{itemize}
                \item \emph{latency gradient}:
                $x_{t} \in [-0.007, 0.007]$, for all $t$ s.t.
                $(t \mod 3) = 0$

                 \item \emph{latency ratio}:
                $x_{t} \in [1, 1.04]$, for all $t$ s.t.
                $(t \mod 3) = 1$

                 \item \emph{sending ratio}:
                $x_{t} \in [0.7, 8]$, for all $t$ s.t.
                $(t \mod 3) = 2$
            
            \end{itemize}

        \end{enumerate}
        
\item Encoding a \emph{postcondition} Q which encodes (for a fixed slack $\alpha$) and a given distance function $\distanceFn: \outputSpace\times\outputSpace \mapsto \mathbb{R^+}$, that for an input $x'\in \feasibleStatesSpace$ the following holds: $\distanceFn(N_{1}(x'),N_{2}(x')) \geq \alpha$

Examples of distance functions:

 \begin{enumerate}
 
           \item \textbf{$L_{1}$ norm:}

            \[\distanceFn(N_{1}, N_{2}) = \argmax_{x\in \feasibleStatesSpace}(|N_{1}(x) - N_{2}(x)|)\]

            This distance function was used in the case of the Aurora benchmark.

           \item $\mathbf{condition-distance (``c-distance'')}$\textbf{:} the 
           \emph{$c-distance$} function returns the maximal $L_{1}$ norm of two 
           DNNs, for all inputs $x \in \feasibleStatesSpace$ such that both 
           outputs $N_{1}(x)$, $N_{2}(x)$ comply to constraint $\mathbf{c}$.

           \[c-distance(N_{1}, N_{2}) \triangleq \max_{x\in \feasibleStatesSpace \text{ s.t. } N_{1}(x),N_{2}(x) \vDash c}(|N_{1}(x) - N_{2}(x)|)\]

        For the Cartpole and Mountain Car benchmarks, we defined the distance function to be:
        \[ \distanceFn (N_{1}, N_{2}) =\min_{c, c'} (c-distance (N_{1}, N_{2}), c'-distance (N_{1}, N_{2})) \] 

We chose $c: = N_{1}(x)\geq 0  \wedge N_{2}(x) \geq 0$ and $c':= N_{1}(x)\leq 0 \wedge N_{2}(x) \leq 0$

This distance function is tailored to find the maximal difference between the 
outputs (actions) of two models, in a given category of inputs (non-positive or 
non-negative, in our case). The intuition behind this function is that in some 
benchmarks, good and bad models may differ in the \emph{sign} (rather than only 
the magnitude) of their actions. For example, consider a scenario of the 
Cartpole benchmark where the cart is located on the ``edge'' of the platform: 
an 
action to one direction (off the platform) will cause the episode to end, while 
an action to the other direction will allow the agent to increase its rewards 
by continuing the episode.

    \end{enumerate}

\end{enumerate}

\clearpage
	
\section{Algorithm Variations and Hyperparameters}
	\label{sec:appendix:AlgorithmAdditionalInformation}
		
	In this appendix, we elaborate on our algorithms' additional 
	hyperparameters and filtering criteria, used throughout our evaluation. As 
	the results demonstrate, our method is highly robust in a myriad of 
	settings.

    \subsection{Precision}
    For each benchmark and each experiment, we arbitrarily selected $k=16$ models which reached our reward threshold for the in-distribution data. Then, we used these models for our empirical evaluation. The PDT scores were calculated up to a finite precision of $0.5\leq \epsilon\leq 1$, depending on the
benchmark ($1$ for Cartpole and Aurora, and $0.5$ for Mountain Car).

\subsection{Filtering Criteria}
As elaborated in Section~\ref{sec:Approach}, our algorithm iteratively filters out (Line~\ref{lst:line:modelsToRemove} in Alg.~\ref{alg:modelSelection}) models with a relatively high disagreement score, i.e., models that may disagree with their peers in the inputs domain. We present three different criteria in which we may choose the models to remove in a given iteration, after sorting the models based on their DS score:

\begin{enumerate}
     \item \conditionPercentile: in which we remove the \emph{top $p \% $} of 
     models with the highest disagreement scores, for a predefined value $p$. 
     In our experiments, we chose $p=25\%$.
     
    \item \conditionMax: in which we: \begin{enumerate}
        \item sort the DS scores of all models in a descending order
        \item calculate the difference between every two adjacent scores
        \item search for the \emph{greatest difference} of any two subsequent 
        DS scores
        \item for this difference, use the larger DS as a threshold
        \item remove all models with a DS that is greater than or equal to this threshold
    \end{enumerate}
    
   \item \conditionCombined: in which we remove models based either on \conditionMax or  
   \conditionPercentile, depending on which criterion eliminates more models in a specific iteration.  
\end{enumerate}

\clearpage

\section{Cartpole: Supplementary Results}
\label{sec:appendix:CartPoleSupplementaryResults}

Throughout our evaluations of this benchmark, we use a threshold of \textbf{250} to distinguish between \emph{good} and \emph{bad} models --- this threshold value induces a large margin from rewards of poorly-performing models (which are usually less than $100$).

\begin{figure}[ht]
    \centering
    \captionsetup{justification=centering}
    \includegraphics[width=0.5\textwidth]{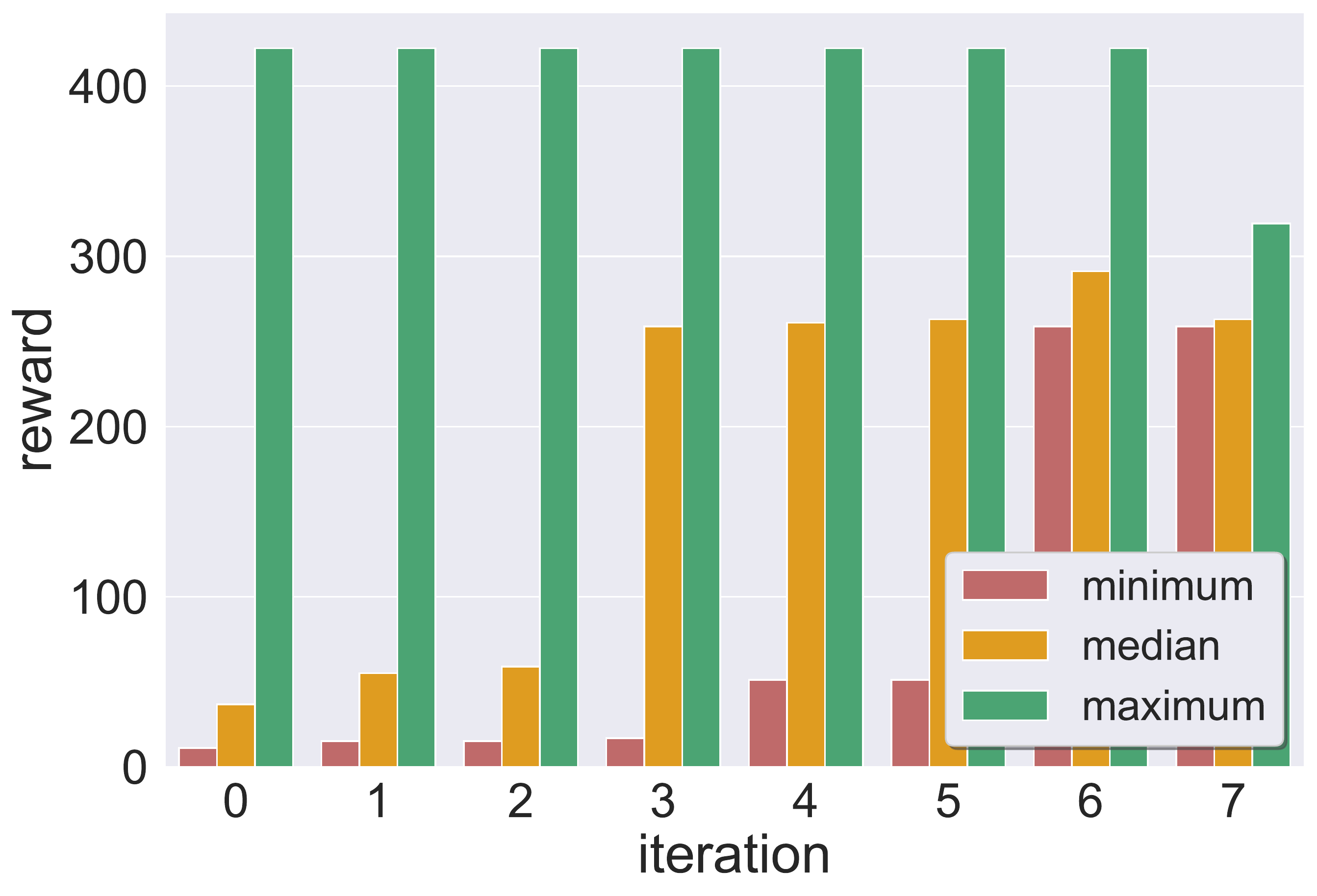}
    \caption{Cartpole: model selection results: minimum, median, and maximum 
    rewards of the models selected after each iteration.}
    \label{fig:cartpolePercentileMinMaxRewards}
\end{figure}

Note that as seen in Fig.~\ref{fig:cartpolePercentileMinMaxRewards}, our 
algorithm eventually also removes \textit{some} of the more successful models. 
However, the final result contains \textit{only} well-performing models, as in 
the other benchmarks. 

\begin{figure}[ht]
    \centering
    \subfloat[In-distribution \label{subfig:cartpoleRewards:inDist}]{\includegraphics[width=0.49\textwidth]{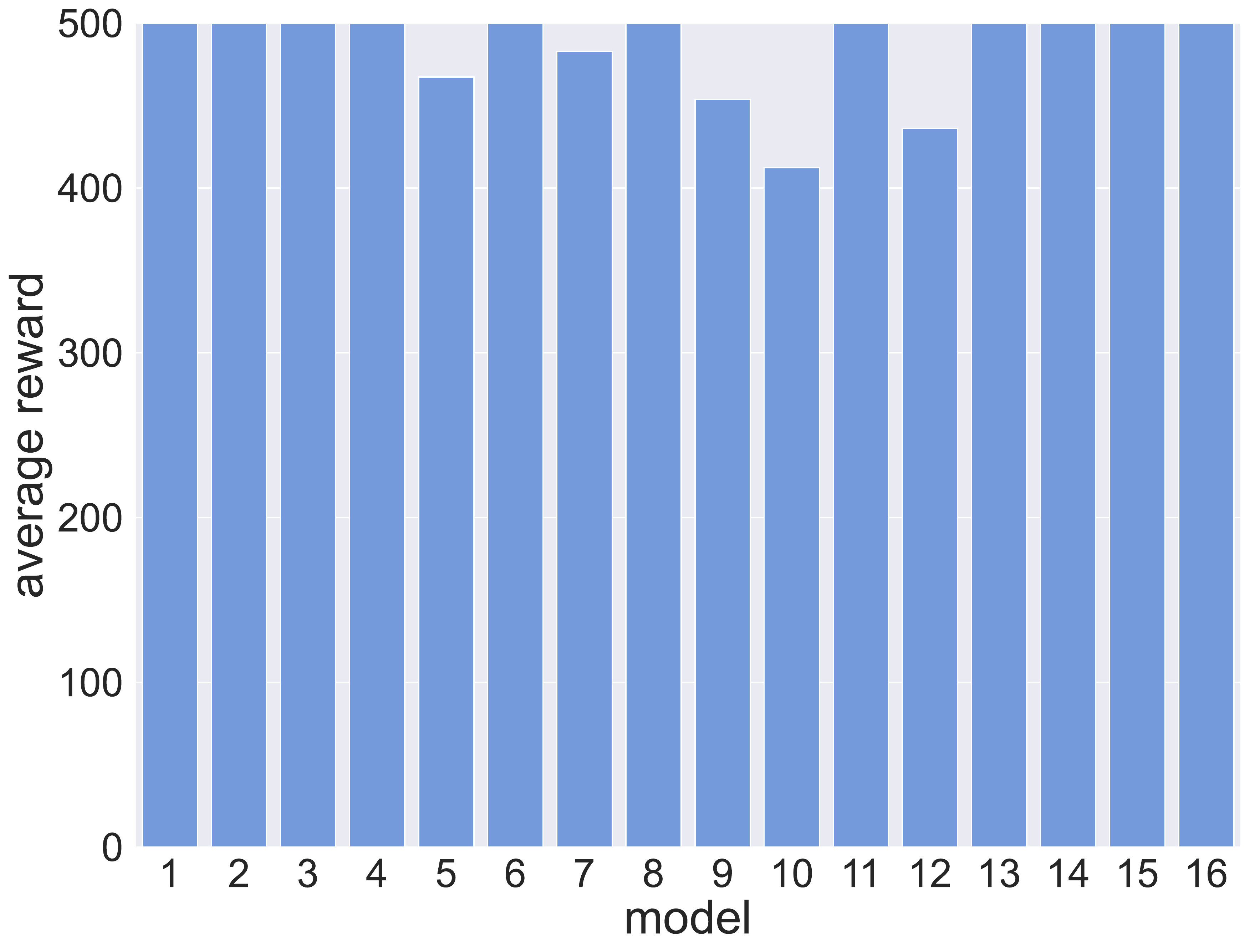}}
    \hfill
    \subfloat[OOD\label{subfig:cartpoleRewards:OOD}]{\includegraphics[width=0.49\textwidth]{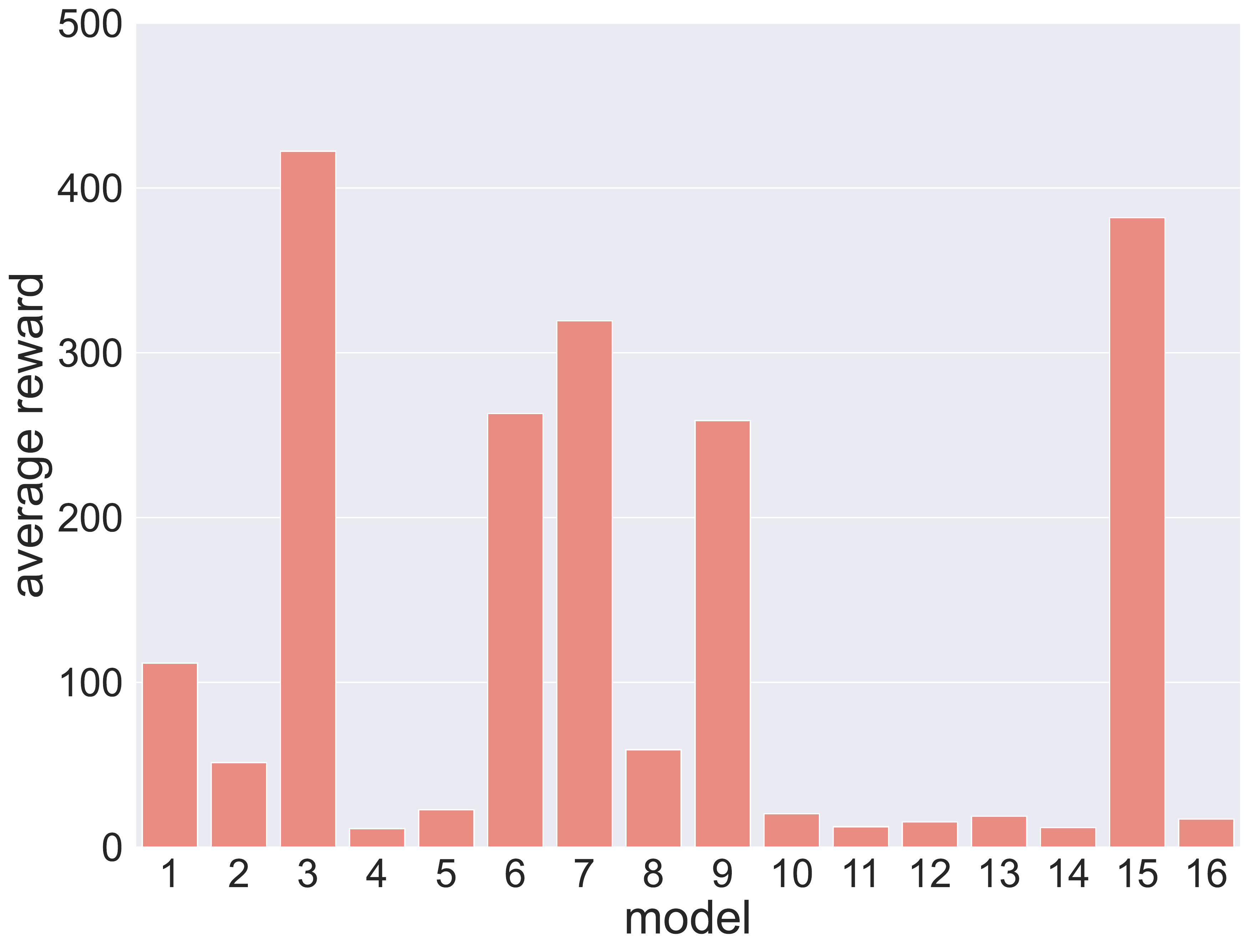}}\\
    
    \caption{Cartpole: models' average rewards in different distributions.}
    \label{fig:cartpoleRewards}
\end{figure}

\FloatBarrier

\clearpage

\subsection{Result per Filtering Criteria}

\begin{figure}[!ht]
    \centering
    \captionsetup[subfigure]{justification=centering}
    \captionsetup{justification=centering} 
     \begin{subfigure}[t]{0.49\linewidth}
         \centering
    \includegraphics[width=\textwidth]{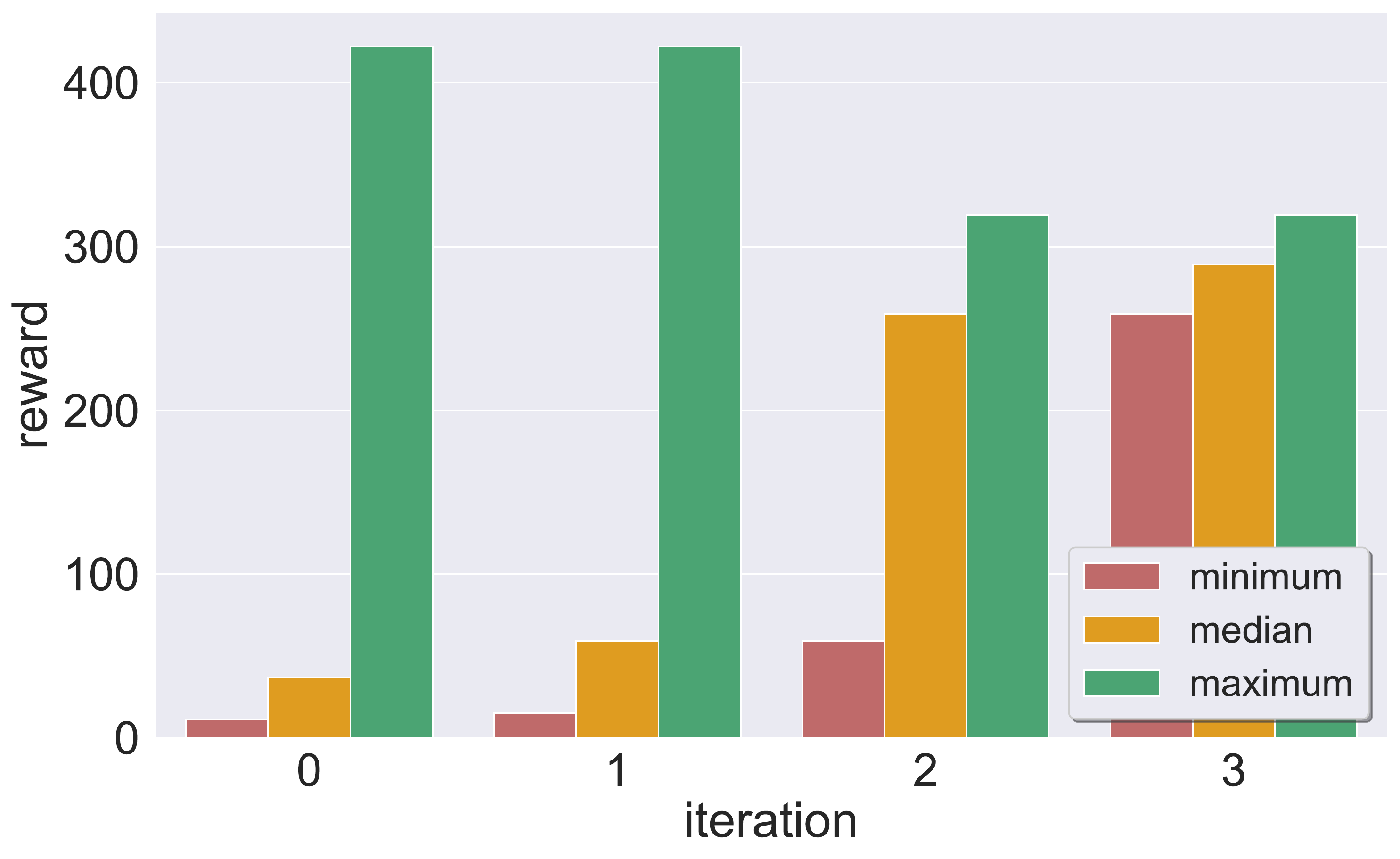}
         \caption{Rewards statistics}
        \label{}
     \end{subfigure}
     \hfill
     \begin{subfigure}[t]{0.49\linewidth}
        \includegraphics[width=\textwidth, height=0.61\textwidth]{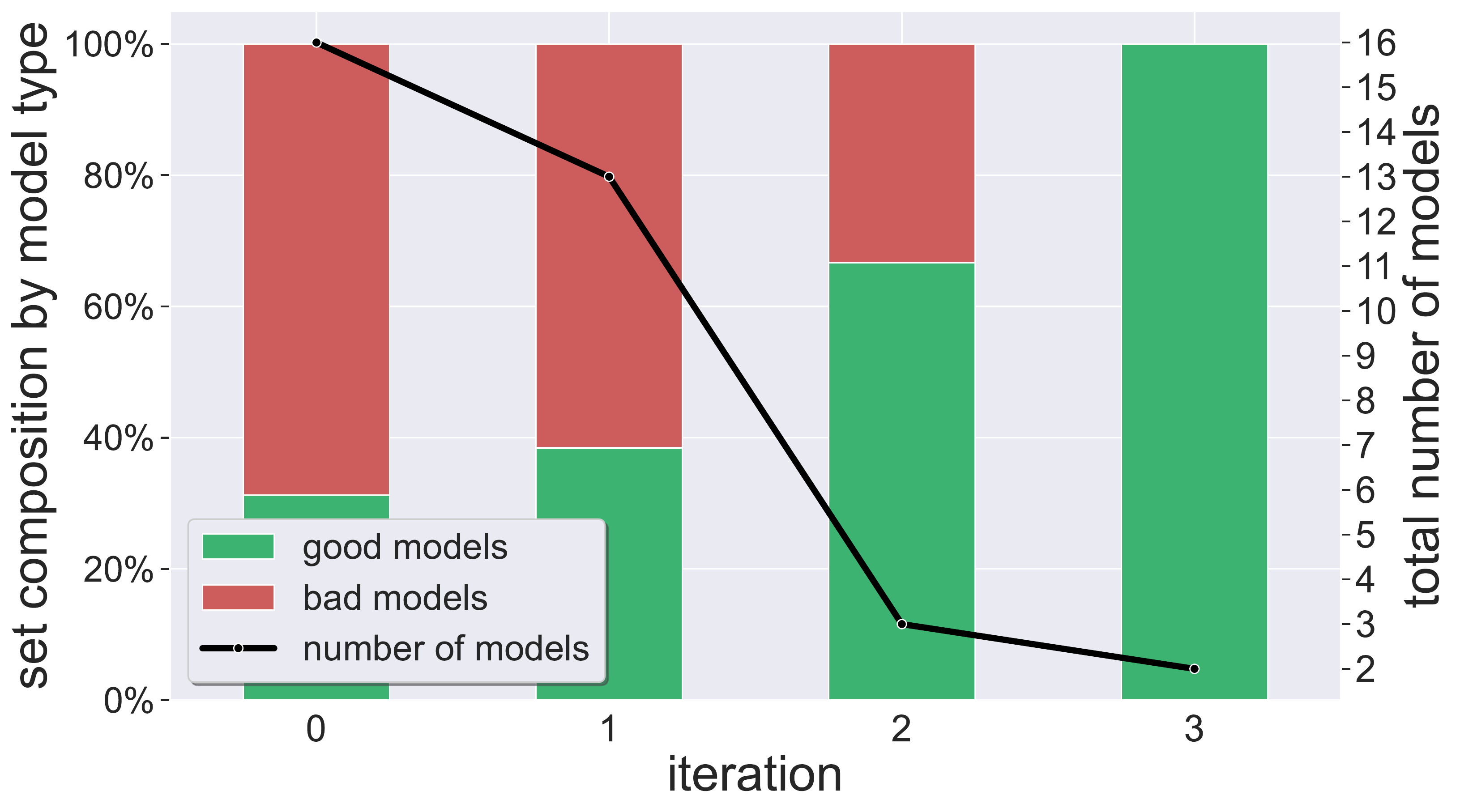}
         \caption{Remaining good and bad models ratio}
        \label{}
     \end{subfigure}
     
    \caption{Cartpole: results using the \maxAgg filtering criterion.}
    \label{fig:cartpole:maxResults}
\end{figure}

\begin{figure}[!h]
    \centering
    \captionsetup[subfigure]{justification=centering}
    \captionsetup{justification=centering} 
     \begin{subfigure}[t]{0.49\linewidth}
         \centering
    \includegraphics[width=\textwidth]{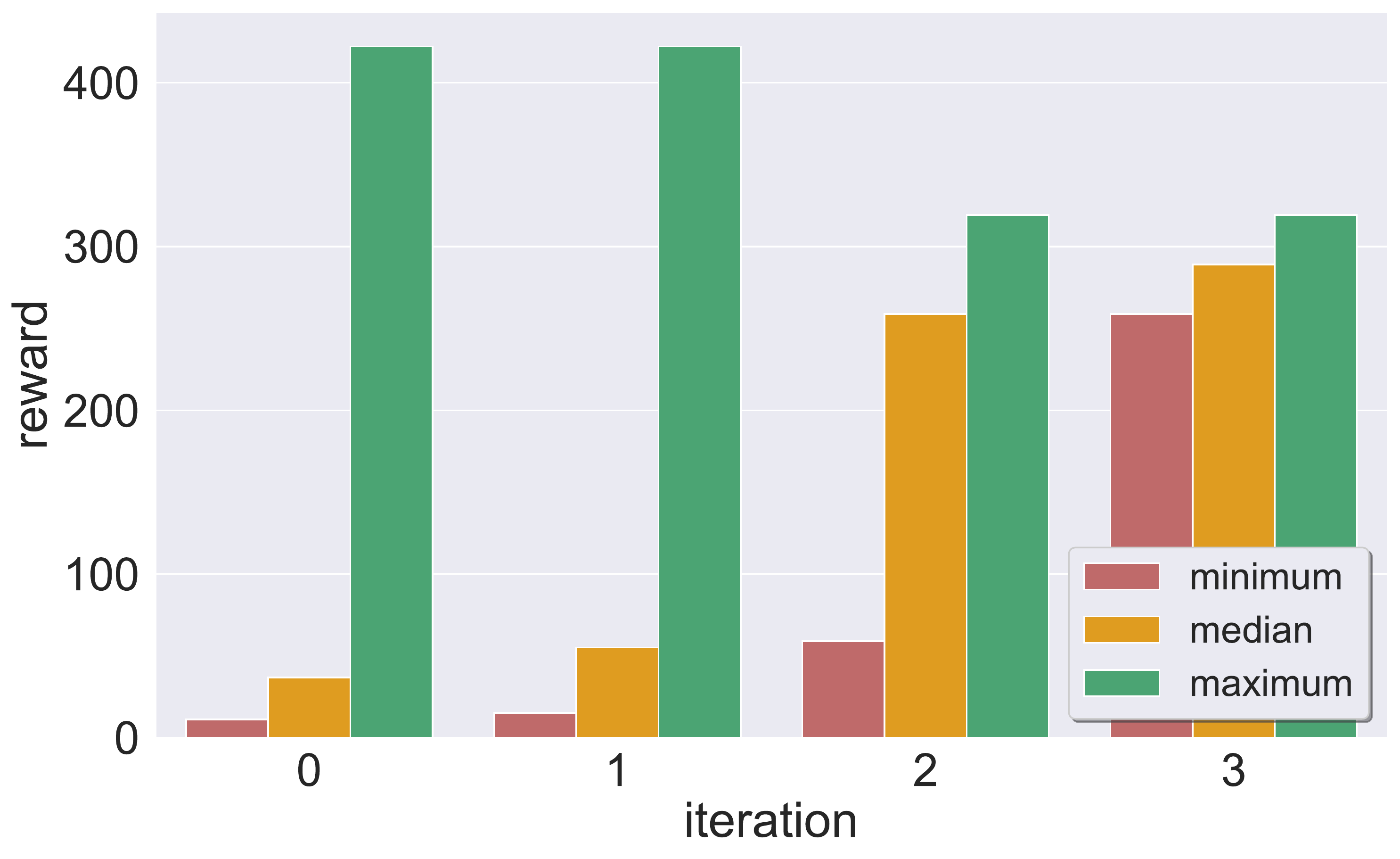}
         \caption{Rewards statistics}
        \label{}
     \end{subfigure}
     \hfill
     \begin{subfigure}[t]{0.49\linewidth}
        \includegraphics[width=\textwidth]{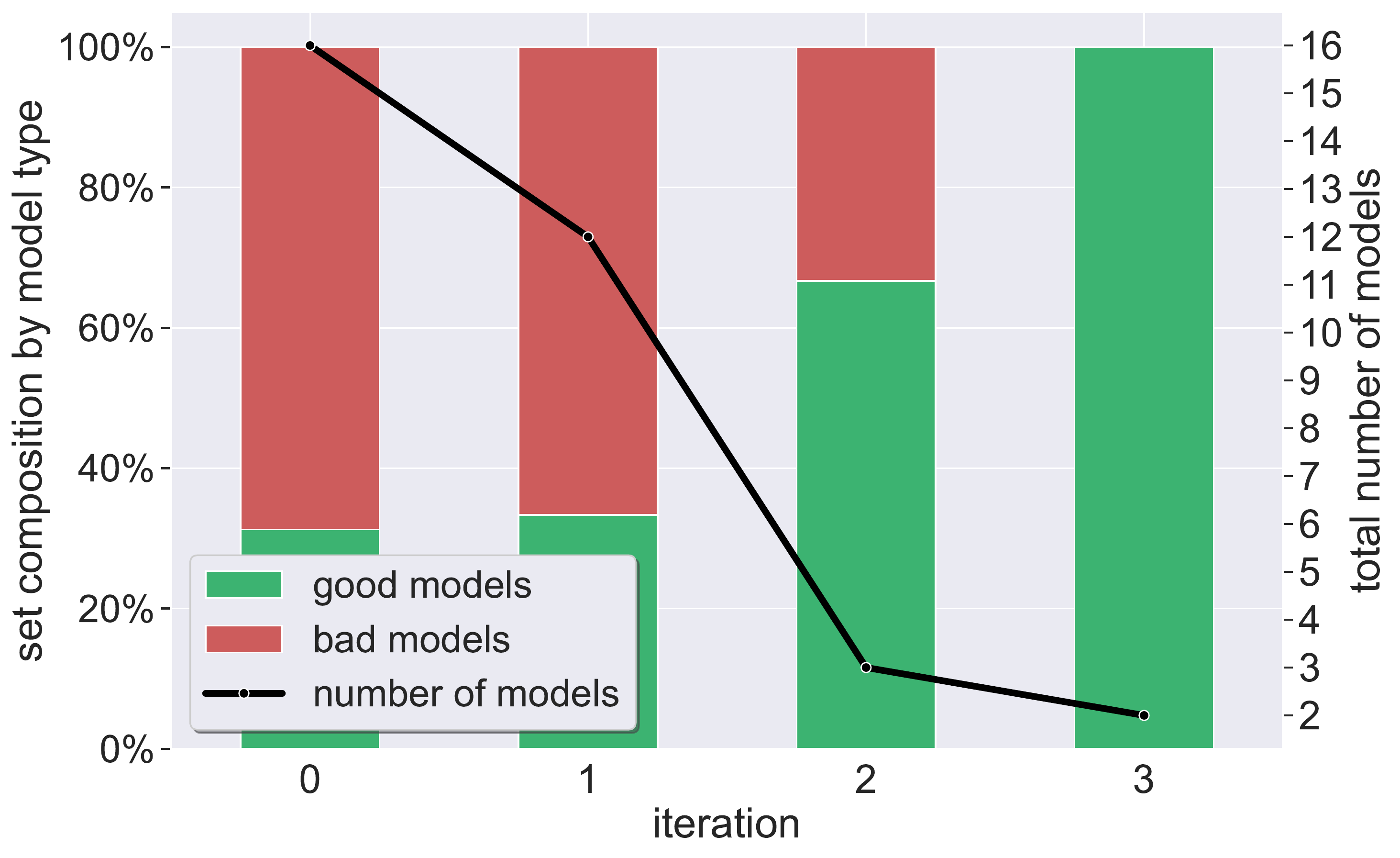}
         \caption{Remaining good and bad models ratio}
        \label{}
     \end{subfigure}
    \caption{Cartpole: results using the \conditionCombined filtering 
    criterion.}
    \label{fig:cartpole:CombinedResults}
\end{figure}
\FloatBarrier

\newpage

\section{Mountain Car: Supplementary Results}
\label{sec:appendix:MountainCarSupplementaryResults}

\subsection{The Mountain Car Benchmark}

Mountain Car is a classic, relatively simple RL benchmark. In this benchmark, a car
(agent) is placed in a valley between two hills (at $x\in[-1.2, 0.6]$), and needs to reach a flag on top of one of the hills. The state, $s=(x, v_{x})$ represents the car's location (along the x-axis) and velocity.
The agent's action (output) is the \emph{applied force}: a continuous
value indicating the magnitude and direction in which the agent wishes
to move. During training, the agent is incentivized to reach the flag (placed at the top of a valley, originally at $x=0.45$). For each time-step until the flag is reached, the agent receives a small, negative reward; if it reaches the flag, the agent is rewarded with a large positive reward. An episode terminates when the flag is reached, or when the
number of steps exceeds some predefined value ($300$ in our
experiments). Good and bad models are distinguished by an average reward threshold of \textbf{90}.

\begin{figure}
    \centering
    \captionsetup[subfigure]{justification=centering}
    \captionsetup{justification=centering}
    \begin{subfigure}[t]{0.49\linewidth}
         \includegraphics[width=\textwidth]{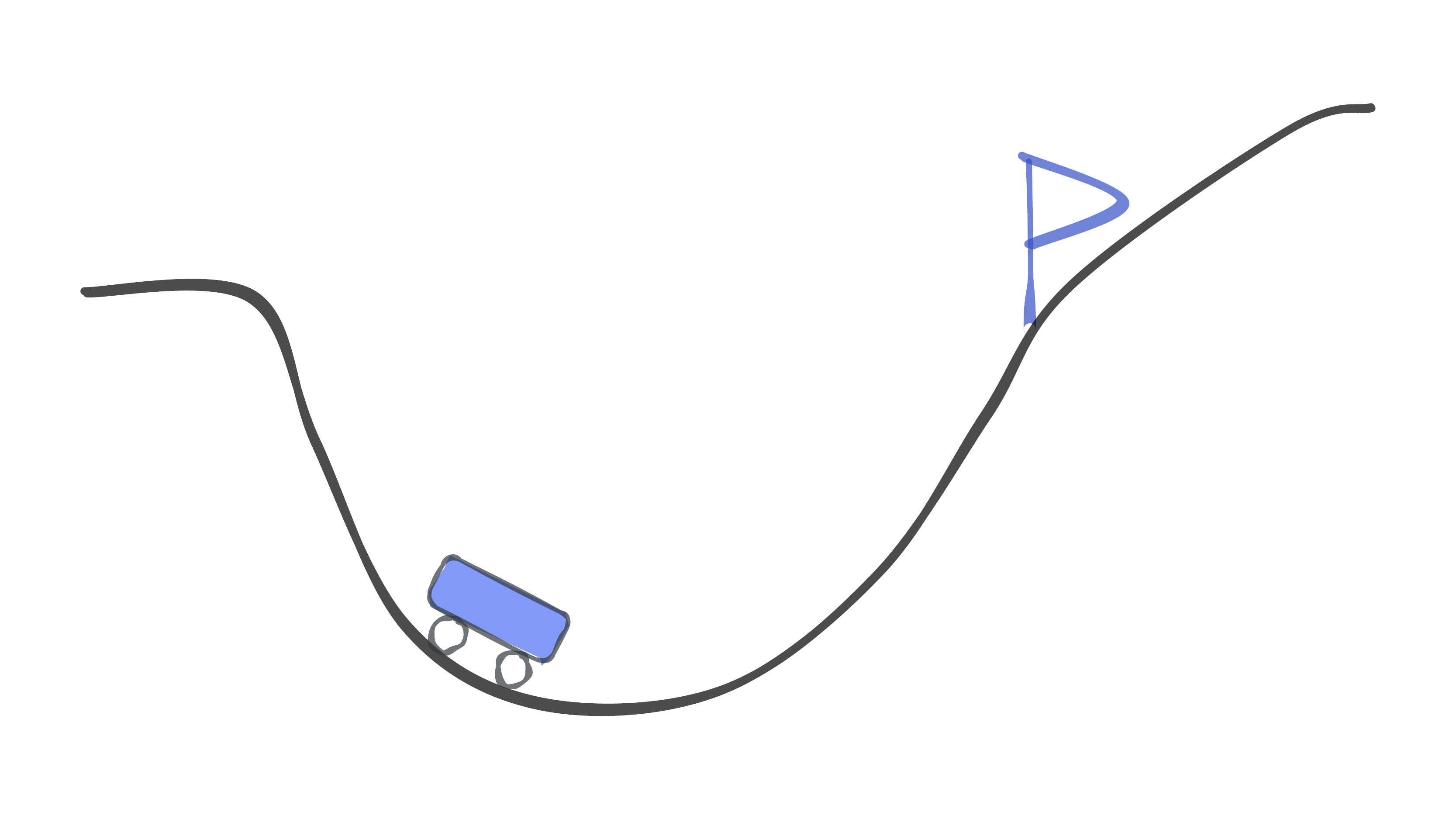}
             \caption{Training (in-distribution) setting: the agent's initial point is on the left side, and the goal is at a nearby point}
         \label{fig:MountainCarInDistribution:a}
     \end{subfigure}
     \hfill
     \begin{subfigure}[t]{0.49\linewidth}
         \includegraphics[width=\textwidth]{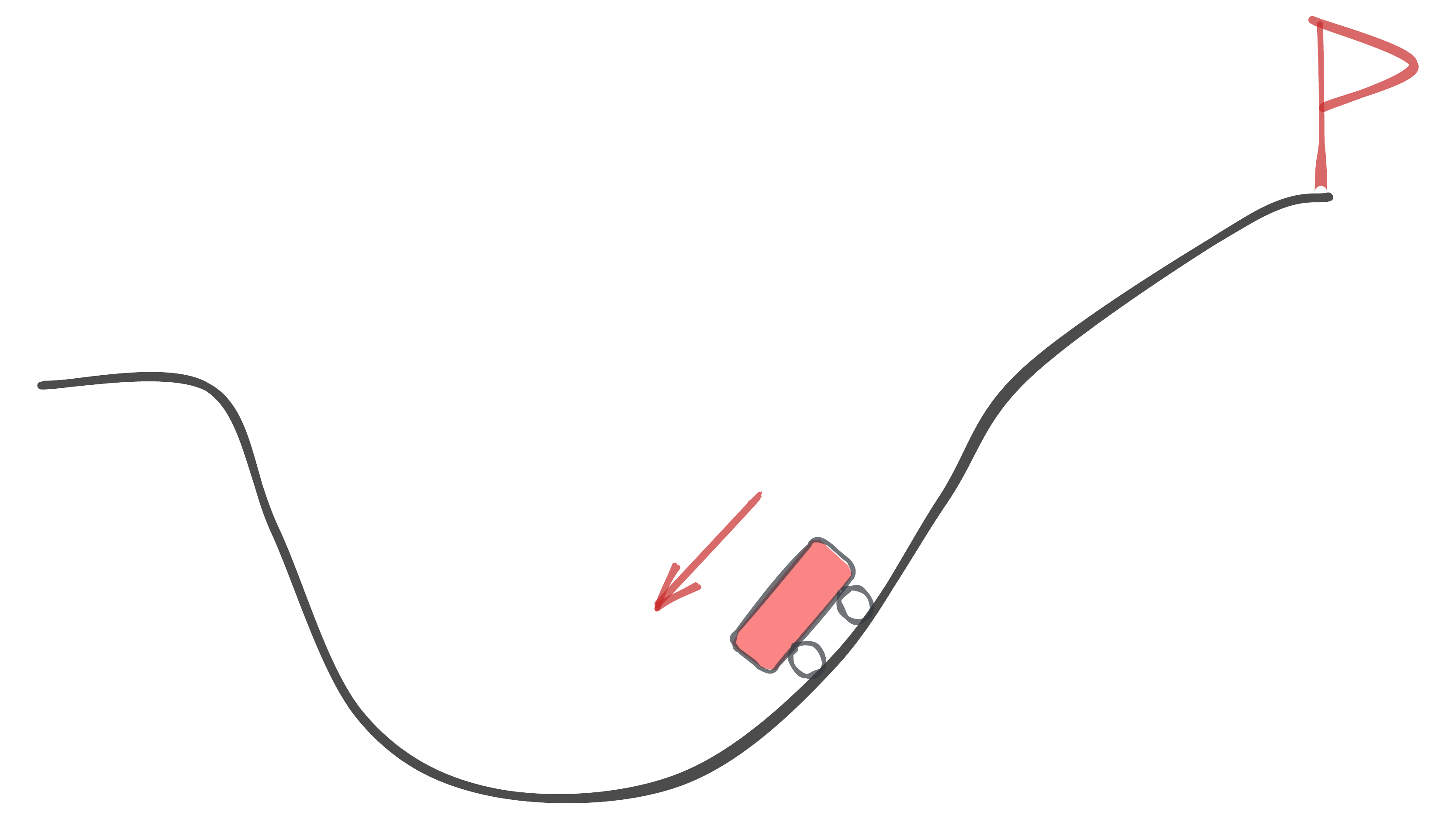}
         
         \caption{OOD setting: the agent's initial point is farther away, with 
         initial negative velocity; the goal is significantly farther up than 
         in training}
         \label{fig:MountainCarOod:b}
     \end{subfigure}
    
    \caption{Mountain Car: figure (a) depicts the setting on which the agents 
    were trained, and figure (b) depicts the harder, OOD 
    setting.}\label{fig:MountainCarInDistributionAndOod}
\end{figure}

During training (in-distribution), the car is initially placed on the \emph{left} side
of the valley's bottom, with a low, random velocity (see
Fig.~\ref{fig:MountainCarInDistribution:a}). We trained $k=16$ agents (denoted as
$\{1, 2, \ldots 16\}$), which all perform well (i.e., achieve an average reward higher than our
threshold) in-distribution. This evaluation was done for over $10,000$ 
episodes.  

According to the scenarios used by the training environment, we specified the (OOD) input domain by:
\begin{inparaenum}[(i)]
\item extending the x-axis, from $[-1.2, 0.6]$ to $[-2.4,
  0.9]$;
\item moving the flag further to the right, from $x=0.45$ to
  $x=0.9$; and 
\item setting the car's initial location further to the right of
  the valley's bottom, and with a large initial \textit{negative}
  velocity (to the left).
\end{inparaenum}
An illustration appears in Fig.~\ref{fig:MountainCarOod:b}.  These new
settings represent a novel state distribution, which causes the agents
to respond to states that they had not observed during training:
different locations, greater velocity, and different combinations of
location and velocity directions.

\begin{figure}
    \centering
    \subfloat[In-distribution \label{subfig:mountaincarRewards:inDist}]{\includegraphics[width=0.49\textwidth]{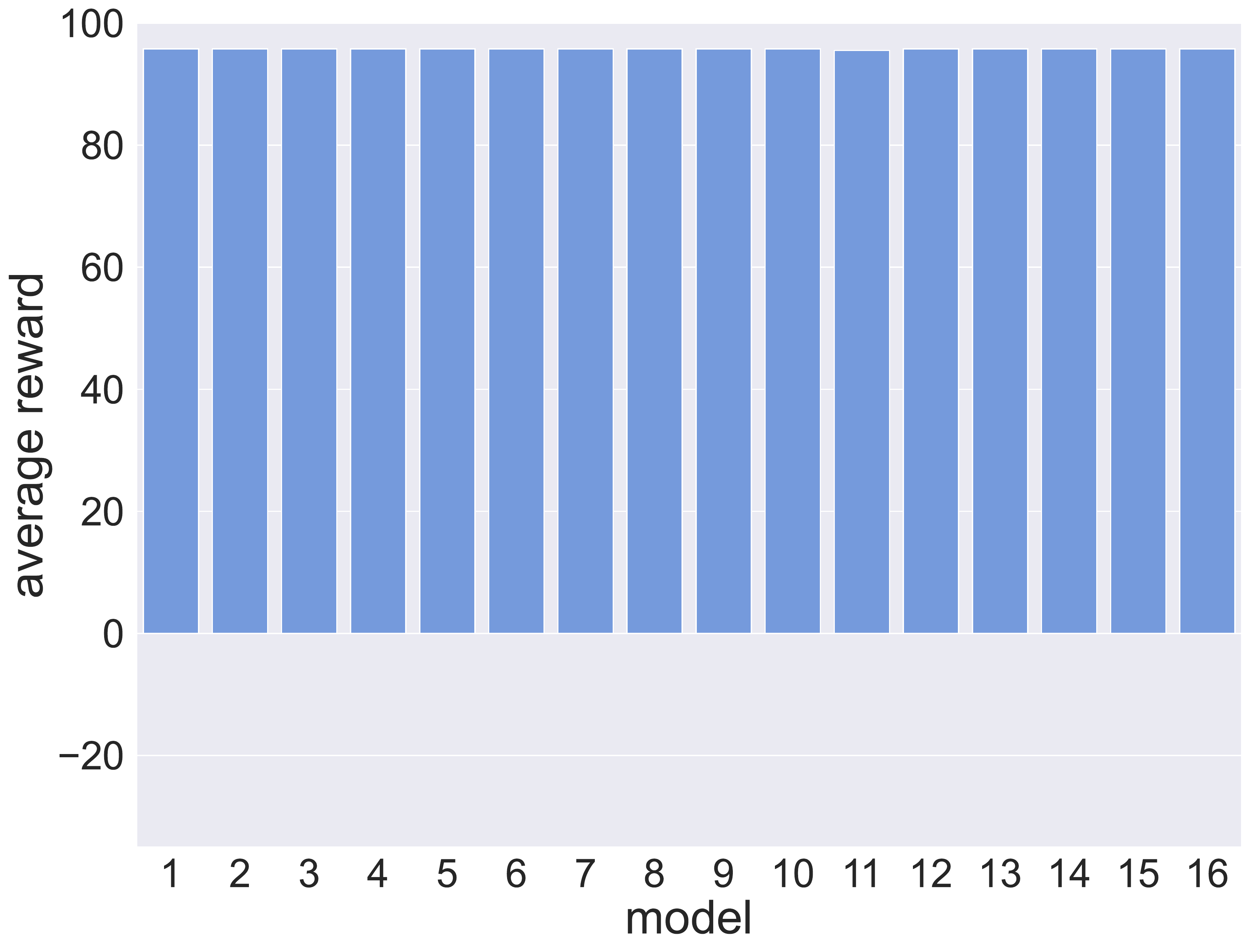}}
    \hfill
    \subfloat[OOD\label{subfig:mountaincarRewards:OOD}]{\includegraphics[width=0.49\textwidth]{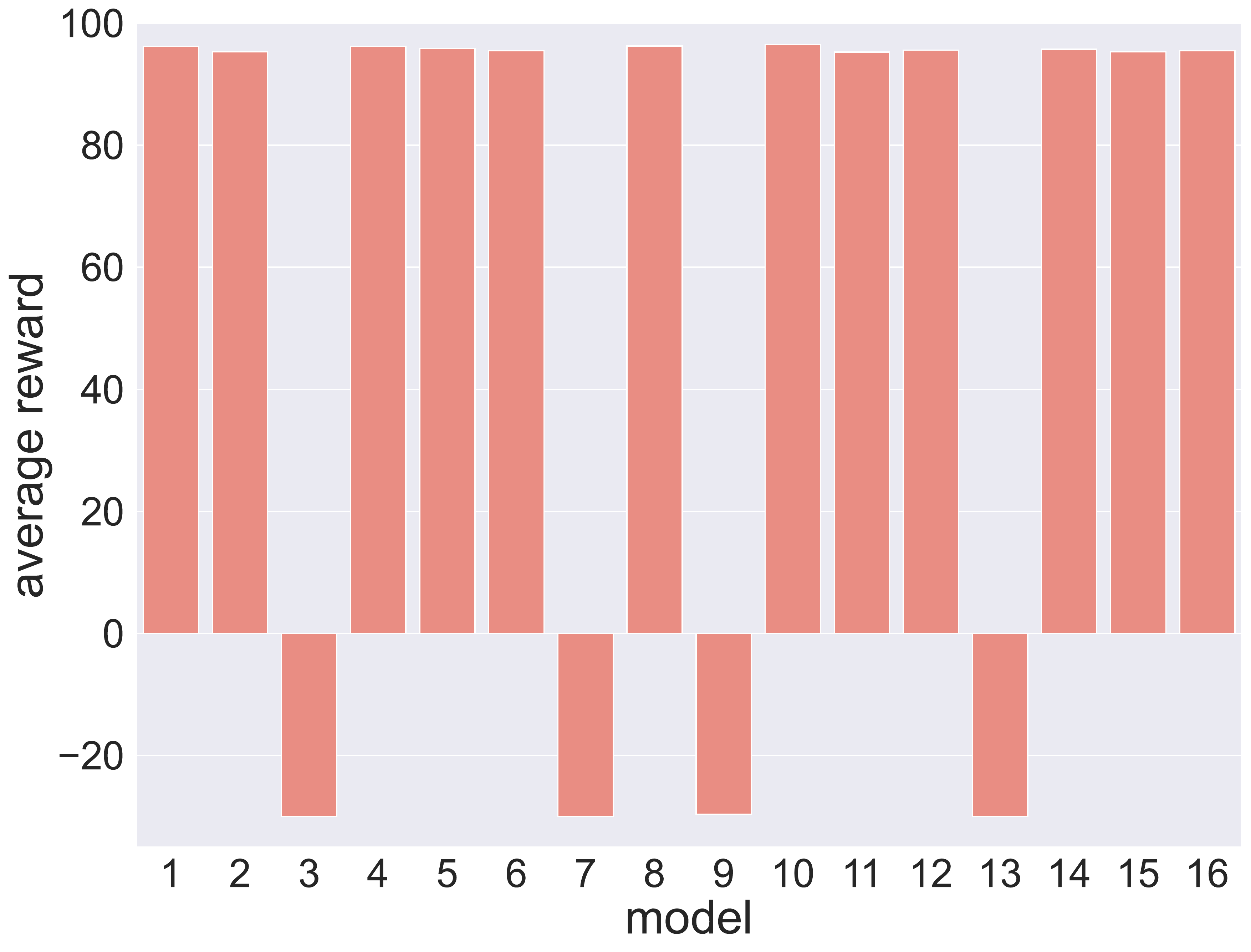}}\\
    
    \caption{Mountain Car: the models' average rewards in different 
    distributions.}\label{fig:mountaincarRewards}
\end{figure}

Out of the $k=16$ models that performed well in-distribution, $4$ models
failed (i.e., did not reach the flag, ending their episodes with a negative
average reward) in the OOD scenario, while the remaining $12$
succeeded, i.e., reached a high average reward when simulated on the OOD data (see Fig.~\ref{fig:mountaincarRewards}).  
The large ratio
of successful models is not surprising, as Mountain Car is a relatively easy benchmark.

To evaluate our algorithm, we ran it on these models, and the aforementioned (OOD) input domain, and checked whether it removed the
models that (although successful in-distribution) fail in the new, harder,
setting. Indeed, our method was able to filter out all unsuccessful
models, leaving only a subset of $5$ models ($\{2,4,8,10,15\}$), all of which perform well in the OOD scenario.



\begin{figure}[ht]
    \centering
    \captionsetup[subfigure]{justification=centering}
    \captionsetup{justification=centering} 
    \begin{subfigure}[t]{0.49\linewidth}
        \includegraphics[width=\textwidth]{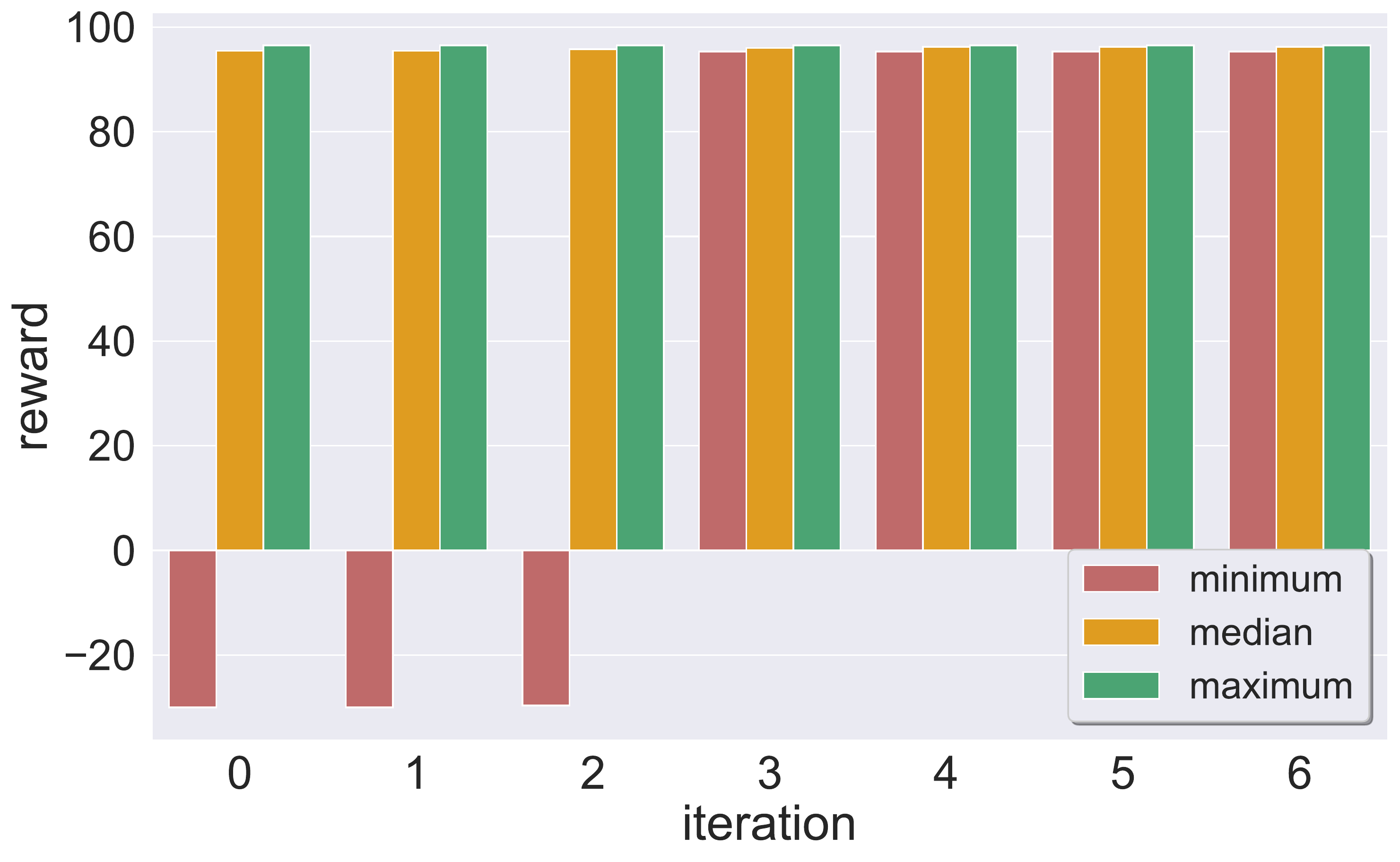}
         \caption{Rewards statistics}
        \label{fig:mountaincar:PercentileMinMaxRewards}
     \end{subfigure}
     \hfill
     \begin{subfigure}[t]{0.49\linewidth}
         \centering
    \includegraphics[width=\textwidth]{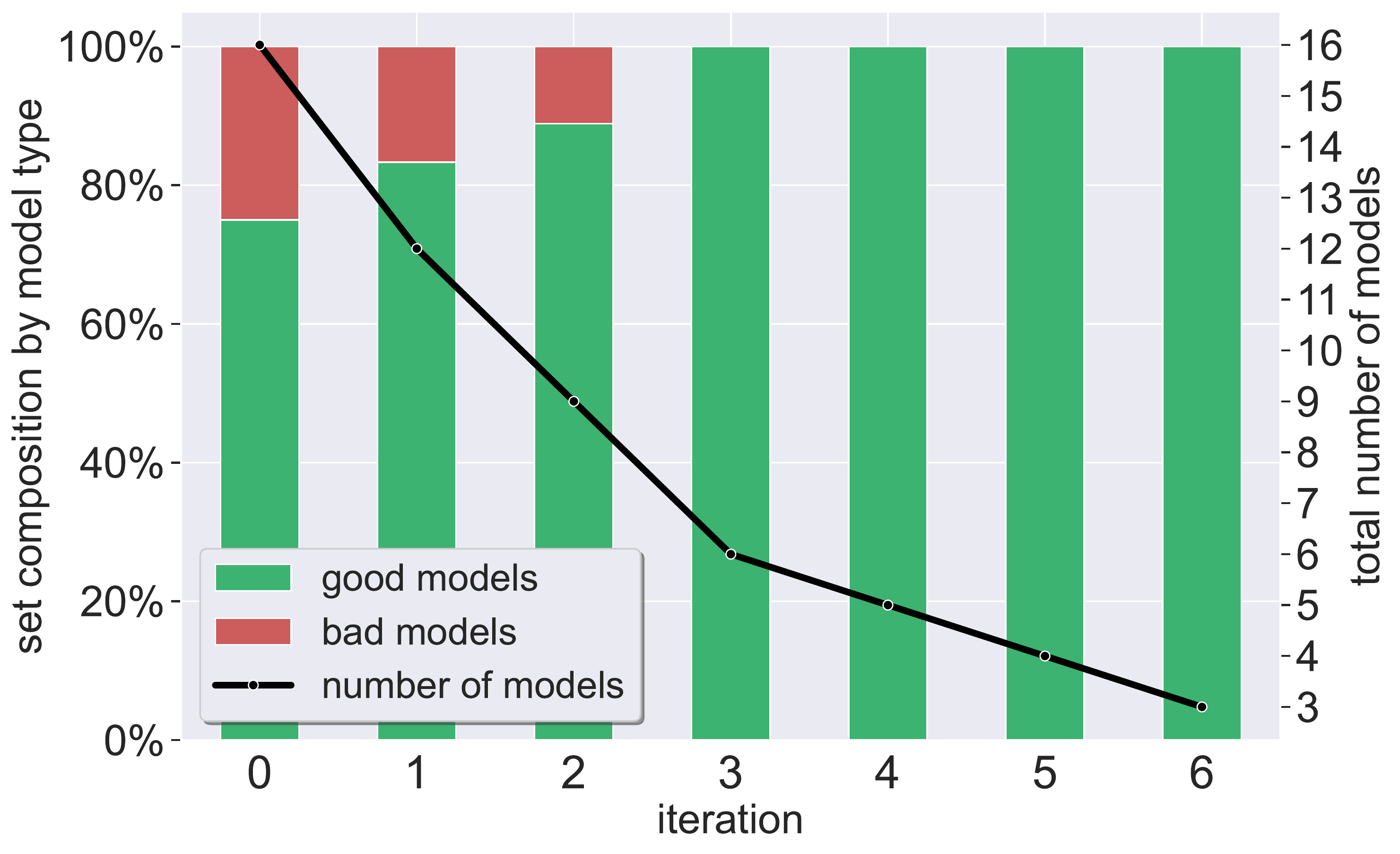}
         \caption{Remaining good and bad models ratio}
        \label{fig:mountaincar:goodBadModelsPercentages}
     \end{subfigure}
    \caption{Mountain Car: results using the \conditionPercentile filtering 
    criterion.}
    \label{fig:mountaincar:PrecentileCritResults}
\end{figure}

\begin{figure}[!h]
    \centering
    \captionsetup[subfigure]{justification=centering}
    \captionsetup{justification=centering} 
    \begin{subfigure}[t]{0.49\linewidth}
         \centering
    \includegraphics[width=\textwidth]{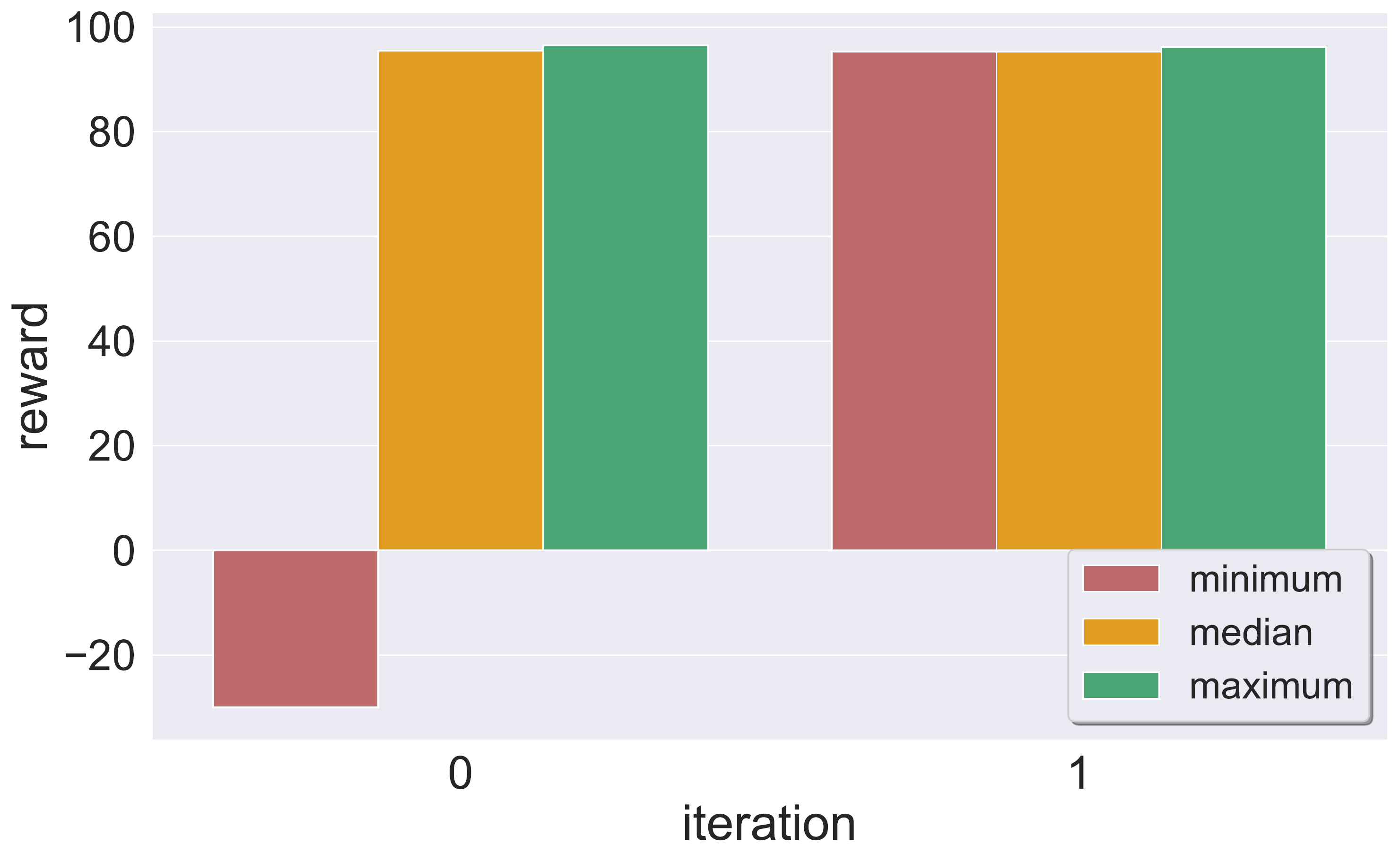}
         \caption{Rewards statistics}
        \label{}
     \end{subfigure}
     \hfill
     \begin{subfigure}[t]{0.49\linewidth}
        \includegraphics[width=\textwidth]{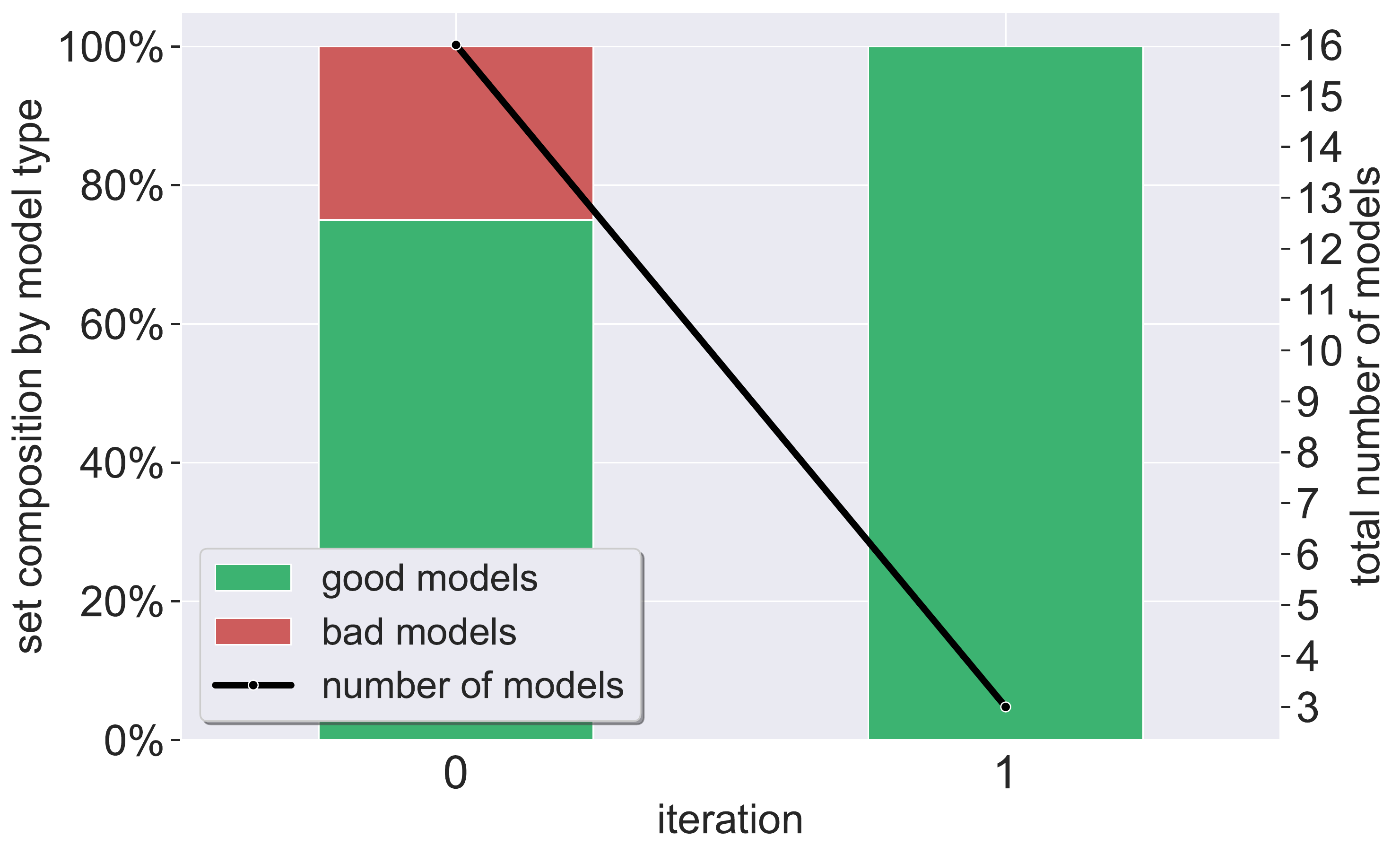}
         \caption{Remaining good and bad models ratio}
        \label{}
     \end{subfigure}
    \caption{Mountain Car: results using the \maxAgg filtering criterion.}
    \label{fig:mountaincar:MaxCritResults}
\end{figure}

\begin{figure}[!h]
    \centering
    \captionsetup[subfigure]{justification=centering}
    \captionsetup{justification=centering} 
    \begin{subfigure}[t]{0.49\linewidth}
         \centering
    \includegraphics[width=\textwidth]{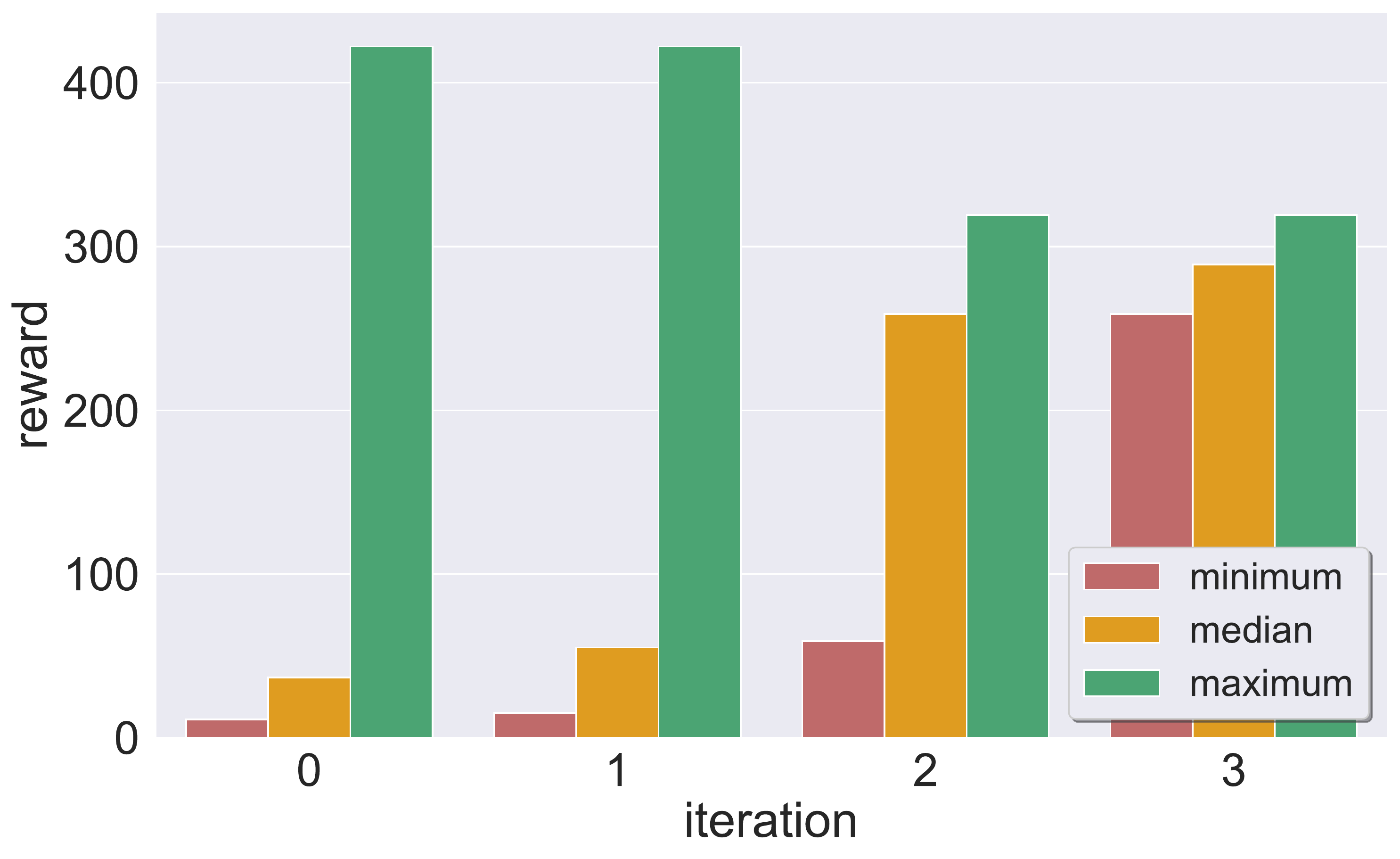}
         \caption{Rewards statistics}
        \label{}
     \end{subfigure}
     \hfill
     \begin{subfigure}[t]{0.49\linewidth}
        \includegraphics[width=\textwidth]{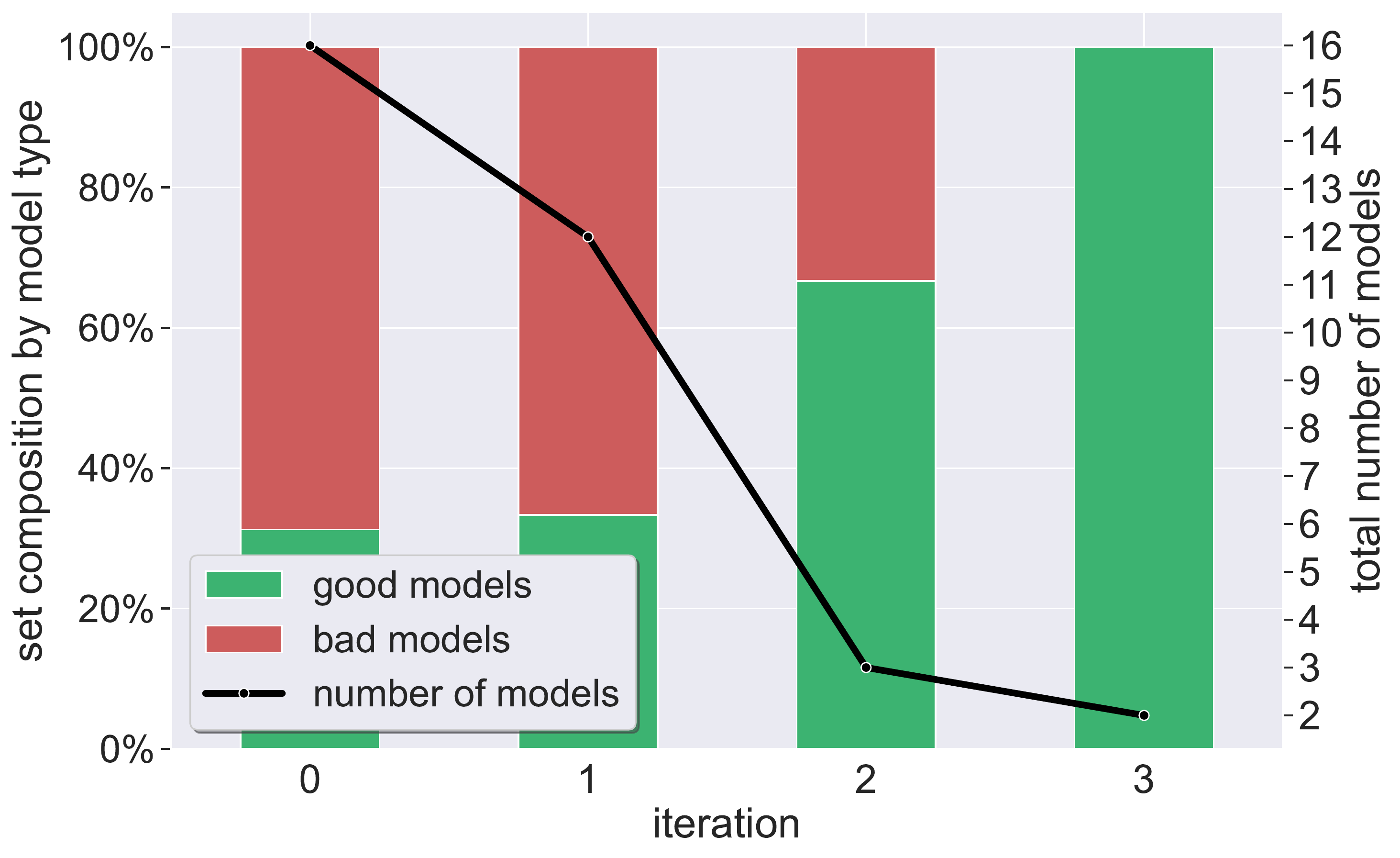}
         \caption{Remaining good and bad models ratio}
        \label{}
     \end{subfigure}
    \caption{Mountain Car: results using the \conditionCombined filtering 
    criterion.}
    \label{fig:mountaincar:CombinedCritResults}
\end{figure}

We also note that our algorithm is robust to various hyperparameters, as demonstrated in Fig.~\ref{fig:mountaincar:PrecentileCritResults}, Fig.~\ref{fig:mountaincar:MaxCritResults} and Fig.~\ref{fig:mountaincar:CombinedCritResults} which depict the results of each iteration of our algorithm, when applied with various filtering criteria (elaborated in Appendix~\ref{sec:appendix:AlgorithmAdditionalInformation}).

\FloatBarrier


\newpage
\subsection{Additional Filtering Criteria}

\FloatBarrier


\subsection{Combinatorial Experiments}

\begin{figure}[!h]
    \centering
\captionsetup{justification=centering}
    \includegraphics[width=0.6\textwidth]{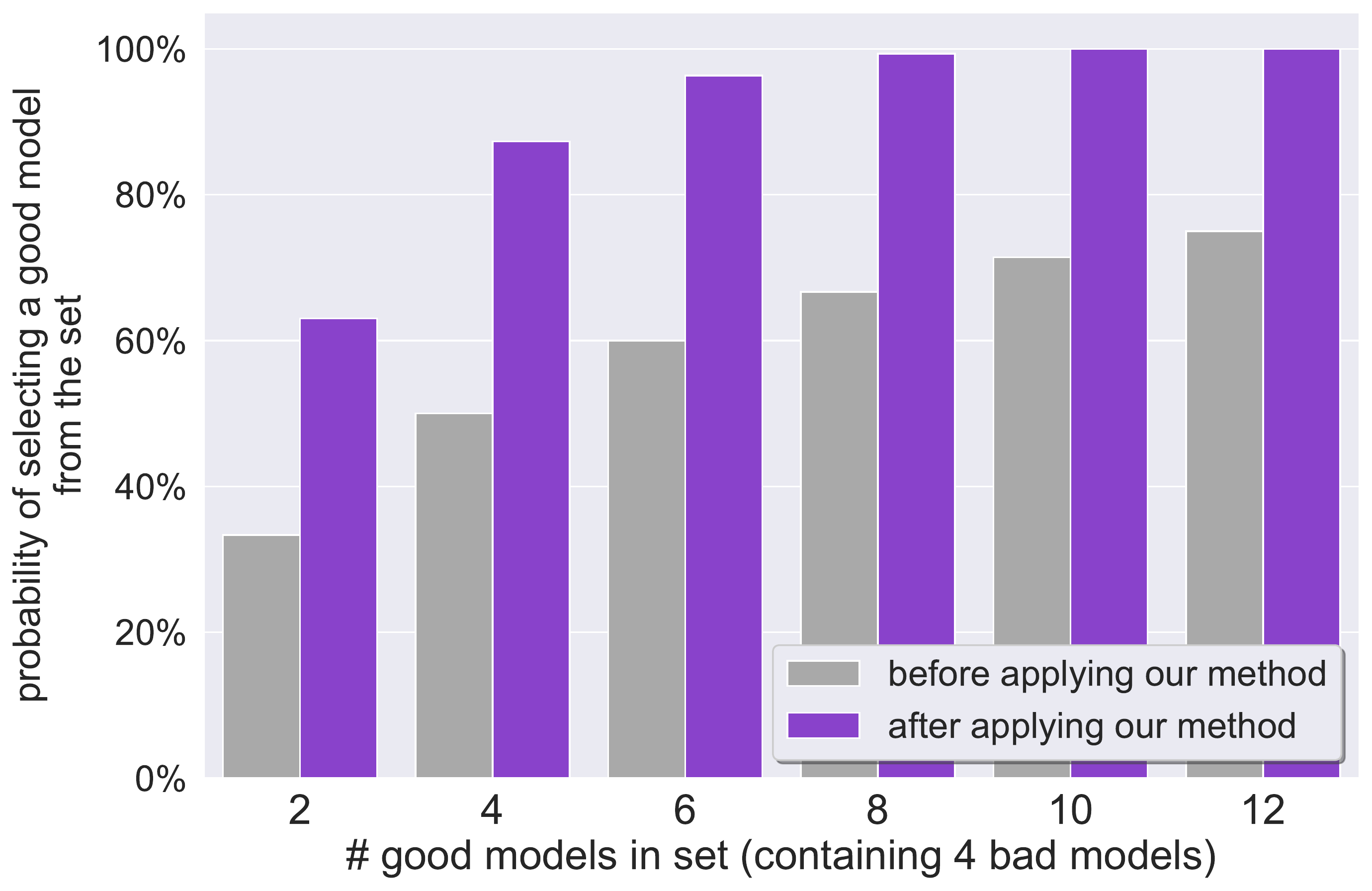}
    
     \caption{Mountain Car: our algorithm effectively increases the probability 
     to choose a good model, due to effective filtering. The plot corresponds 
     to Table~\ref{table:results:MountainCarCombinatorialResultsTableMaxConditionMaxHyperparameter2}.}
    \label{fig:mountaincarProbabilitiesMaxConditionHyperparameter2}
\end{figure}

Due to the original bias of the initial set of candidates, in which $12$ out of the original $16$ models are good in the OOD setting, we set out to validate that the fact that our algorithm succeeded in returning solely good models is indeed due to its correctness, and not due to the inner bias among the set of models, to contain good models. In our experiments (summarized below) we artificially generated new sets of models in which the ratio of good models is deliberately lower than in the original set. We then reran our algorithm on all possible combinations of the initial subsets, and calculated (for each subset) the probability to select a good model in this new setting, from the models surviving our filtering process. As we show, our method significantly improves the chances of selecting a good model \emph{even when these are a minority in the original set}. For example, the leftmost column of Fig.~\ref{fig:mountaincarProbabilitiesMaxConditionHyperparameter2} shows that over sets consisting of $4$ bad models and only $2$ good ones, the probability of selecting a good model after running our algorithm is over $60\%$ (!) --- almost double the probability of randomly selecting a good model from the original set before running our algorithm. These results were consistent across multiple subset sizes, and with various filtering criteria.
We note that for the calculations demonstrating the chance to select a good model, we assume random selection from a subset of models: \emph{before} applying our algorithm, the subset is the original set of models; and \emph{after} our algorithm is applied --- the subset is updated based on the result of our filtering procedure. The probability is computed based on the number of combinations of bad models surviving the filtering process, and their ratio relative to all the models returned in those cases (we assume uniform probability, per subset).

\begin{table}[ht]
	\centering
        \captionsetup{justification=centering}     
	\begin{tabular}
    { |P{0.07\linewidth}:P{0.07\linewidth}| P{0.2\linewidth}|P{0.18\linewidth}:P{0.18\linewidth}|P{0.15\linewidth}:P{0.15\linewidth}|}
		\hline
		\multicolumn{2}{|P{0.14\linewidth}|}{\texttt{\textbf{COMPOSITION}}} & 
		\multicolumn{1}{P{0.2\linewidth}|}{\texttt{\textbf{total $\#$ experiments}}} & 
		\multicolumn{2}{ P{0.36\linewidth}|}{\texttt{\textbf{$\#$ experiments with}}} & 
		\multicolumn{2}{ P{0.3\linewidth}|}{\texttt{\textbf{probability to choose 
		\textcolor{forestGreen}{good} models}}} \\ 
		\hline
		\textcolor{forestGreen}{good} & \textcolor{red}{bad} & 
		{\shortstack{subgroup \\combinations\\of 
		\textcolor{forestGreen}{good} models}} & 
		\shortstack{\textbf{all} \\surviving\\ models 
		\textcolor{forestGreen}{good}} & 
		\shortstack{\textbf{some}\\surviving\\ models \textcolor{red}{bad}} & 
		naive 
		& our method \\ 
		\hline
		
		\textcolor{forestGreen}{12} & 
		\textcolor{red}{4} & 
		${12 \choose\textcolor{forestGreen}{12}} = 1$ & 
		1 & 
		0 & 
        75 \%
		 &
		100\% \\
		
		\hline
		\textcolor{forestGreen}{10} &
		\textcolor{red}{4} & 
		${12 \choose \textcolor{forestGreen}{10}} = 66$ & 
		65 & 
		1 & 
		71.43 \%
        & 
		99.49 \% 
        \\
		 
		\hline
		\textcolor{forestGreen}{8} & 
		\textcolor{red}{4} & 
		${12 \choose \textcolor{forestGreen}{8}} =  495$ & 
		423 & 
		72 & 
		66.67 \% &
       95.15 \%
		\\

		\hline
		\textcolor{forestGreen}{6} & 
		\textcolor{red}{4} & 
		${12 \choose \textcolor{forestGreen}{6}} = 924$ & 
		549 & 
		375 & 
		60 \% &
		86.26 \%  
		\\ 
		
		\hline
		\textcolor{forestGreen}{4} 
		& 
		\textcolor{red}{4} & 
		${12 \choose \textcolor{forestGreen}{4}} =  495$ & 
		123 & 372 & 
		50 \%
		 & 
       74.01 \%
		\\

		\hline
		\textcolor{forestGreen}{2} & 
		\textcolor{red}{4} & 
		${12 \choose \textcolor{forestGreen}{2}} = 66$ & 
		66 & 
		0 & 
		33.33 \%
		 & 
        54.04 \%   
        \\ 
		
		\hline
	
    \end{tabular}
	
	\vspace{3mm}
    \caption{Results summary of the combinatorial Mountain Car experiment, 
    using the \conditionPercentile filtering criterion.}
    \label{table:results:MountainCarCombinatorialResultsTablePercentileCondition}
\end{table}
\FloatBarrier


\begin{table}[ht]
	\centering
        \captionsetup{justification=centering} 
	\begin{tabular}
    { |P{0.07\linewidth}:P{0.07\linewidth}| P{0.2\linewidth}|P{0.18\linewidth}:P{0.18\linewidth}|P{0.15\linewidth}:P{0.15\linewidth}|}
		\hline
		\multicolumn{2}{|P{0.14\linewidth}|}{\texttt{\textbf{COMPOSITION}}} & 
		\multicolumn{1}{P{0.2\linewidth}|}{\texttt{\textbf{total $\#$ experiments}}} & 
		\multicolumn{2}{ P{0.36\linewidth}|}{\texttt{\textbf{$\#$ experiments with}}} & 
		\multicolumn{2}{ P{0.3\linewidth}|}{\texttt{\textbf{probability to choose 
		\textcolor{forestGreen}{good} models}}} \\ 
		\hline
		\textcolor{forestGreen}{good} & \textcolor{red}{bad} & 
		{\shortstack{subgroup \\combinations\\of 
		\textcolor{forestGreen}{good} models}} & 
		\shortstack{\textbf{all} \\surviving\\ models 
		\textcolor{forestGreen}{good}} & 
		\shortstack{\textbf{some}\\surviving\\ models \textcolor{red}{bad}} & 
		naive 
		& our method \\ 
		\hline

		\textcolor{forestGreen}{12} & 
		\textcolor{red}{4} & 
		${12 \choose\textcolor{forestGreen}{12}} = 1$ & 
		1 & 
		0 & 
        75 \%
		 &
		100\% \\
		
		\hline
		\textcolor{forestGreen}{10} &
		\textcolor{red}{4} & 
		${12 \choose \textcolor{forestGreen}{10}} = 66$ & 
		66 & 
		0 & 
		71.43 \%
            & 
		100 \% 
        
        \\
		 
		\hline
		\textcolor{forestGreen}{8} & 
		\textcolor{red}{4} & 
		${12 \choose \textcolor{forestGreen}{8}} =  495$ & 
		486 & 
		9 & 
		66.67 \% &
       99.34 \%
		\\

		\hline
		\textcolor{forestGreen}{6} & 
		\textcolor{red}{4} & 
		${12 \choose \textcolor{forestGreen}{6}} = 924$ & 
		844 & 
		80 & 
		60 \% &
		96.33 \%  
		\\ 
		
		\hline
		\textcolor{forestGreen}{4} 
		& 
		\textcolor{red}{4} & 
		${12 \choose \textcolor{forestGreen}{4}} =  495$ & 
		375 & 120 & 
		50 \%
		 & 
       87.32 \%
		\\

		\hline
		\textcolor{forestGreen}{2} & 
		\textcolor{red}{4} & 
		${12 \choose \textcolor{forestGreen}{2}} = 66$ & 
		31 & 
		35 & 
		33.33 \%
		 & 
        63.03 \%   
        \\ 
		
		\hline
	\end{tabular}
	
	\vspace{3mm}
    \caption{Results summary of the combinatorial Mountain Car experiment, 
    using the \conditionMax filtering criterion.}
    \label{table:results:MountainCarCombinatorialResultsTableMaxConditionMaxHyperparameter2}
\end{table}

\begin{table}[ht]
	\centering
 \captionsetup[subfigure]{justification=centering}
    \captionsetup{justification=centering} 
	\begin{tabular}
    { |P{0.07\linewidth}:P{0.07\linewidth}| P{0.2\linewidth}|P{0.18\linewidth}:P{0.18\linewidth}|P{0.15\linewidth}:P{0.15\linewidth}|}
		\hline
		\multicolumn{2}{|P{0.14\linewidth}|}{\texttt{\textbf{COMPOSITION}}} & 
		\multicolumn{1}{P{0.2\linewidth}|}{\texttt{\textbf{total $\#$ experiments}}} & 
		\multicolumn{2}{ P{0.36\linewidth}|}{\texttt{\textbf{$\#$ experiments with}}} & 
		\multicolumn{2}{ P{0.3\linewidth}|}{\texttt{\textbf{probability to choose 
		\textcolor{forestGreen}{good} models}}} \\ 
		\hline
		\textcolor{forestGreen}{good} & \textcolor{red}{bad} & 
		{\shortstack{subgroup \\combinations\\of 
		\textcolor{forestGreen}{good} models}} & 
		\shortstack{\textbf{all} \\surviving\\ models 
		\textcolor{forestGreen}{good}} & 
		\shortstack{\textbf{some}\\surviving\\ models \textcolor{red}{bad}} & 
		naive 
		& our method \\ 
		\hline
		\textcolor{forestGreen}{12} & 
		\textcolor{red}{4} & 
		${12 \choose\textcolor{forestGreen}{12}} = 1$ & 
		1 & 
		0 & 
        75 \%
		 &
		100\% \\
		
		\hline
		\textcolor{forestGreen}{10} &
		\textcolor{red}{4} & 
		${12 \choose \textcolor{forestGreen}{10}} = 66$ & 
		66 & 
		0 & 
		71.43 \%
        & 
		100 \% \\
		 
		\hline
		\textcolor{forestGreen}{8} & 
		\textcolor{red}{4} & 
		${12 \choose \textcolor{forestGreen}{8}} =  495$ & 
		481 & 
		12 & 
		66.67 \% &
        99.26 \%
		\\

		\hline
		\textcolor{forestGreen}{6} & 
		\textcolor{red}{4} & 
		${12 \choose \textcolor{forestGreen}{6}} = 924$ & 
		842 & 
		82 & 
		60 \% &
		96.74 \%  
		\\ 
		
		\hline
		\textcolor{forestGreen}{4} 
		& 
		\textcolor{red}{4} & 
		${12 \choose \textcolor{forestGreen}{4}} =  495$ & 
		372 & 123 & 
		50 \%
		 & 
        88.09 \%
		\\

		\hline
		\textcolor{forestGreen}{2} & 
		\textcolor{red}{4} & 
		${12 \choose \textcolor{forestGreen}{2}} = 66$ & 
		31 & 
		35 & 
		33.33 \%
		 & 
        63.36 \%   
        \\ 
		
		\hline
	\end{tabular}
	
	\vspace{3mm}
    \caption{Results summary of the combinatorial Mountain Car experiment, 
    using the \conditionCombined filtering criterion.}
    \label{table:results:MountainCarCombinatorialResultsTableCombinedConditionMaxHyperparameter2}
\end{table}
\FloatBarrier

\clearpage
	
\section{Aurora: Supplementary Results}
\label{sec:appendix:AuroraSupplementaryResults}
\subsection{Additional Information}

\begin{figure}[ht]
    \centering
    \captionsetup{justification=centering}
    \subfloat[In-distribution \label{subfig:auroraRewards:inDist}]{\includegraphics[width=0.49\textwidth]{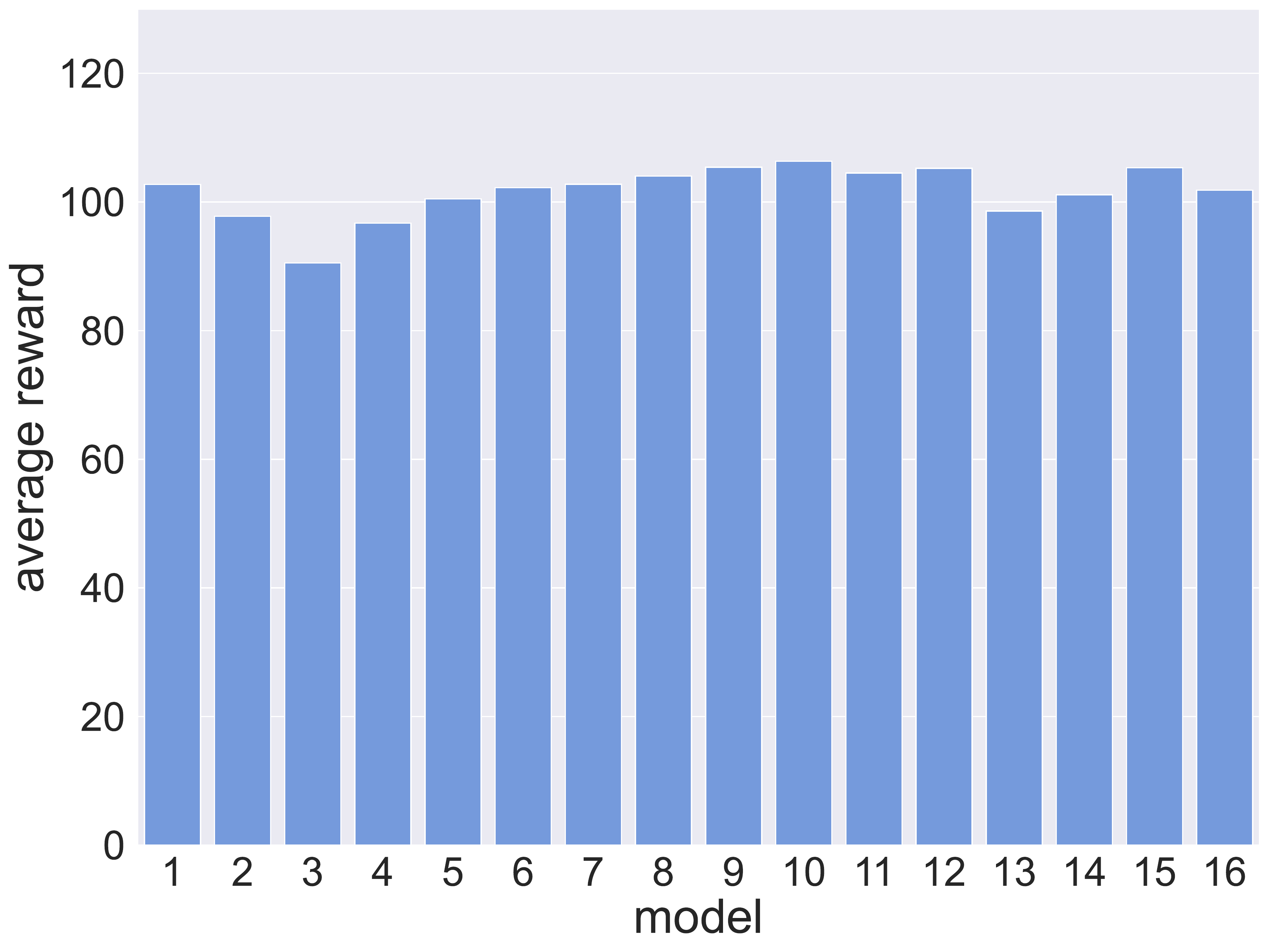}}
    \hfill
    \subfloat[OOD\label{subfig:auroraRewards:OOD}]{\includegraphics[width=0.49\textwidth]{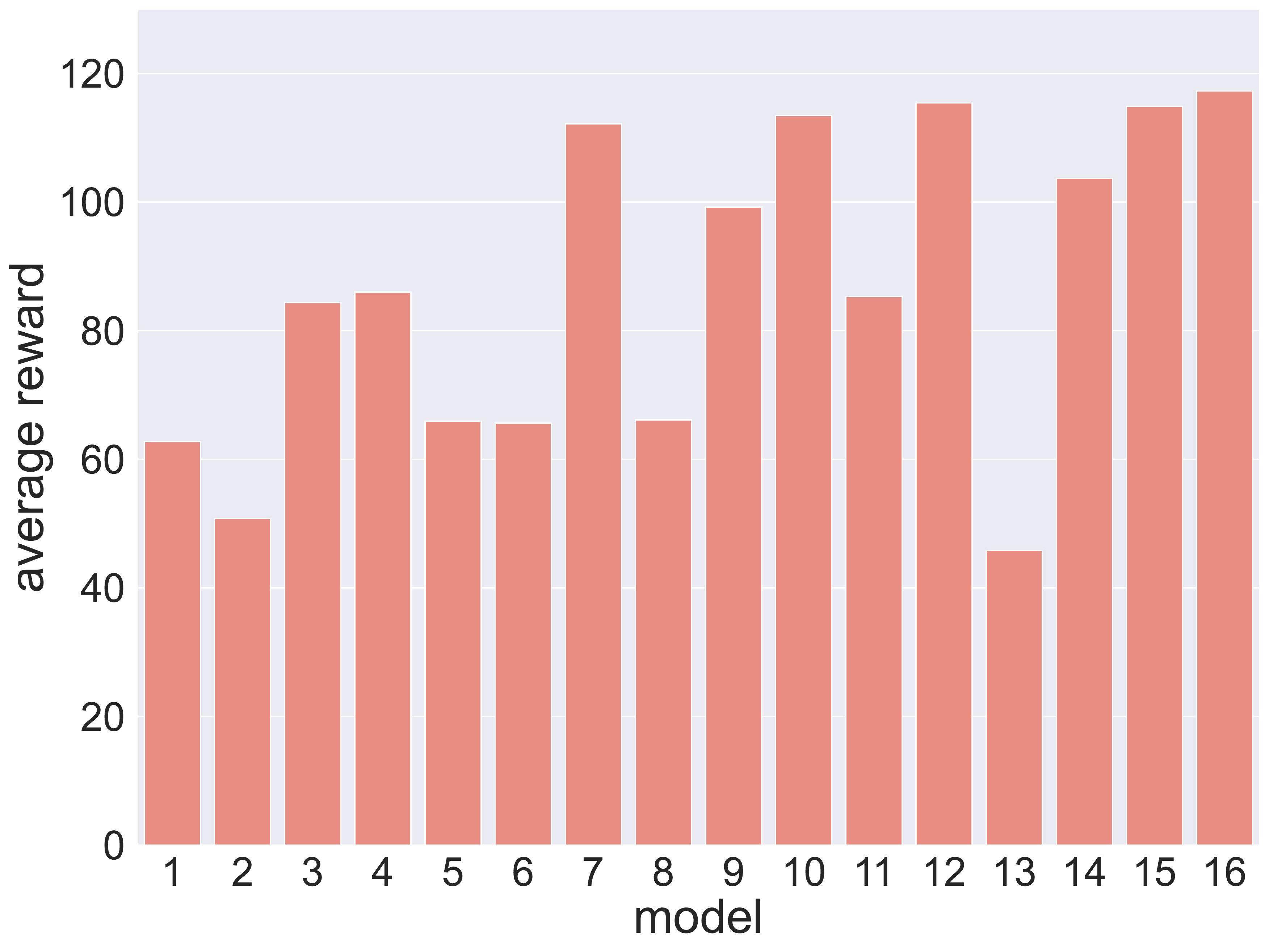}}\\
    
    \caption{Aurora Experiment~\ref{exp:auroraShort}: the models' average 
    rewards when simulated on different distributions.}\label{fig:auroraRewards}
\end{figure}

\begin{enumerate}
    \item A more detailed explanation of Aurora's input statistics:
    \begin{inparaenum}[(i)]
      \item \textit{Latency Gradient}: a derivative of latency (packet delays) 
      over the recent MI (``monitor interval'');
    \item \textit{Latency Ratio}: the ratio between the average latency in the current MI to the
      minimum latency previously observed; and
    \item \textit{Sending Ratio}: the ratio between the number of packets sent to the number of acknowledged packets over the recent MI.
    \end{inparaenum} As mentioned, these metrics indicate the link's congestion level.

    \item For all our experiments on this benchmark, we defined ``good'' models as models that achieved an average reward greater than a threshold of \textbf{99}; ``bad'' models are models that achieved a reward lower than this threshold.
    
    \item \emph{In-distribution}, the average reward is not necessarily correlated with the average reward \emph{OOD}. For example, in Exp.~\ref{exp:auroraShort} with the short episodes during training (see Fig.~\ref{fig:auroraRewards}):

    \begin{enumerate}
        \item In-distribution, model $\{4\}$ achieved a lower reward than models $\{2\}$ and $\{5\}$, but a higher reward OOD.

        \item In-distribution, model $\{16\}$ achieved a lower reward than model $\{15\}$, but a higher reward OOD.
    \end{enumerate}
\end{enumerate}

\begin{figure}
    \centering
    \captionsetup[subfigure]{justification=centering}
    \captionsetup{justification=centering}
    \begin{subfigure}[t]{0.48\linewidth}
         \includegraphics[width=\textwidth]{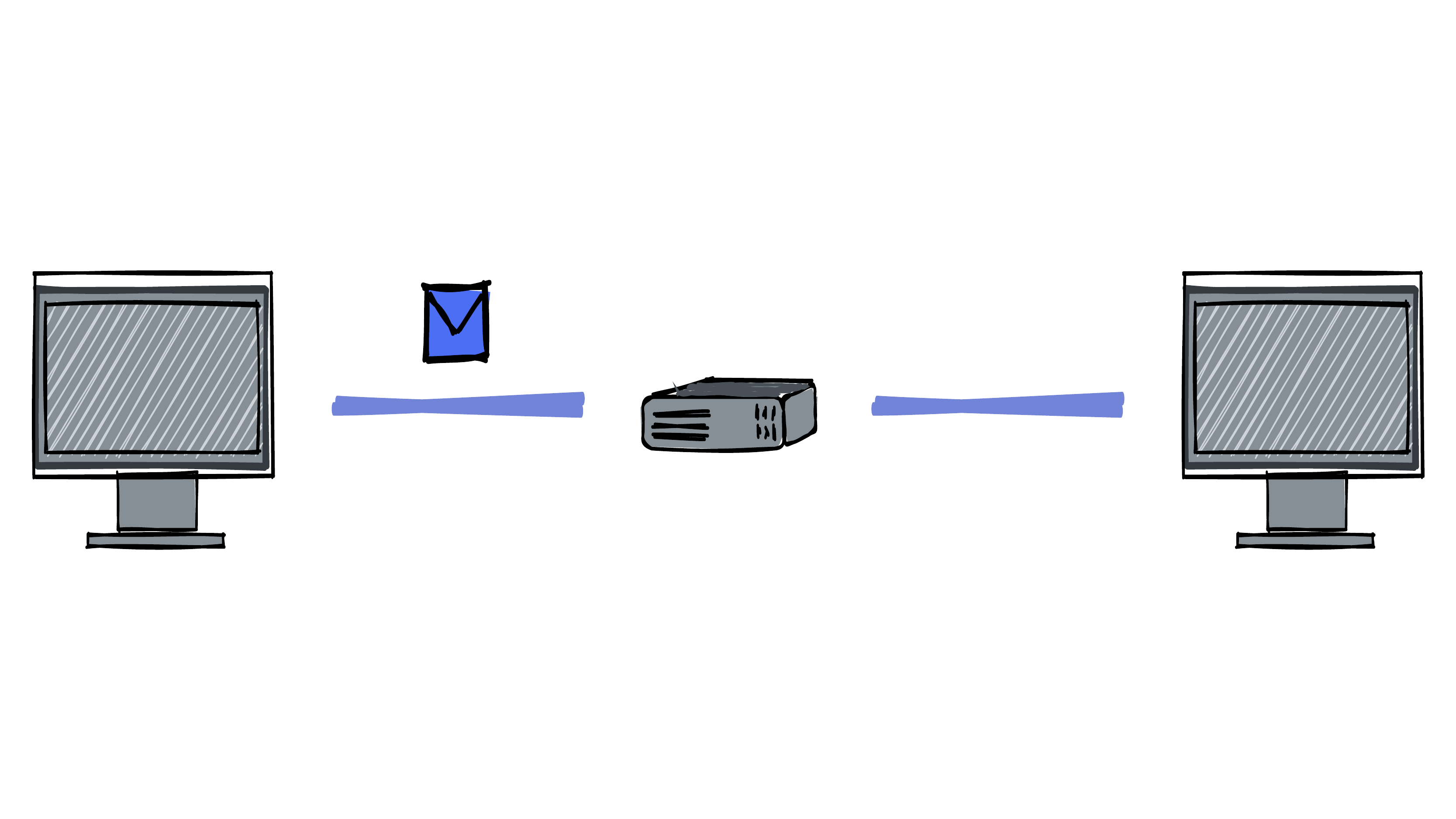}
         \caption{In-distribution setting, in which the agent was trained, with low packet congestion}
         \label{fig:AuroraInDistribution:a}
     \end{subfigure}
     \hfill
     \begin{subfigure}[t]{0.48\linewidth}
         \includegraphics[width=\textwidth]{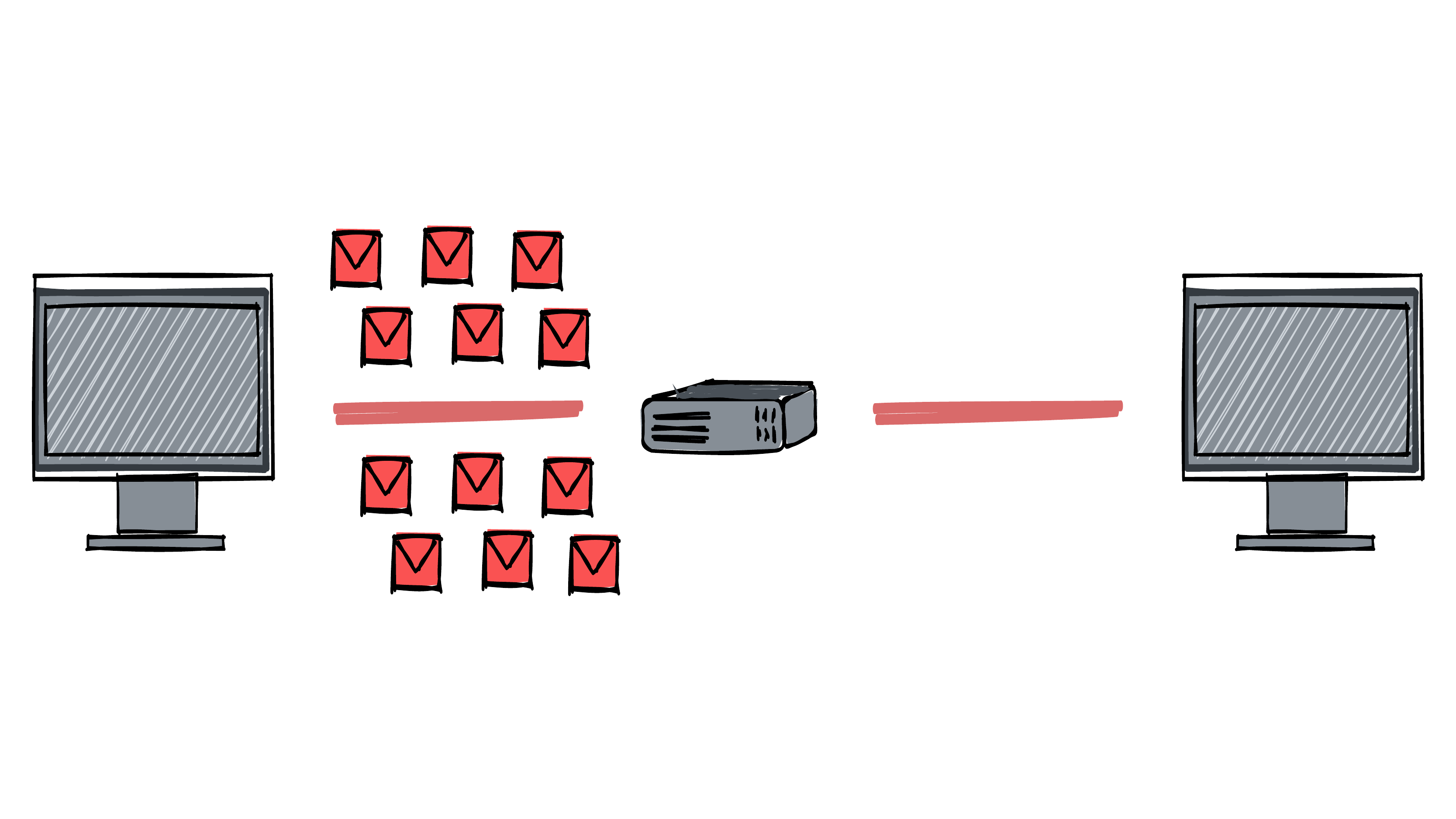}
         \caption{OOD settings, of a much higher congestion rate, with a significantly greater packet loss}
         \label{fig:AuroraOod:b}
     \end{subfigure}
    
    \caption{Aurora: illustration of in-distribution and OOD settings.}
    \label{fig:AuroraInDistributionAndOod}
\end{figure}
\FloatBarrier

\newpage
\noindent \normalsize \textbf{ Experiment(\ref{exp:auroraShort}): Aurora: Short Training Episodes}


\FloatBarrier


\experiment{Aurora: Long Training Episodes}
~\label{exp:auroraLong}

Similar to Experiment~\ref{exp:auroraShort}, we trained a new set of $k=16$ 
agents. In this experiment, we increased each training episode to consist of 
$400$ steps (instead of $50$, as in the ``short'' training). The rest of the 
parameters were identical to the previous setup in 
Experiment~\ref{exp:auroraShort}. This time, $5$ models performed poorly in the 
OOD environment (i.e., did not reach our reward threshold of $99$), while the 
remaining $11$ models performed well both in-distribution and OOD.

When running our method with the \conditionMax criterion, our algorithm returned $4$ models, all being a subset of the group of $11$ models which generalized successfully, and after fully filtering out all the unsuccessful models. Running the algorithm with the \conditionPercentile or the \conditionCombined criteria also yielded a subset of this group, indicating the filtering process was again successful (and robust to various algorithm hyperparameters).

\begin{figure}[ht]
    \centering
    \captionsetup{justification=centering}
    \subfloat[In-distribution \label{subfig:auroraLongRewards:inDist}]{\includegraphics[width=0.49\textwidth]{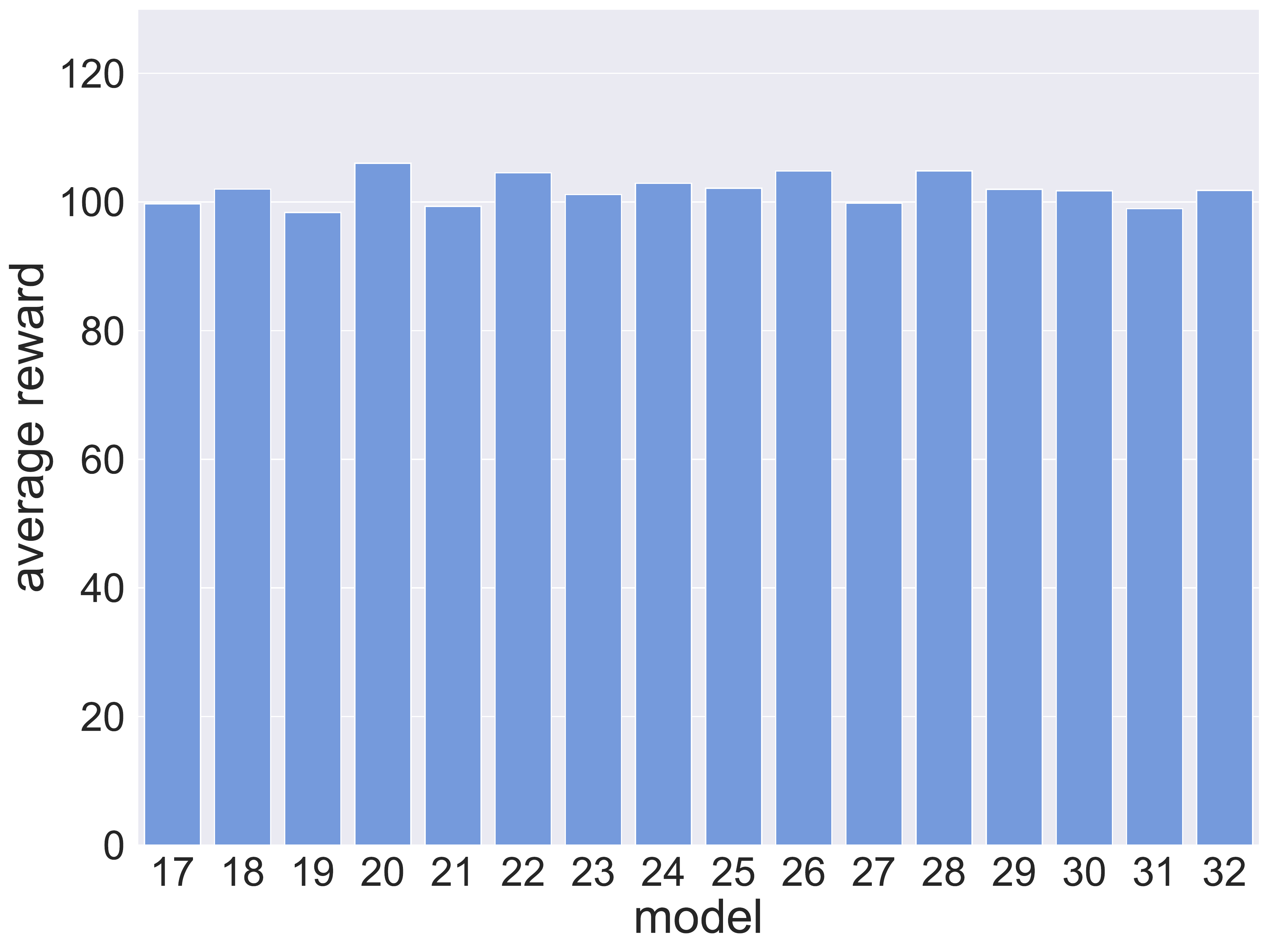}}
    \hfill
    \subfloat[OOD\label{subfig:auroraLongRewards:OOD}]{\includegraphics[width=0.49\textwidth]{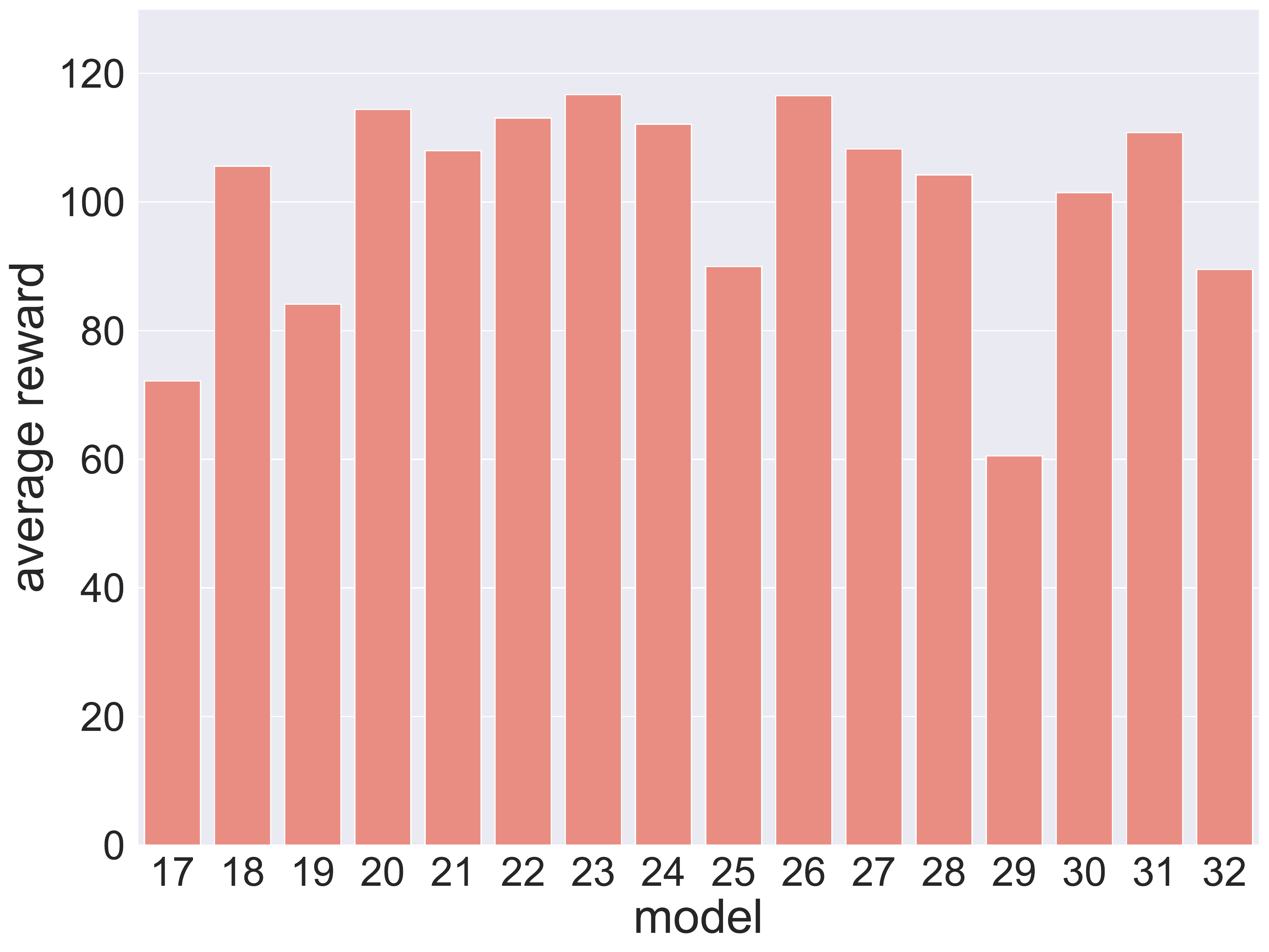}}\\
    
    \caption{Aurora Experiment~\ref{exp:auroraLong}: the models' average 
    rewards when simulated on different distributions.}
    \label{fig:auroraLongRewards}
\end{figure}


\begin{figure}[ht]
    \centering
    \captionsetup{justification=centering}
        \captionsetup[subfigure]{justification=centering}
    \captionsetup{justification=centering} 
     \begin{subfigure}[t]{0.49\linewidth}
         \centering
    \includegraphics[width=\textwidth, height=0.67\textwidth]{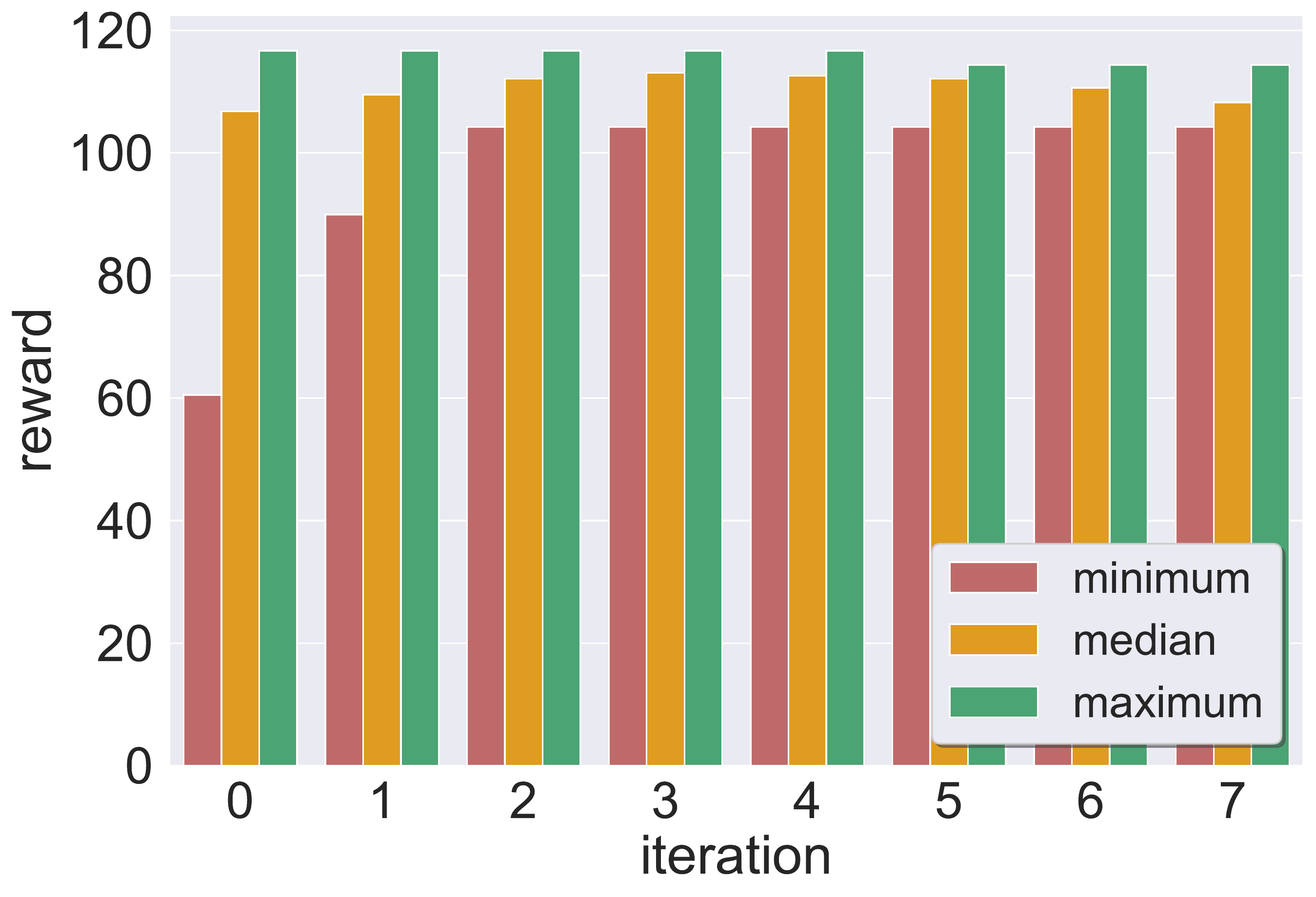}
         \caption{Rewards statistics}
        \label{}
     \end{subfigure}
     \hfill
     \begin{subfigure}[t]{0.49\linewidth}
        \includegraphics[width=\textwidth, height=0.67\textwidth]{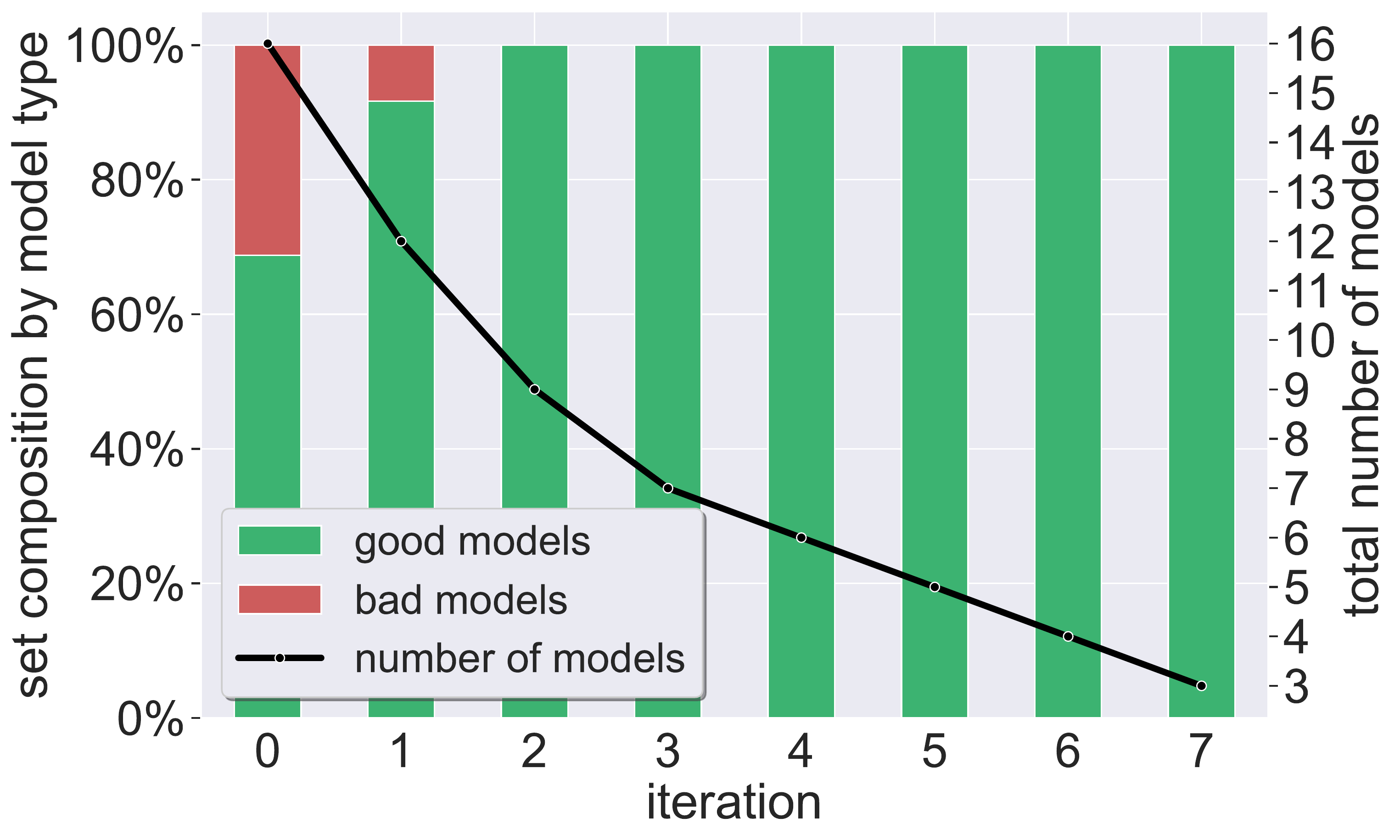}
         \caption{Remaining good and bad models ratio}
        \label{}
     \end{subfigure}

    \caption{Aurora Experiment~\ref{exp:auroraLong}: model selection results.}
    \label{fig:auroraLongGoodBadModelsPercentages}
\end{figure}

\medskip
\noindent
\subsection{Additional Probability Density Functions}  

\begin{figure}
    \centering
    \captionsetup[subfigure]{justification=centering}
    \captionsetup{justification=centering}
    \begin{subfigure}[t]{0.32\linewidth}
         \includegraphics[width=\textwidth]{plots/aurora/models_rewards/short_episodes/rewards_In_distribution.pdf}
         
         \label{subfig:auroraShortRewards:inDist}
     \end{subfigure}
     \hfill
     \begin{subfigure}[t]{0.32\linewidth}
         \centering
    \includegraphics[width=\textwidth]{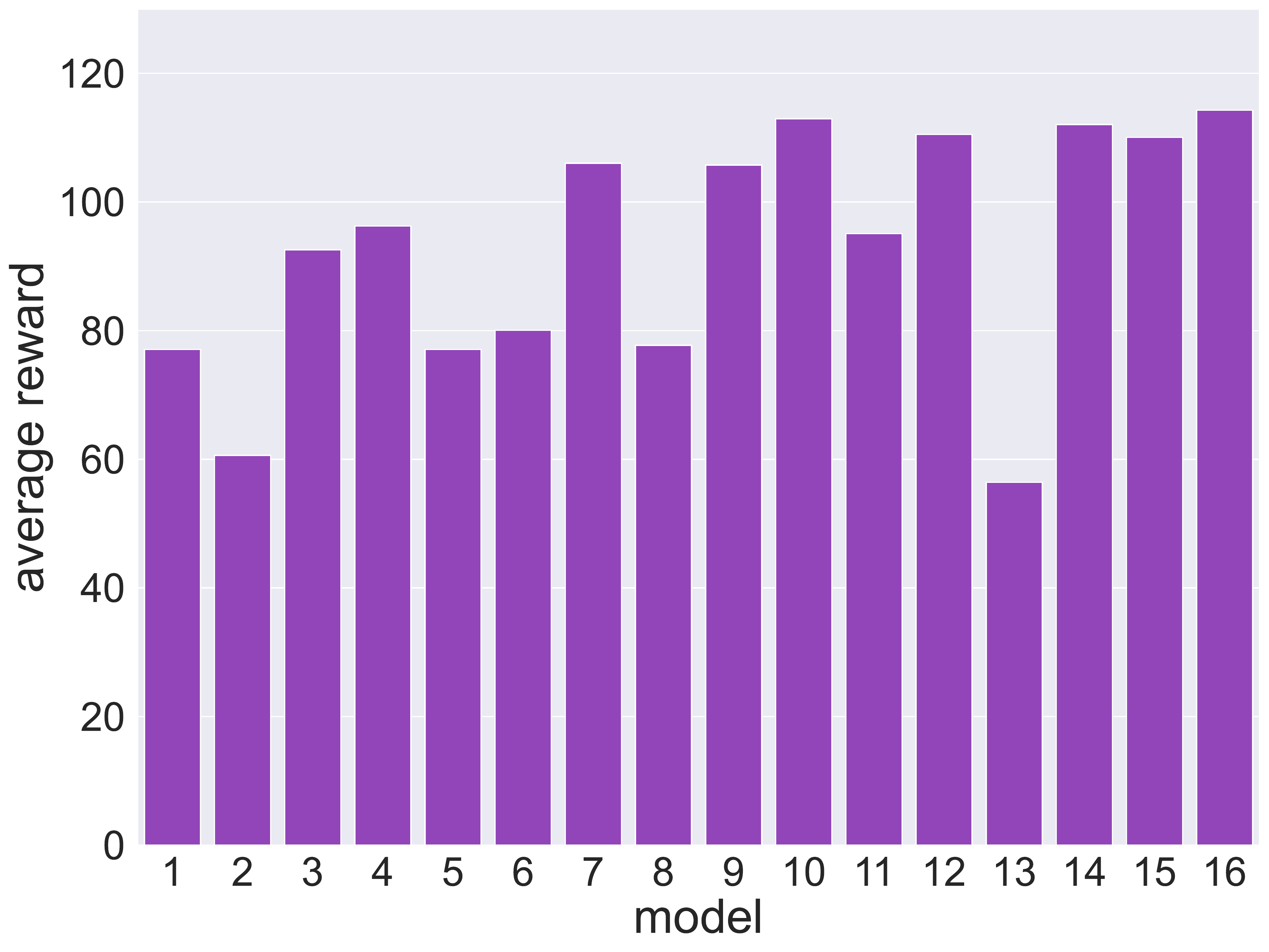}
    \label{subfig:auroraShortRewards:OODLowMean}
     \end{subfigure}
     \hfill
     \begin{subfigure}[t]{0.32\linewidth}
         \centering
         \includegraphics[width=\textwidth]{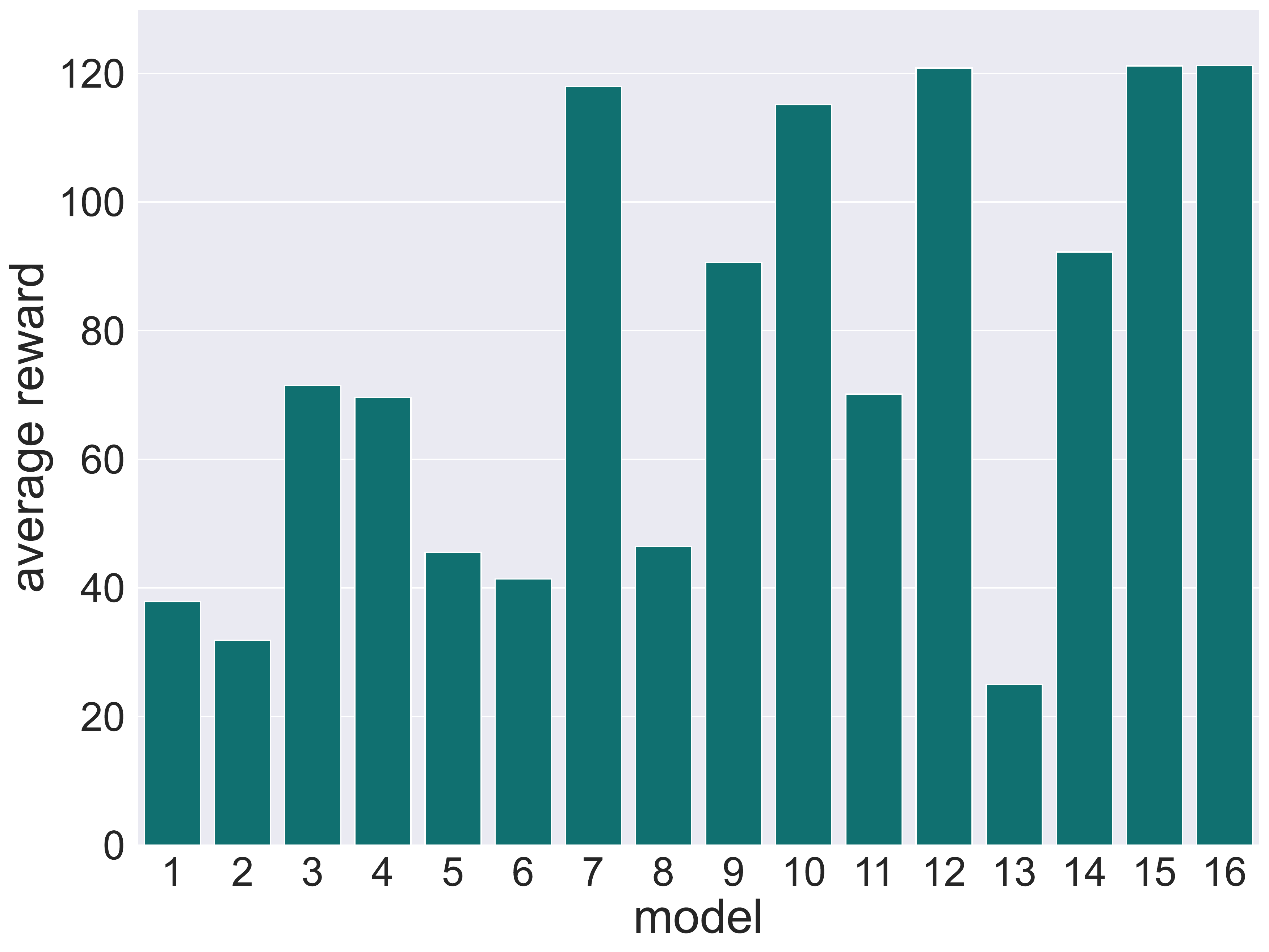}
         \label{subfig:auroraShortRewards:OODHighMean}
     \end{subfigure}     
     \hfill
     \begin{subfigure}[t]{0.32\linewidth}
         \includegraphics[width=\textwidth]{plots/aurora/models_rewards/long_episodes/rewards_In_distribution.pdf}
         
         \caption{In-distribution\\($\sim\mathcal{U}(low,high)$)}
         \label{subfig:auroraLongRewards:inDist}
     \end{subfigure}
     \hfill
     \begin{subfigure}[t]{0.32\linewidth}
         \centering
    \includegraphics[width=\textwidth]{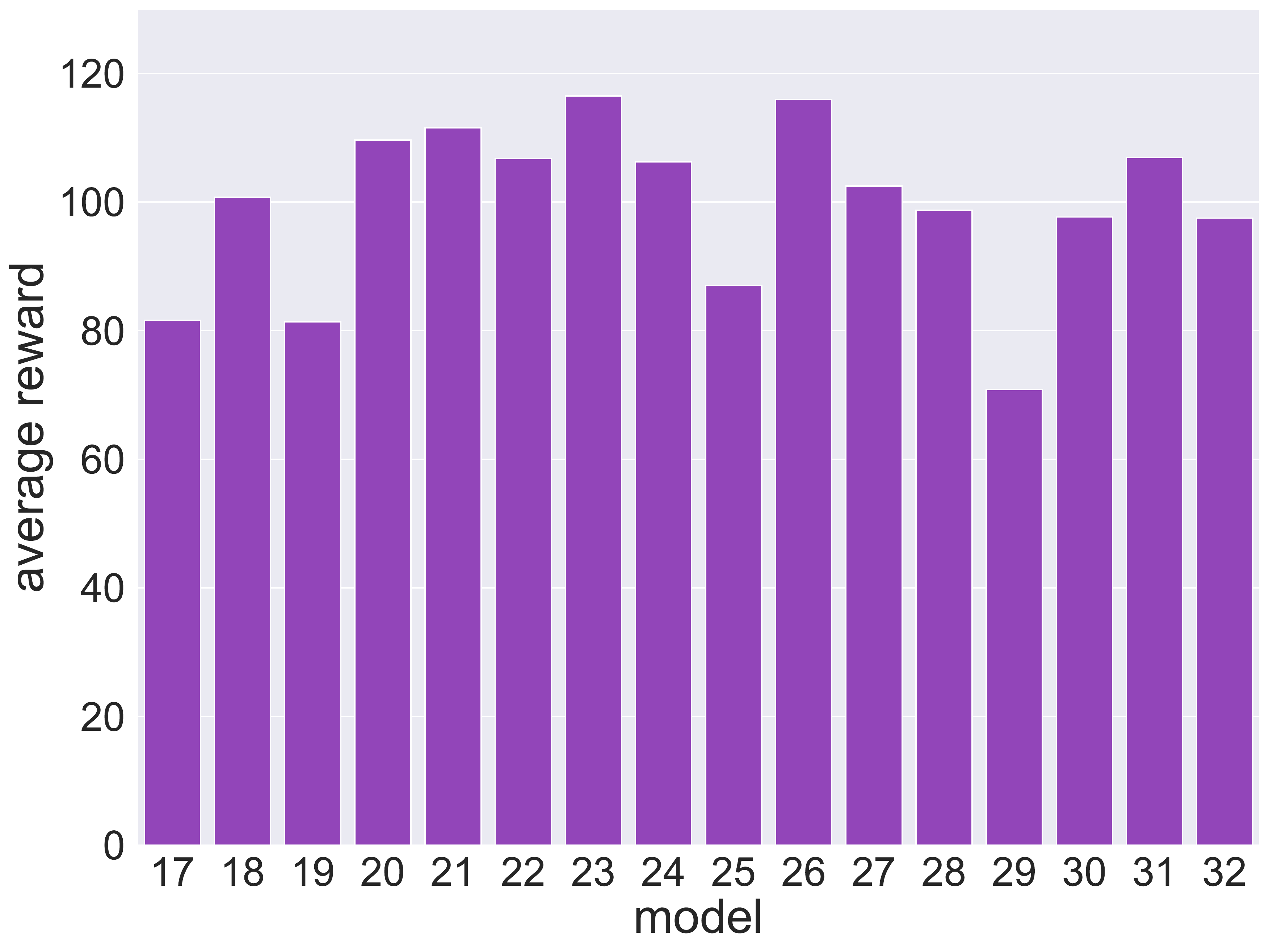}
         \caption{OOD ($\sim\mathcal{TN}(\mu_{low}, \sigma^{2}$))}
    \label{subfig:auroraLongRewards:OODLowMean}
     \end{subfigure}
     \hfill
     \begin{subfigure}[t]{0.32\linewidth}
         \centering
         \includegraphics[width=\textwidth]{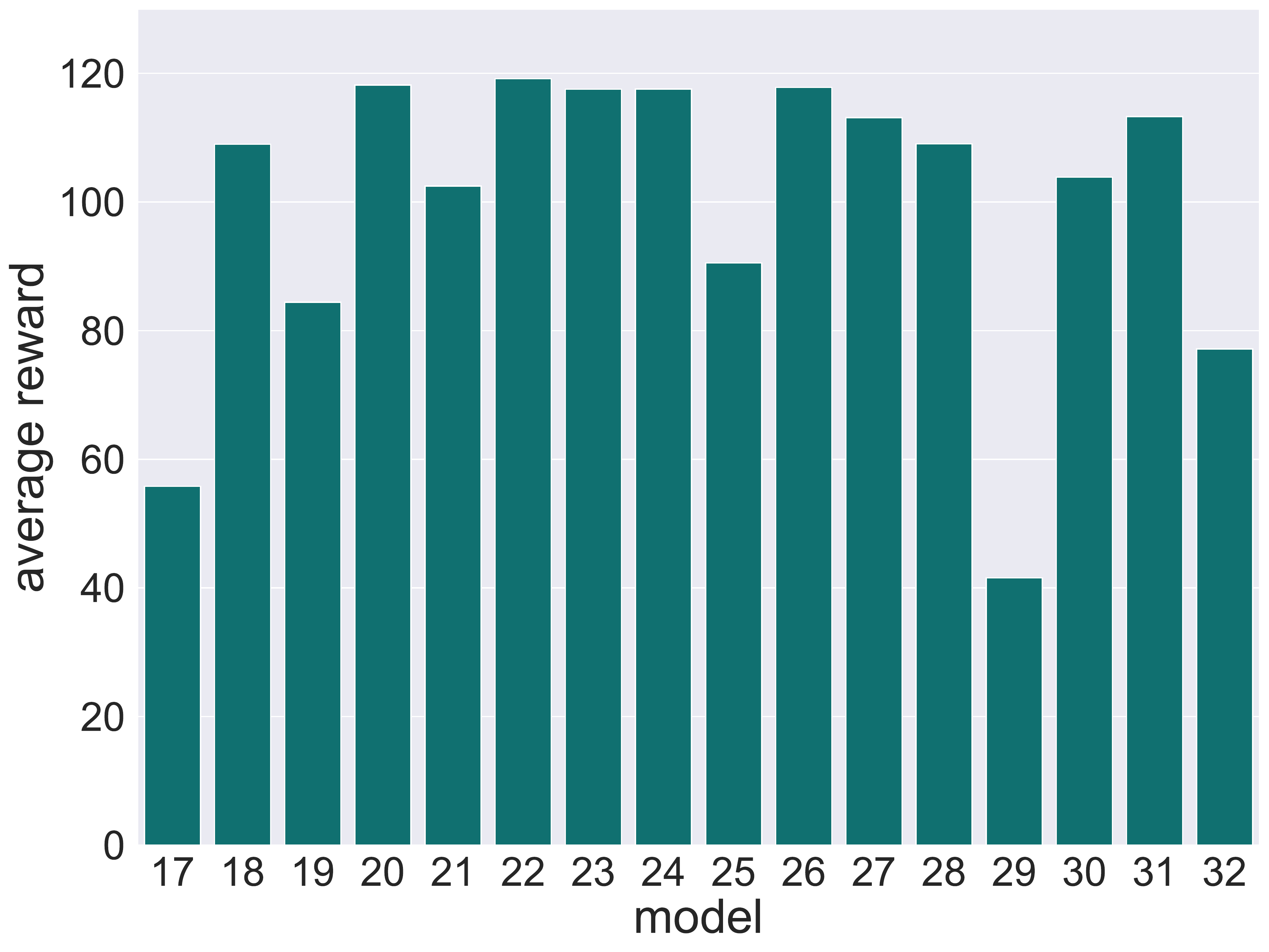}
         \caption{OOD ($\sim\mathcal{TN}(\mu_{high}, \sigma^{2}$))}
         \label{subfig:auroraLongRewards:OODHighMean}
     \end{subfigure}
     
     \caption{Aurora: the models' average rewards under different PDFs.
     \\
     Top row: results for the models used in Experiment~\ref{exp:auroraShort}.
     \\
     Bottom row: results for the models used in 
     Experiment~\ref{exp:auroraLong}.}
    \label{fig:auroraShortDifferentPdsRewards}
\end{figure}

Following are the results discussed in subsection~\ref{subsec:aurora}, and to further demonstrate our method's robustness to different types of
out-of-distribution inputs, we applied it not only to different
\textit{values} (e.g., high \textit{Sending Rate} values) but also to
various \textit{Probability Density Functions} (PDFs) of values in the (OOD) 
input domain in question. More specifically, we repeated the OOD experiments
(Experiment~\ref{exp:auroraShort} and Experiment~\ref{exp:auroraLong}) with 
different
PDFs. In their original settings, all of the environment's parameters
(link's bandwidth, latency, etc.) are uniformly drawn from a range
$[low, high]$. However, in this experiment, we generated two additional PDFs:
\emph{Truncated normal} (denoted as
$\mathcal{TN}_{[low,high]}(\mu, \sigma^{2})$) distributions that are
truncated within the range $[low, high]$). The first PDF was used with
$\mu_{low}=0.3*high+(1-0.3)*low$, and the other with a
$\mu_{high}=0.8*high+(1-0.8)*low$. For both PDFs, the variance,
$\sigma^{2}$, was arbitrarily set to $\frac{high-low}{4}$. These new
distributions are depicted in
Fig.~\ref{fig:auroraShortDifferentPdsRewards} and were used to test
the models from both batches of Aurora experiments (experiments~\ref{exp:auroraShort} and~\ref{exp:auroraLong}).

\begin{figure}[ht]
    \centering
    \captionsetup[subfigure]{justification=centering}
    \captionsetup{justification=centering}
    \begin{subfigure}[t]{0.49\linewidth}
         \includegraphics[width=\textwidth]{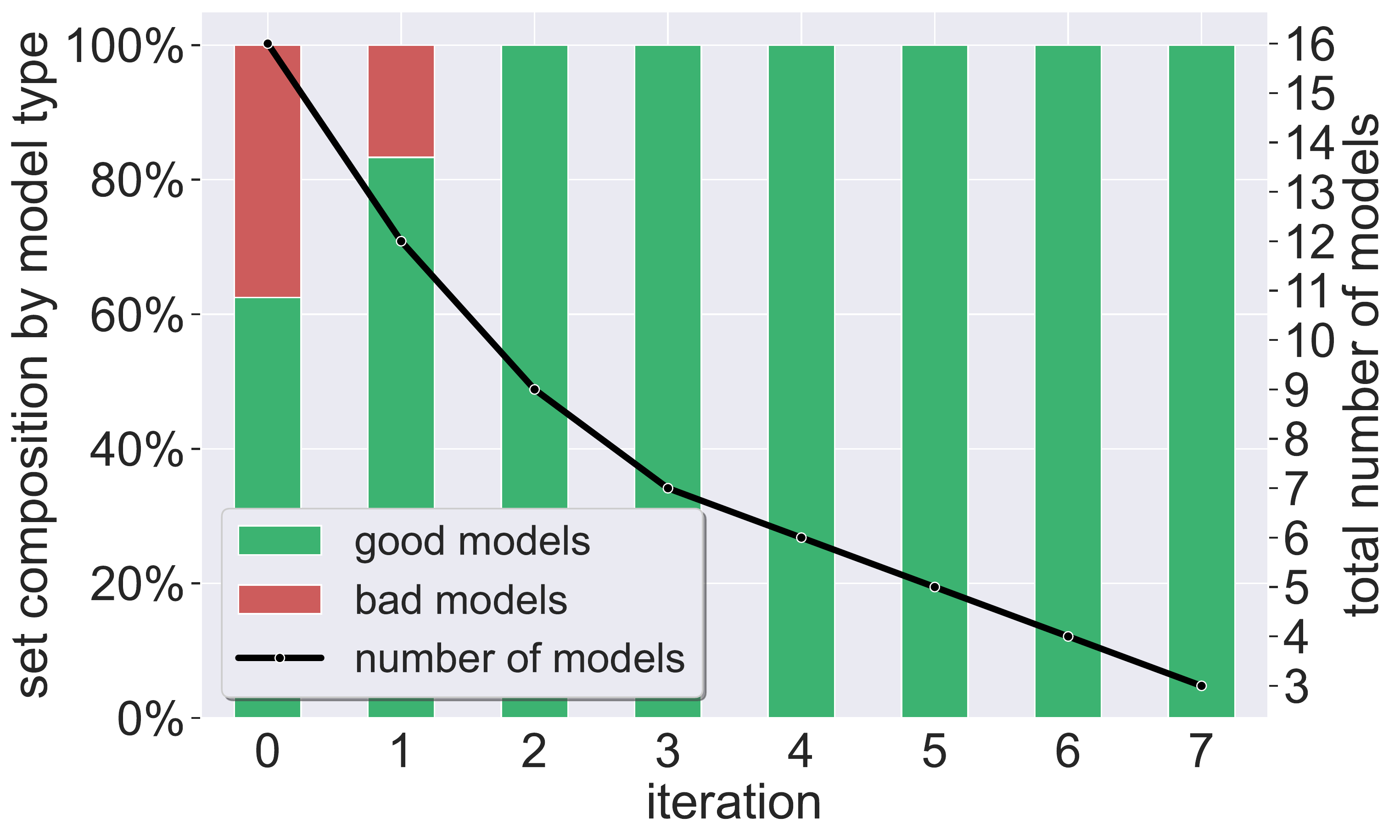}
         \caption{OOD $\sim\mathcal{TN}(\mu_{low}, \sigma^{2}$)}
         \label{subfig:auroraShortOodLowMeanPercentages}
     \end{subfigure}
     \hfill
     \begin{subfigure}[t]{0.49\linewidth}
         \centering
    \includegraphics[width=\textwidth]{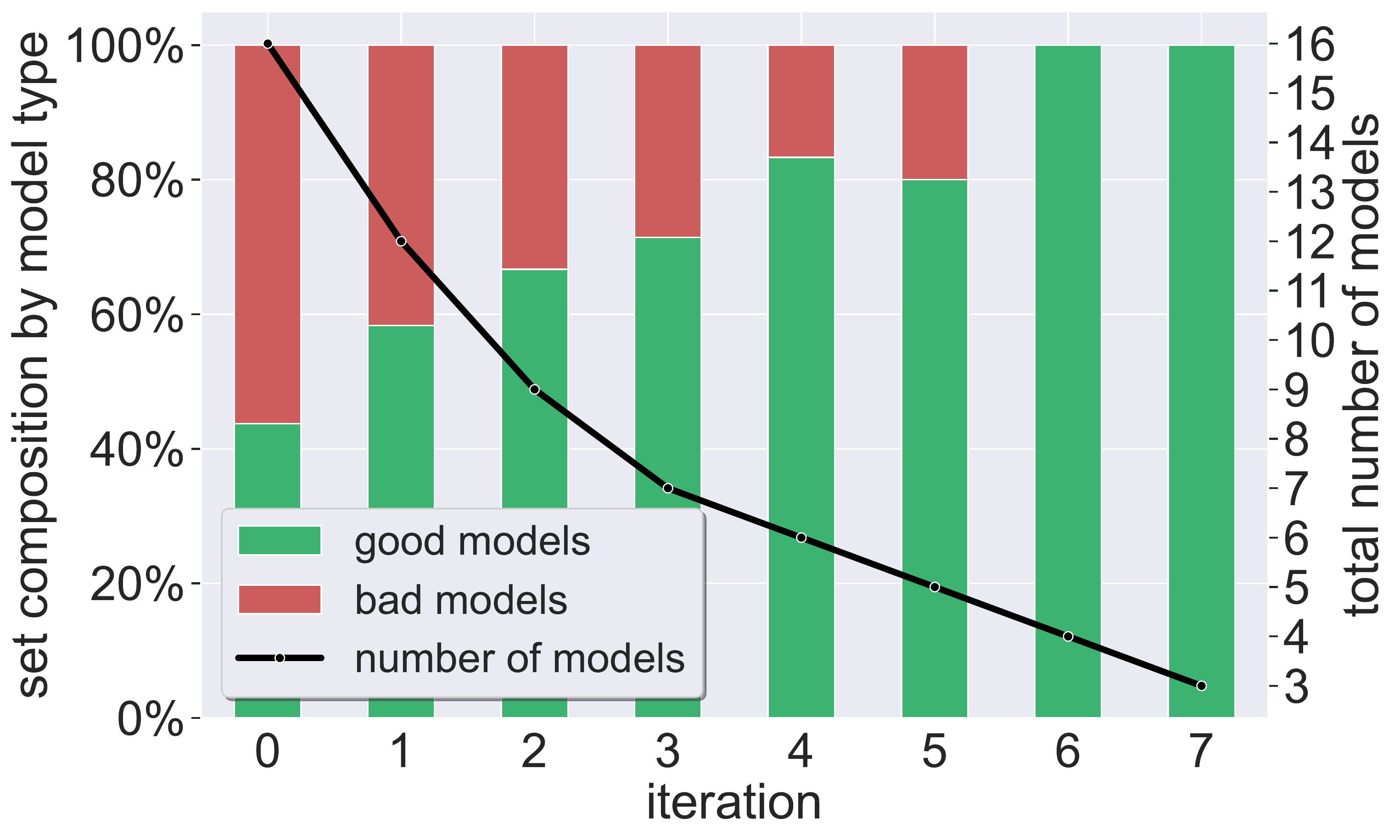}
         \caption{OOD $\sim\mathcal{TN}(\mu_{high}, \sigma^{2}$)}
    \label{subfig:auroraShortOodHighMeanPercentages}
     \end{subfigure}
    \caption{Aurora: Additional PDFs: model selection results for OOD values; 
    the models are the same as in Experiment~\ref{exp:auroraShort}.}
    \label{fig:auroraShortDifferentPdfGoodBadModelsPercentages}
\end{figure}

\begin{figure}[ht]
    \centering
    \captionsetup[subfigure]{justification=centering}
    \captionsetup{justification=centering}
    \begin{subfigure}[t]{0.49\linewidth}
         \includegraphics[width=\textwidth]{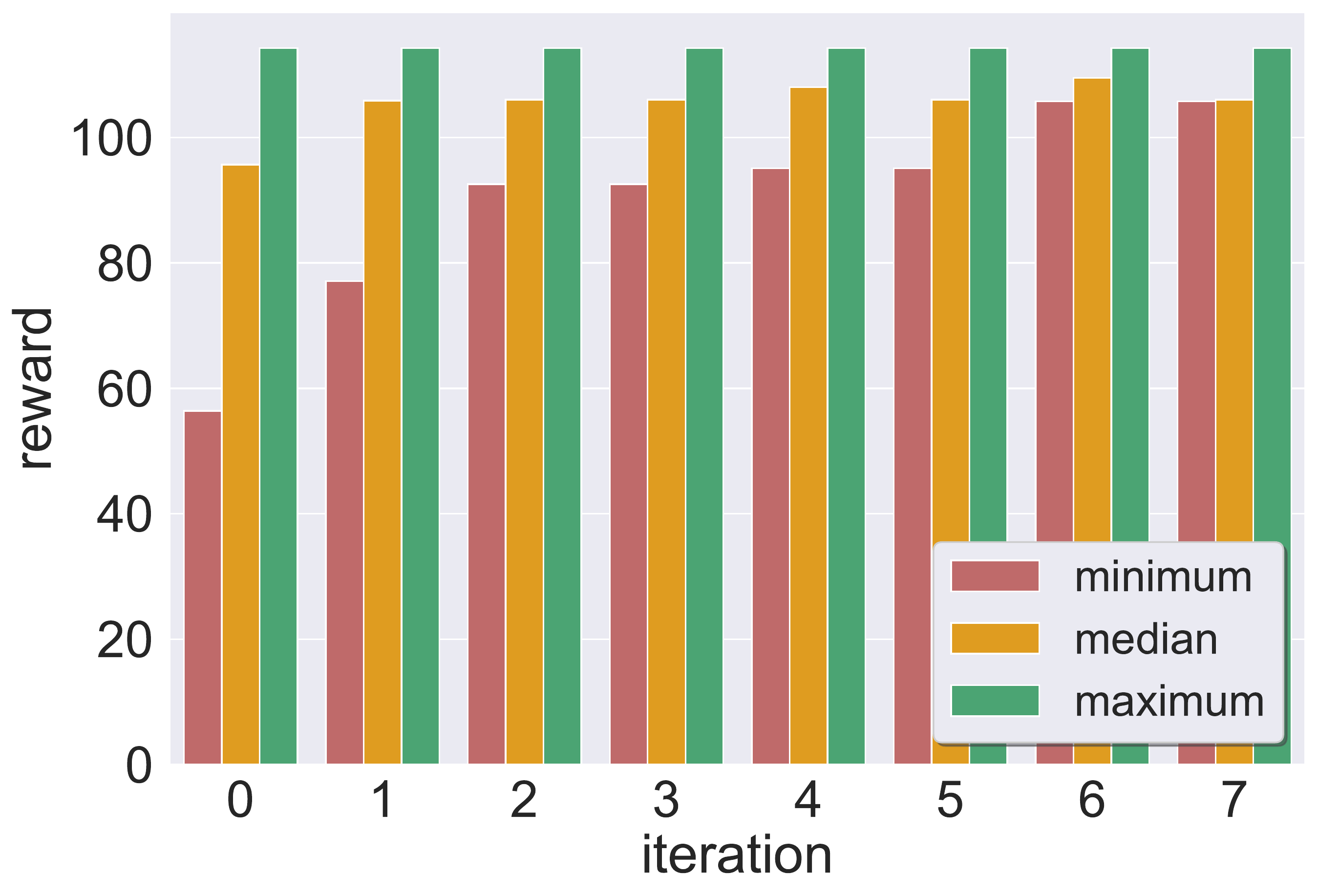}
    \label{subfig:auroraShortOodLowMeanRewardsStats}
         \caption{OOD $\sim\mathcal{TN}(\mu_{low
         }, \sigma^{2}$)}
     \end{subfigure}
     \hfill
     \begin{subfigure}[t]{0.49\linewidth}
         \centering
     \includegraphics[width=\textwidth]{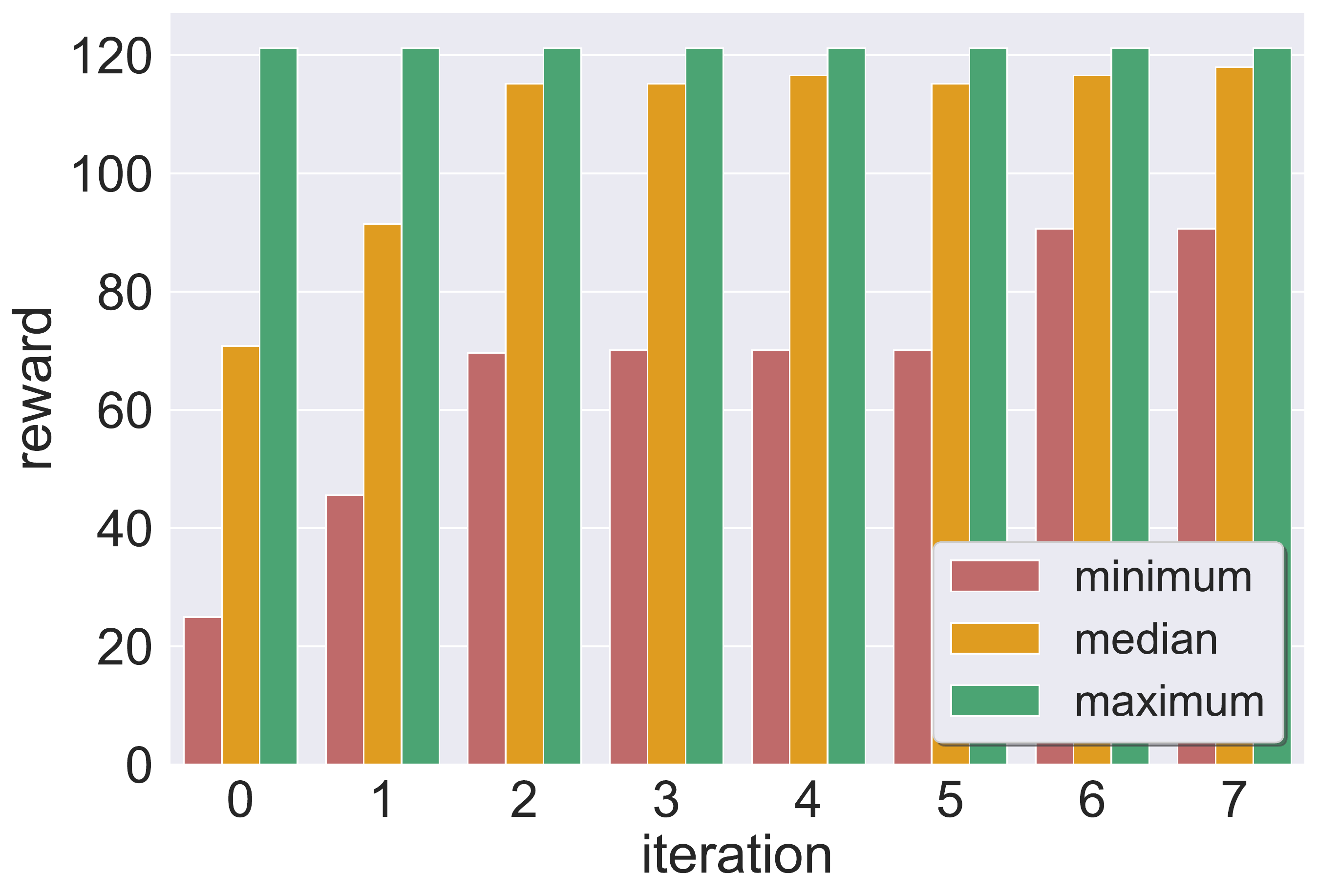}
    \label{subfig:auroraShortOodHighMeanRewardsStat}
        \caption{OOD $\sim\mathcal{TN}(\mu_{high}, \sigma^{2}$)}
     \end{subfigure}
     
    \caption{Aurora: Additional PDFs: model selection results: rewards 
    statistics per iteration;  the models are the same as in 
    Experiment~\ref{exp:auroraShort}.}
    \label{fig:auroraShortDifferentPdfGoodRewardsStats}
\end{figure}

\begin{figure}[ht]
    \centering
    \captionsetup[subfigure]{justification=centering}
    \captionsetup{justification=centering} 
     \begin{subfigure}[t]{0.49\linewidth}
         \includegraphics[width=\textwidth]{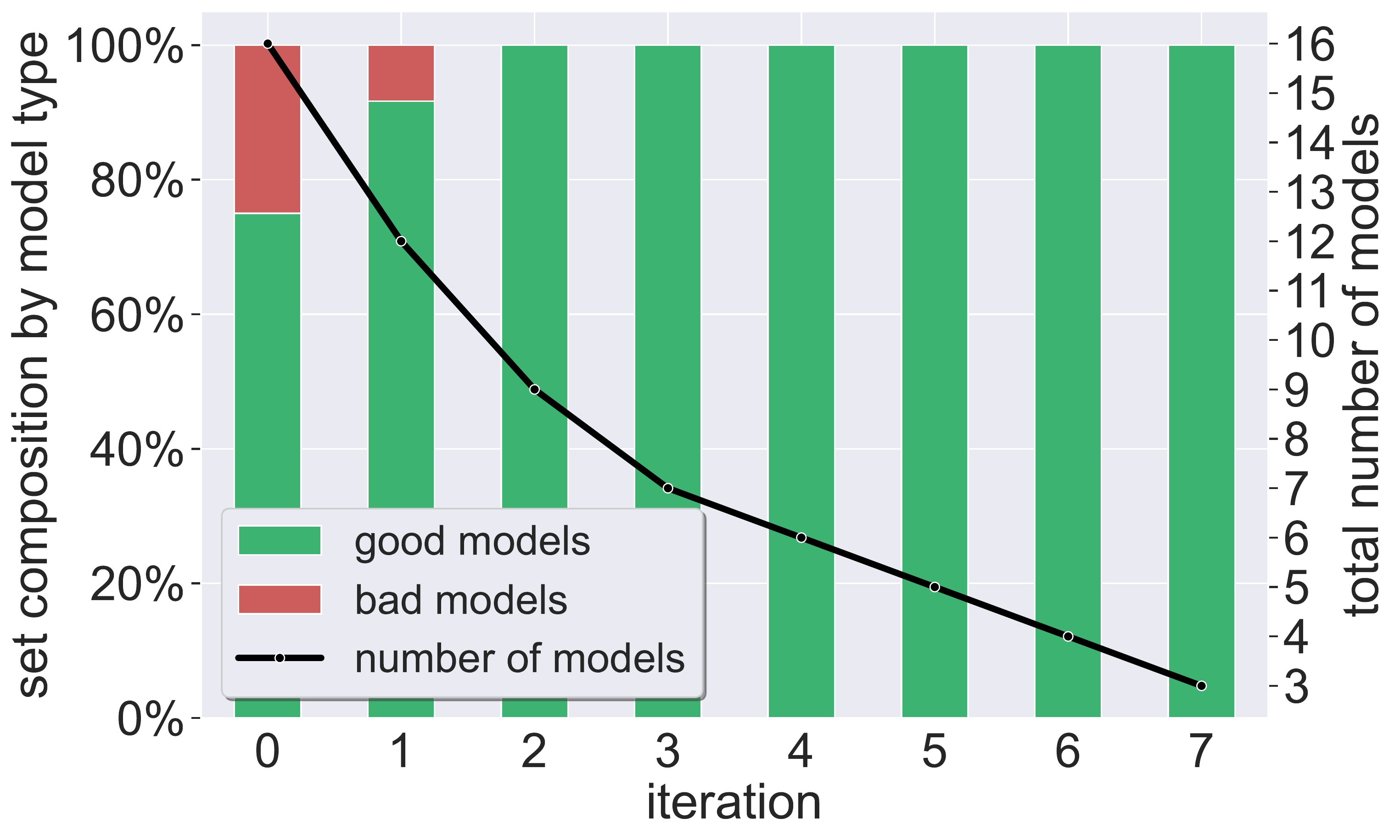}
        \caption{$\sim\mathcal{TN}(\mu_{low}, \sigma^{2}$)} \label{subfig:auroraLongOodHighMeanPercentages}
     \end{subfigure}
     \hfill
     \begin{subfigure}[t]{0.49\linewidth}
         \centering
    \includegraphics[width=\textwidth]{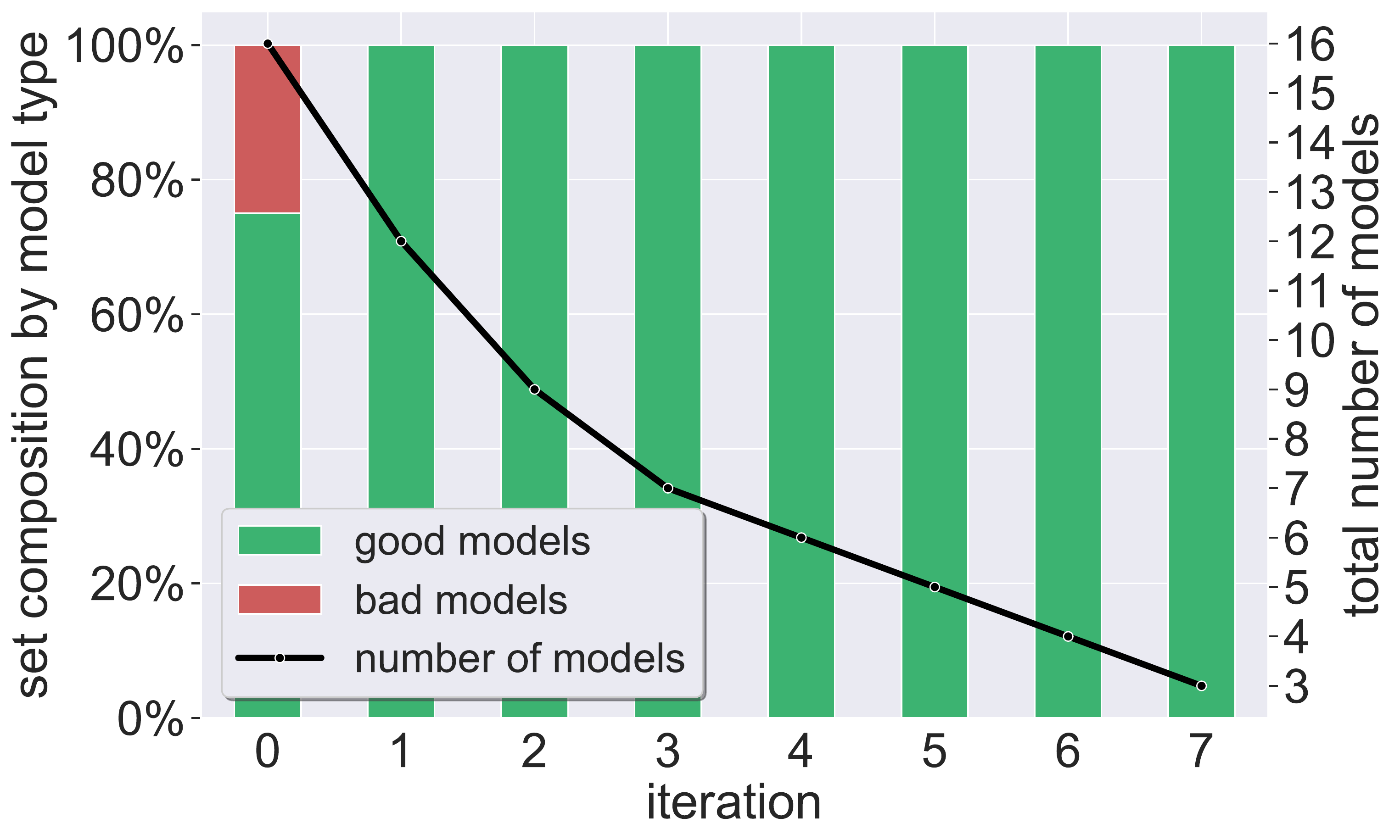}
         \caption{$\sim\mathcal{TN}(\mu_{high}, \sigma^{2}$)}
    \label{subfig:auroraLongOodHighMeanPercentages}
     \end{subfigure}
    \caption{Aurora: Additional PDFs: model selection results for OOD values; 
    the models are the same as in Experiment~\ref{exp:auroraLong}.}
    \label{fig:auroraLongDifferentPdfGoodBadModelsPercentages}
\end{figure}

\begin{figure}[ht]
    \centering
    \captionsetup[subfigure]{justification=centering}
    \captionsetup{justification=centering} 
     \begin{subfigure}[t]{0.49\linewidth}
        \includegraphics[width=\textwidth]{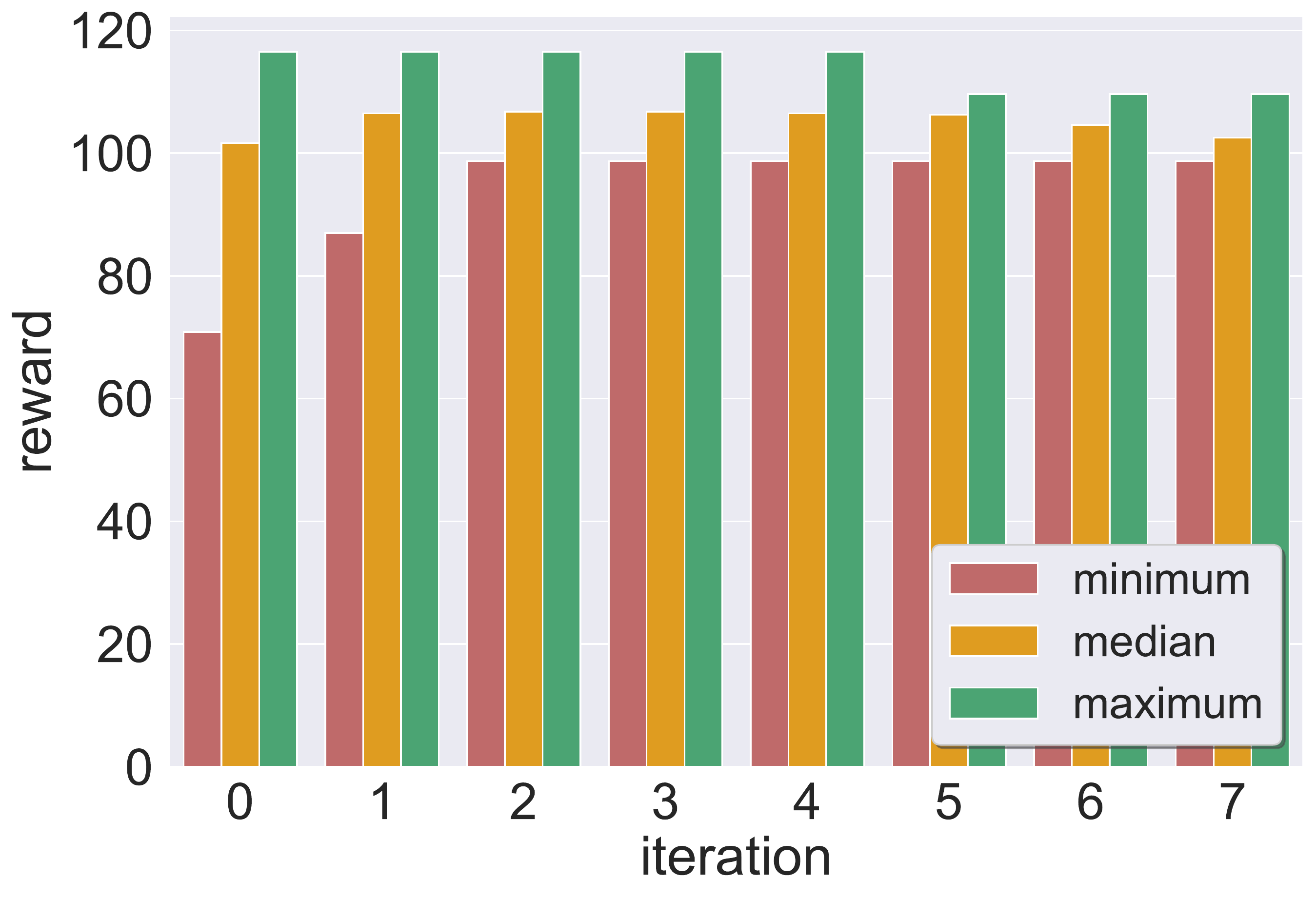}
        \caption{$\sim\mathcal{TN}(\mu_{low}, \sigma^{2}$)}
         \label{subfig:auroraLongOodLowMeanRewardsStats}
     \end{subfigure}
     \hfill
     \begin{subfigure}[t]{0.49\linewidth}
         \centering
    \includegraphics[width=\textwidth]{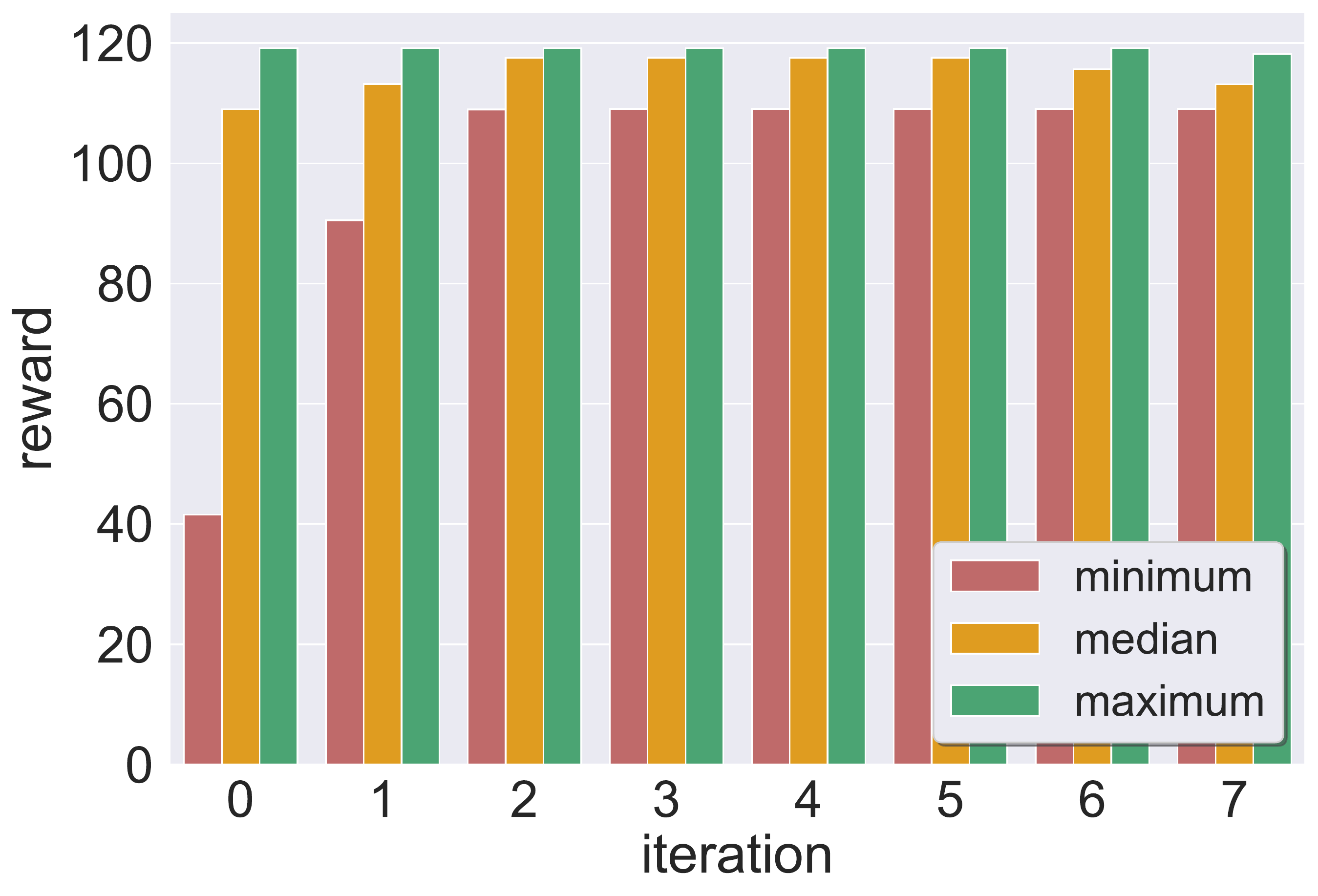}
         \caption{$\sim\mathcal{TN}(\mu_{high}, \sigma^{2}$)}
    \label{subfig:auroraLongOodLowMeanRewardsStats}
     \end{subfigure}
    \caption{Aurora: Additional PDFs: model selection results: rewards 
    statistics per iteration;  the models are the same as in 
    Experiment~\ref{exp:auroraLong}.}    
    \label{fig:auroraLongDifferentPdfGoodRewardsStats}
\end{figure}

\clearpage

\subsection{Additional Filtering Criteria: Experiment~\ref{exp:auroraShort}}

\begin{figure}[h]
    \centering
    \captionsetup[subfigure]{justification=centering}
    \captionsetup{justification=centering} 
     \begin{subfigure}[t]{0.49\linewidth}
         \centering
    \includegraphics[width=\textwidth, height=0.67\textwidth]{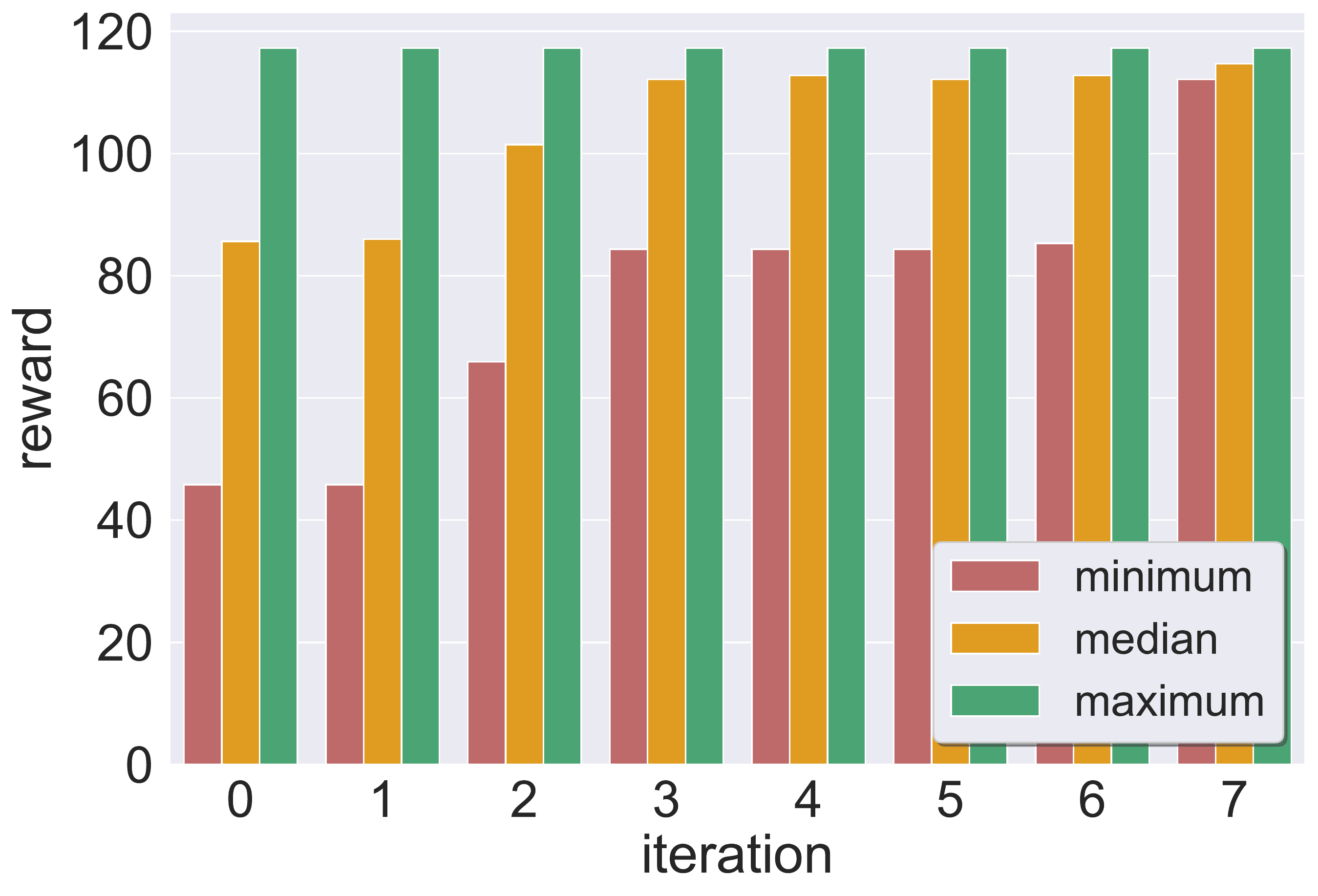}
         \caption{Rewards statistics}
        \label{}
     \end{subfigure}
     \hfill
     \begin{subfigure}[t]{0.49\linewidth}
        \includegraphics[width=\textwidth, height=0.67\textwidth]{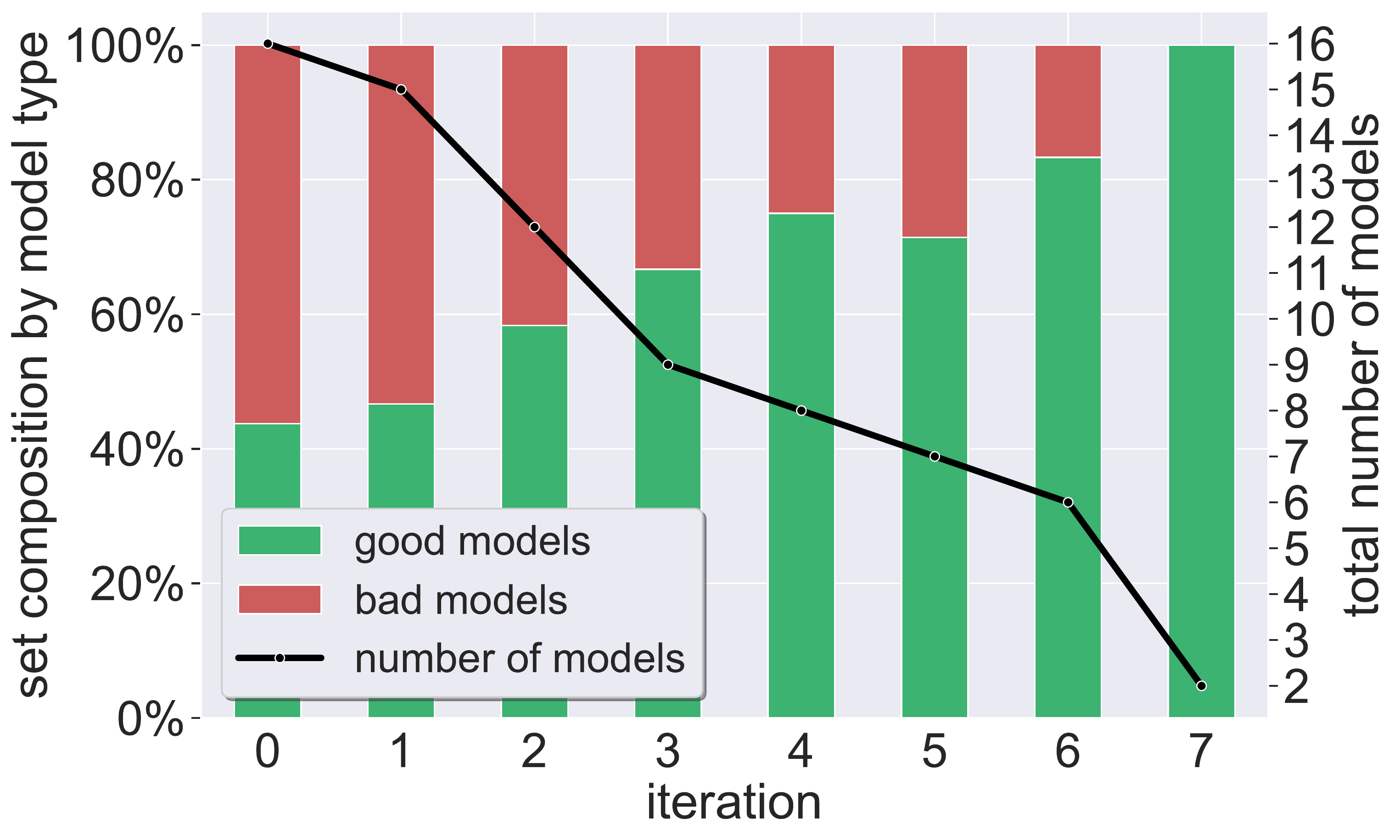}
         \caption{Remaining good and bad models ratio}
        \label{}
     \end{subfigure}
    \caption{Aurora Experiment~\ref{exp:auroraShort}: results using the \maxAgg 
    filtering criterion.}
    \label{fig:auroraShortMaxFiltering}
\end{figure}

\begin{figure}[h]
    \centering
    \captionsetup[subfigure]{justification=centering}
    \captionsetup{justification=centering} 
     \begin{subfigure}[t]{0.49\linewidth}
         \centering
    \includegraphics[width=\textwidth, height=0.67\textwidth]{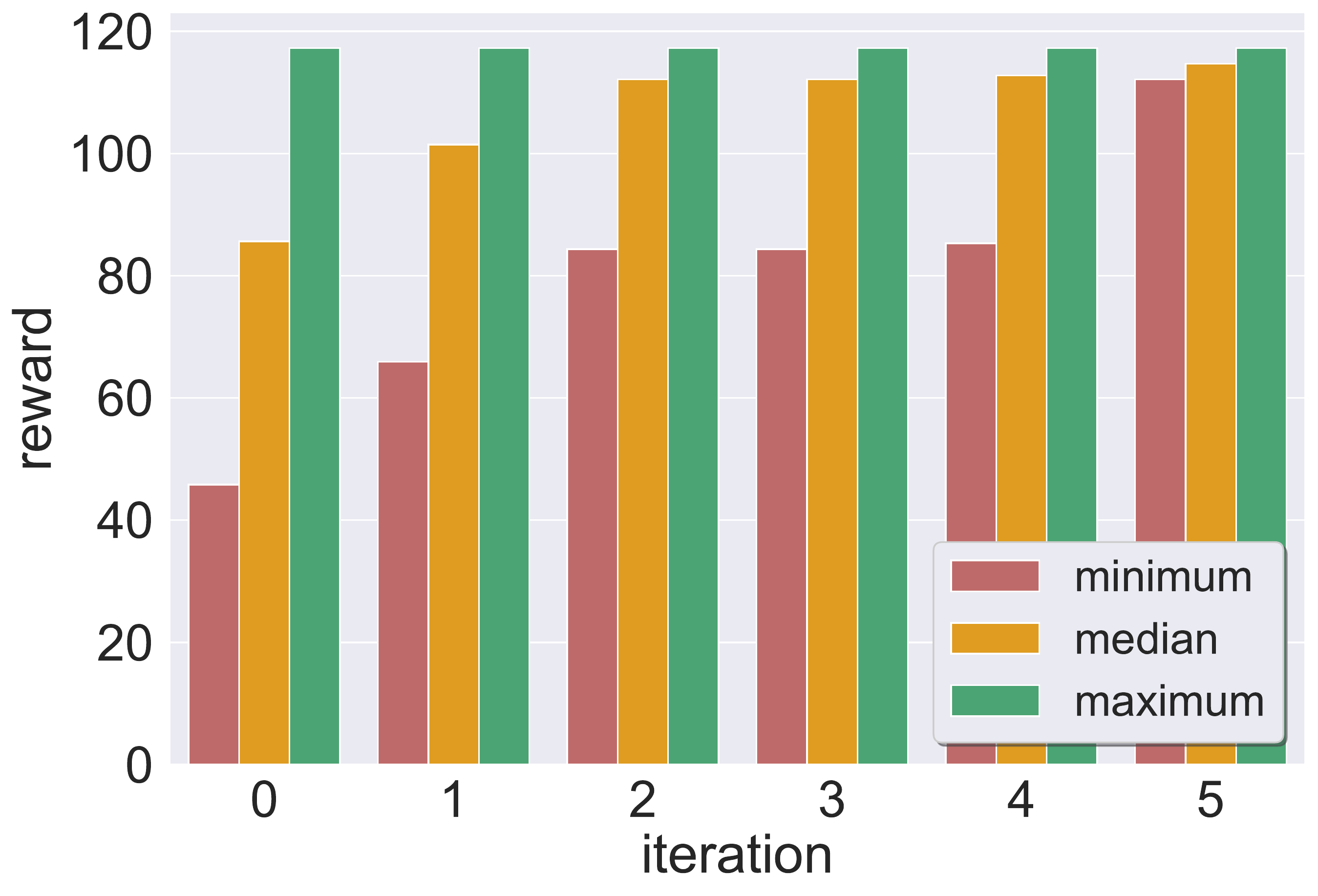}
         \caption{Rewards statistics}
        \label{}
     \end{subfigure}
     \hfill
     \begin{subfigure}[t]{0.49\linewidth}
        \includegraphics[width=\textwidth, height=0.67\textwidth]{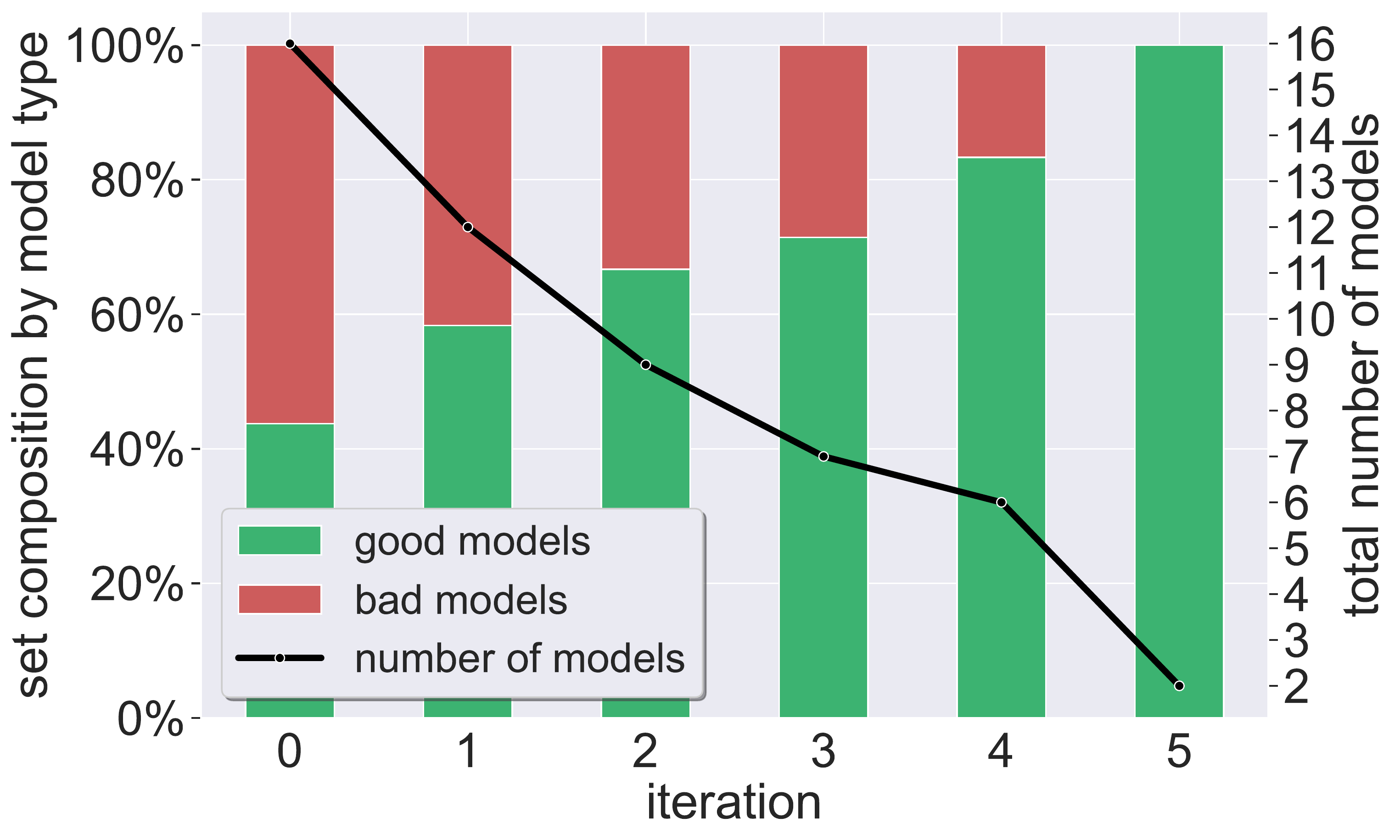}
         \caption{Remaining good and bad models ratio}
        \label{}
     \end{subfigure}
    \caption{Aurora Experiment~\ref{exp:auroraShort}: results using the 
    \conditionCombined filtering criterion.}
\end{figure}
\FloatBarrier

\clearpage

\subsection{Additional Filtering Criteria: Experiment~\ref{exp:auroraLong}}
\begin{figure}[h]
    \centering
    \captionsetup[subfigure]{justification=centering}
    \captionsetup{justification=centering} 
     \begin{subfigure}[t]{0.49\linewidth}
         \centering
    \includegraphics[width=\textwidth, height=0.67\textwidth]{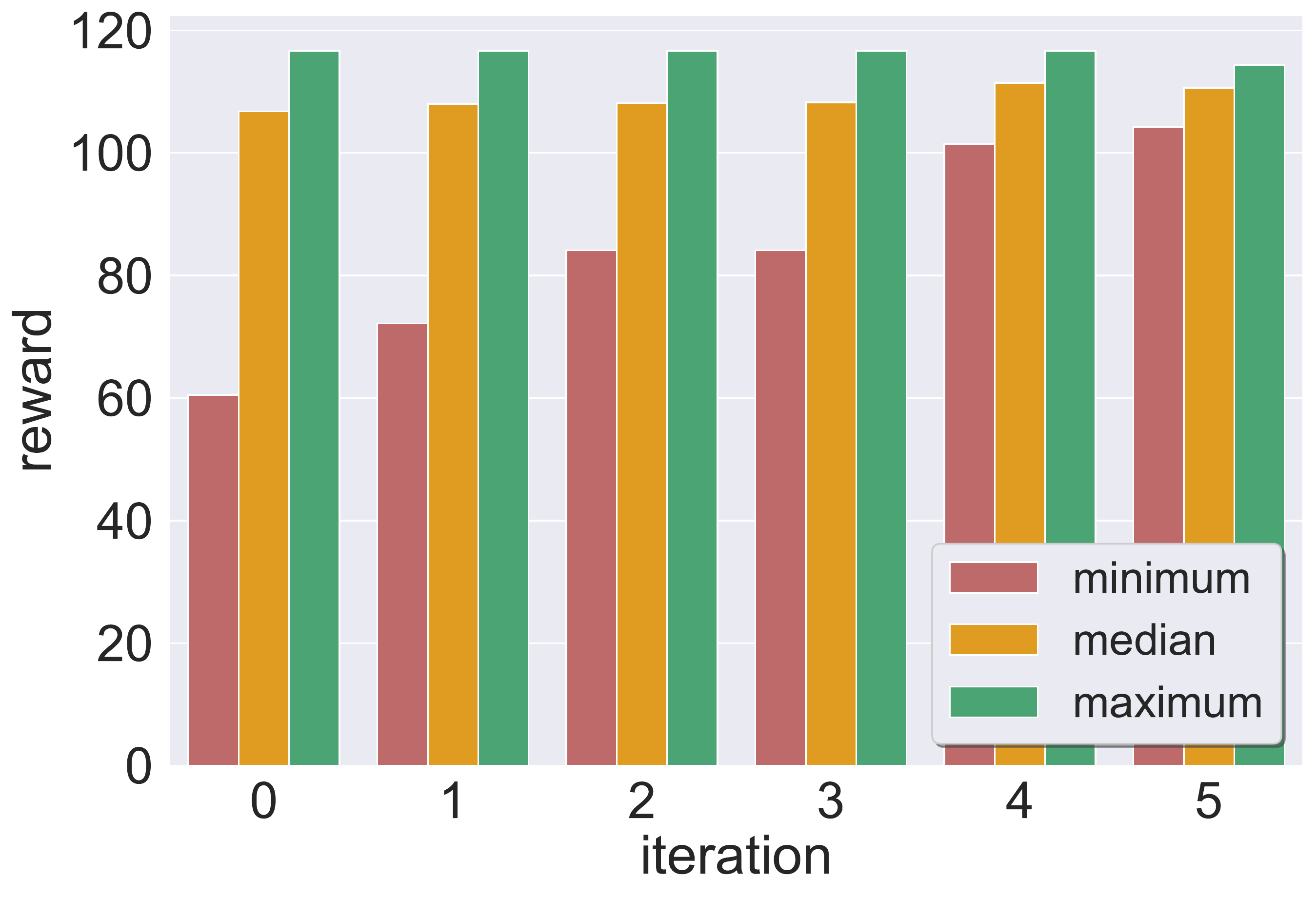}
         \caption{Rewards statistics}
        \label{}
     \end{subfigure}
     \hfill
     \begin{subfigure}[t]{0.49\linewidth}
        \includegraphics[width=\textwidth, height=0.67\textwidth]{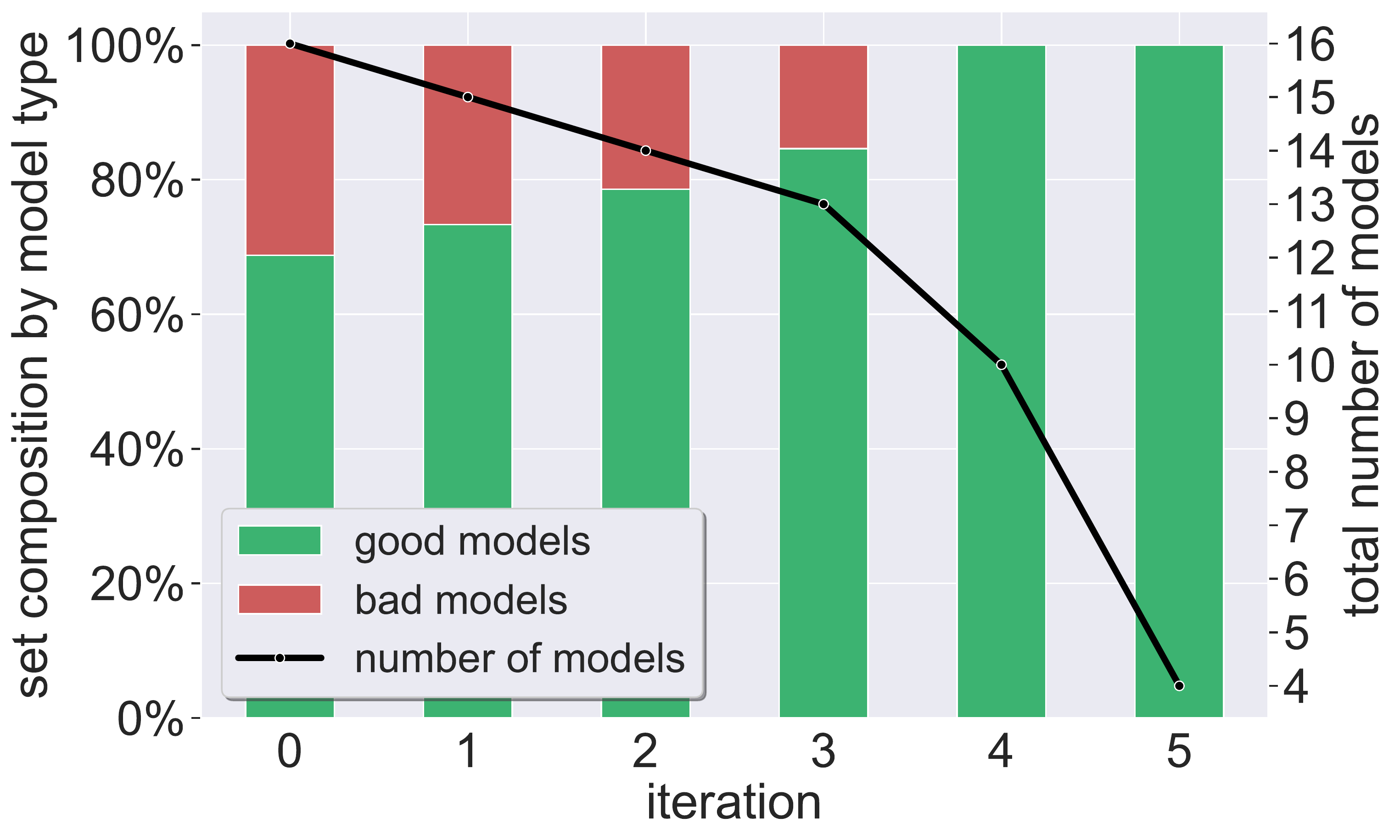}
         \caption{Remaining good and bad models ratio}
        \label{}
     \end{subfigure}
    \caption{Aurora Experiment~\ref{exp:auroraLong}: results using the \maxAgg 
    filtering criterion.}
\end{figure}

\begin{figure}[h]
    \centering
    \captionsetup[subfigure]{justification=centering}
    \captionsetup{justification=centering} 
     \begin{subfigure}[t]{0.49\linewidth}
         \centering
    \includegraphics[width=\textwidth, height=0.67\textwidth]{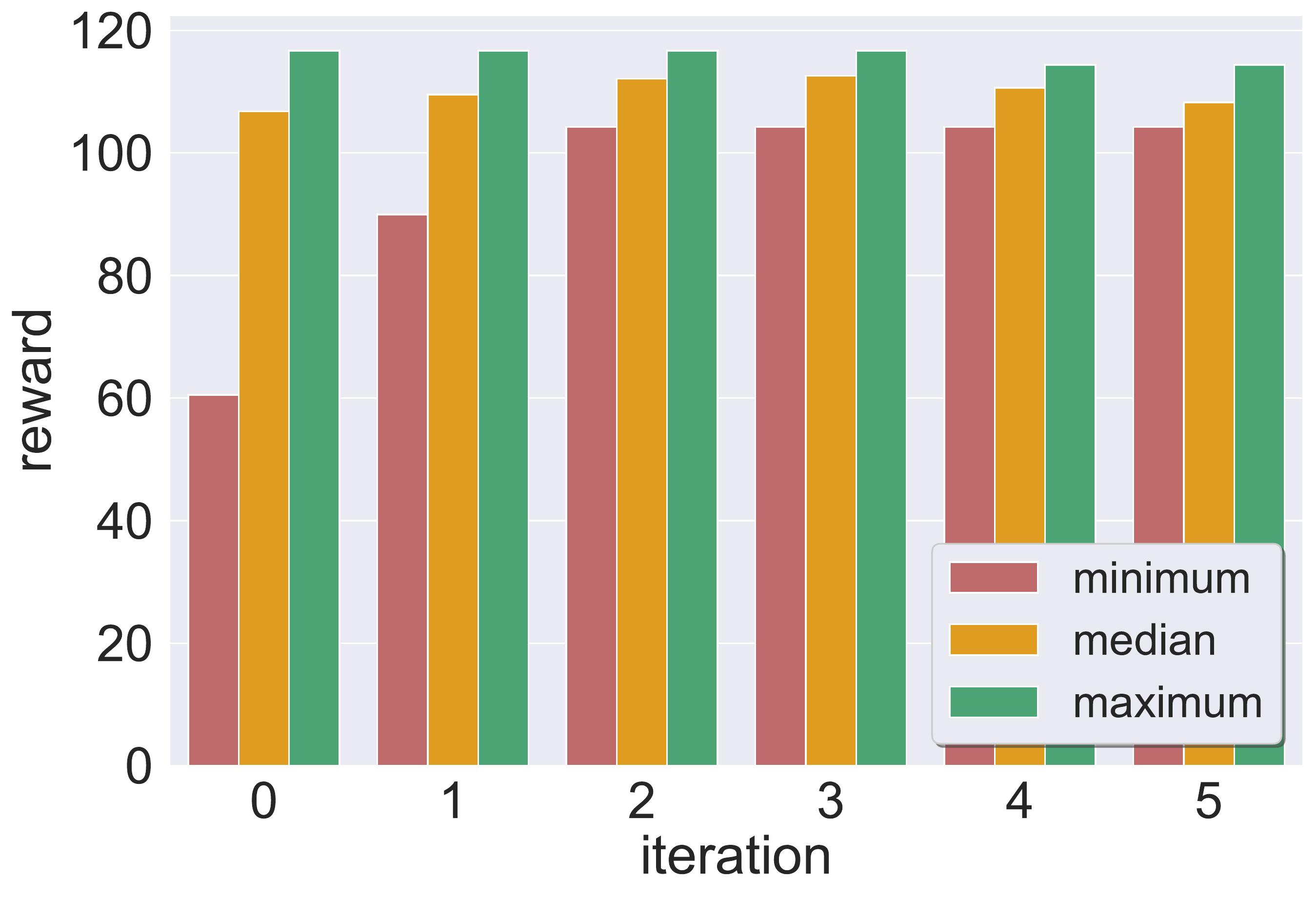}
         \caption{Rewards statistics}
        \label{}
     \end{subfigure}
     \hfill
     \begin{subfigure}[t]{0.49\linewidth}
        \includegraphics[width=\textwidth, height=0.67\textwidth]{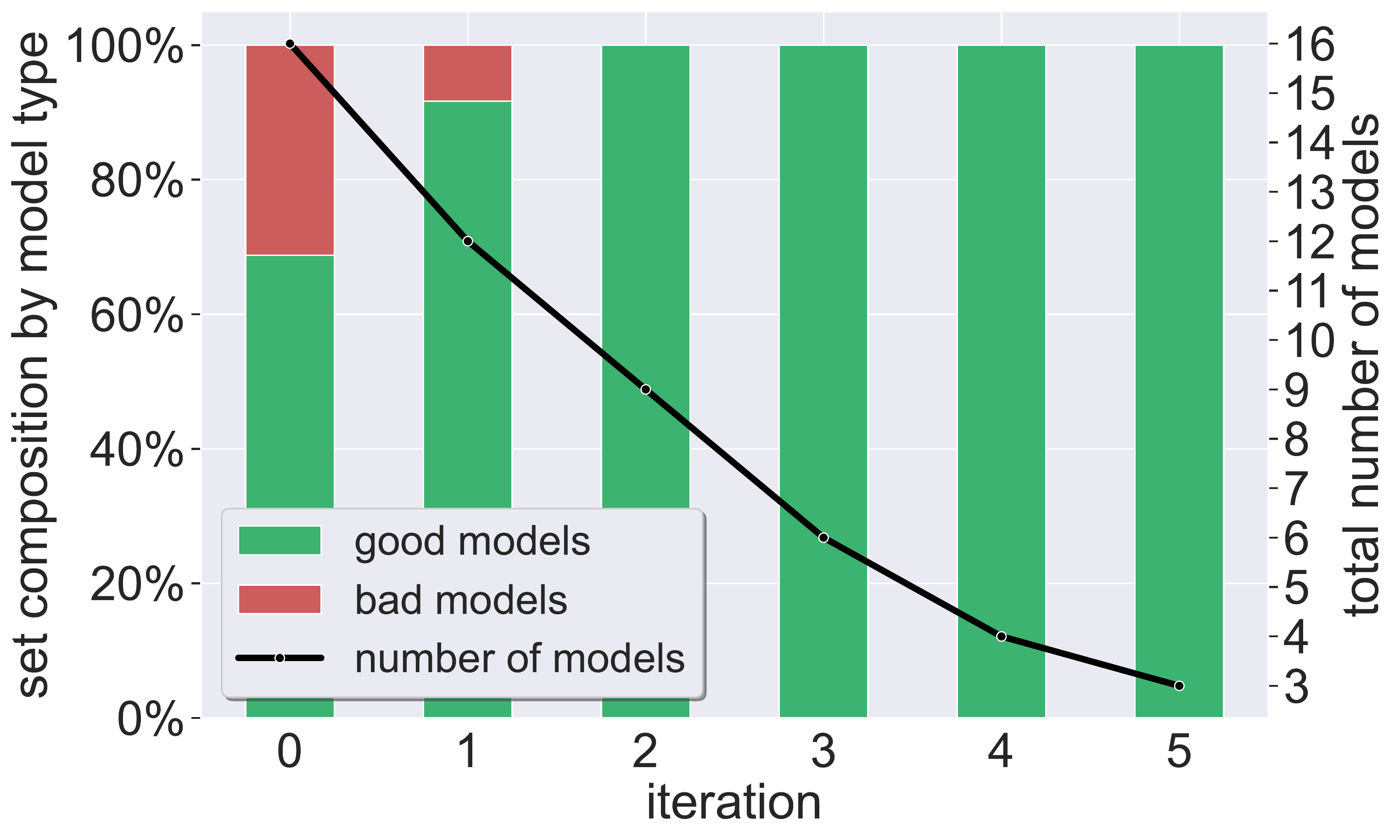}
         \caption{Remaining good and bad models ratio}
        \label{}
     \end{subfigure}
    \caption{Aurora Experiment~\ref{exp:auroraLong}: results using the 
    \conditionCombined filtering criterion.}
\end{figure}
\FloatBarrier

\clearpage

\subsection{Additional Filtering Criteria: Additional PDFs}

\begin{figure}[h]
    \centering
    \captionsetup[subfigure]{justification=centering}
    \captionsetup{justification=centering} 
     \begin{subfigure}[t]{0.49\linewidth}
         \centering
    \includegraphics[width=\textwidth, height=0.67\textwidth]{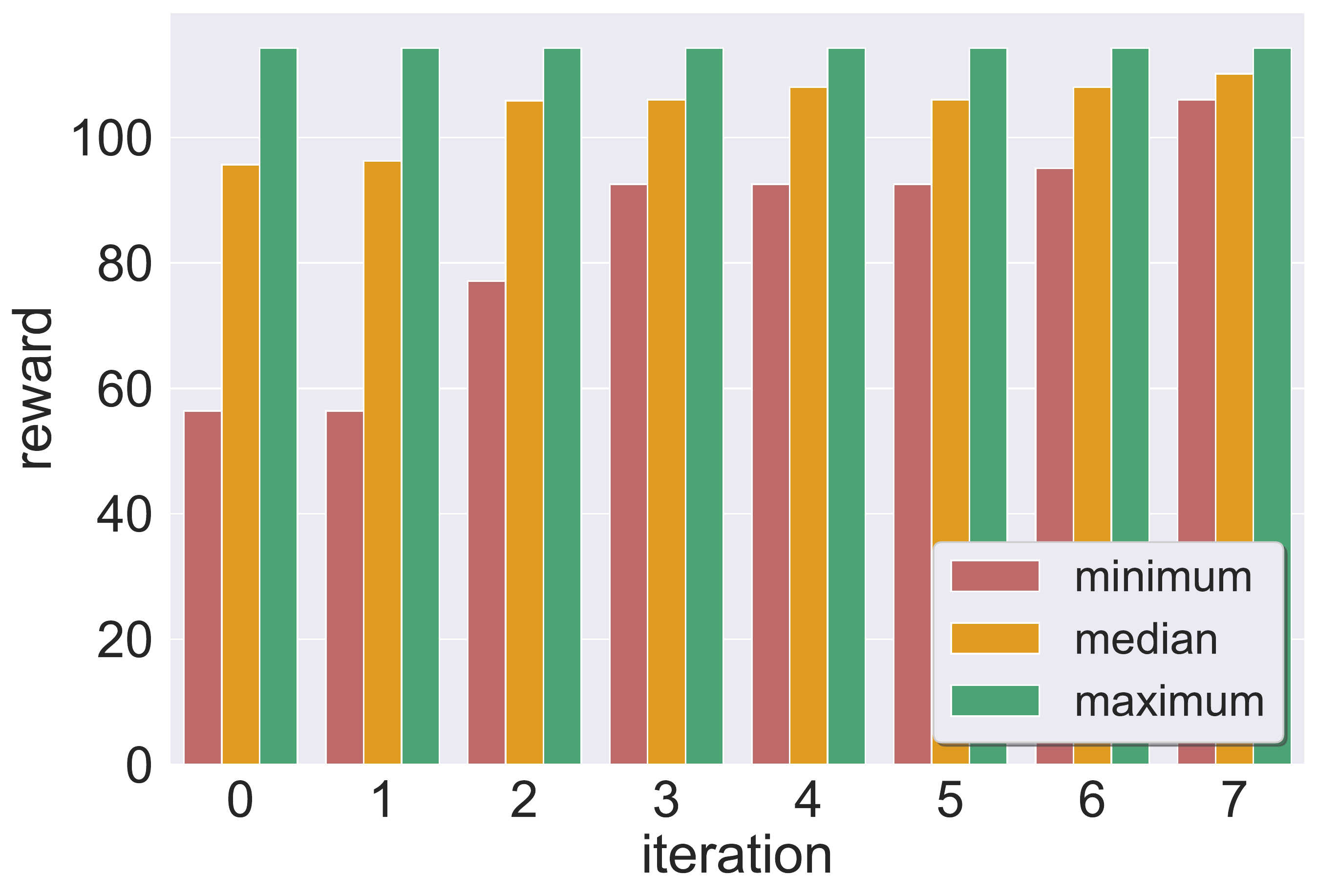}
         \caption{Rewards statistics}
        \label{}
     \end{subfigure}
     \hfill
     \begin{subfigure}[t]{0.49\linewidth}
        \includegraphics[width=\textwidth, height=0.67\textwidth]{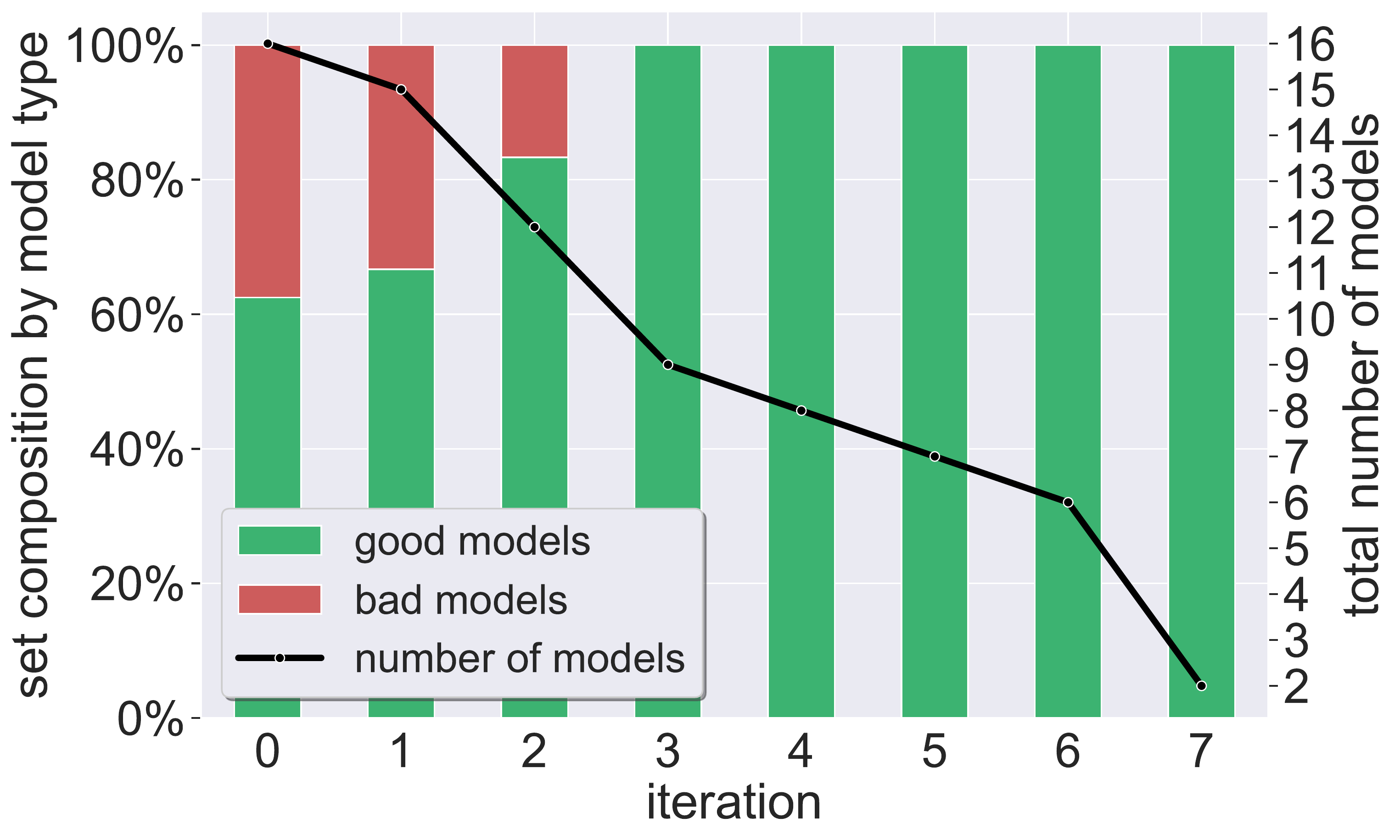}
         \caption{Remaining good and bad models ratio}
        \label{}
     \end{subfigure}
    \caption{Aurora Experiment~\ref{exp:auroraShort}: PDF 
    $\sim\mathcal{TN}(\mu_{low}, \sigma^{2}$): results using the \maxAgg 
    filtering criterion.}
\end{figure}

\begin{figure}[h]
    \centering
    \captionsetup[subfigure]{justification=centering}
    \captionsetup{justification=centering} 
     \begin{subfigure}[t]{0.49\linewidth}
         \centering
    \includegraphics[width=\textwidth, height=0.67\textwidth]{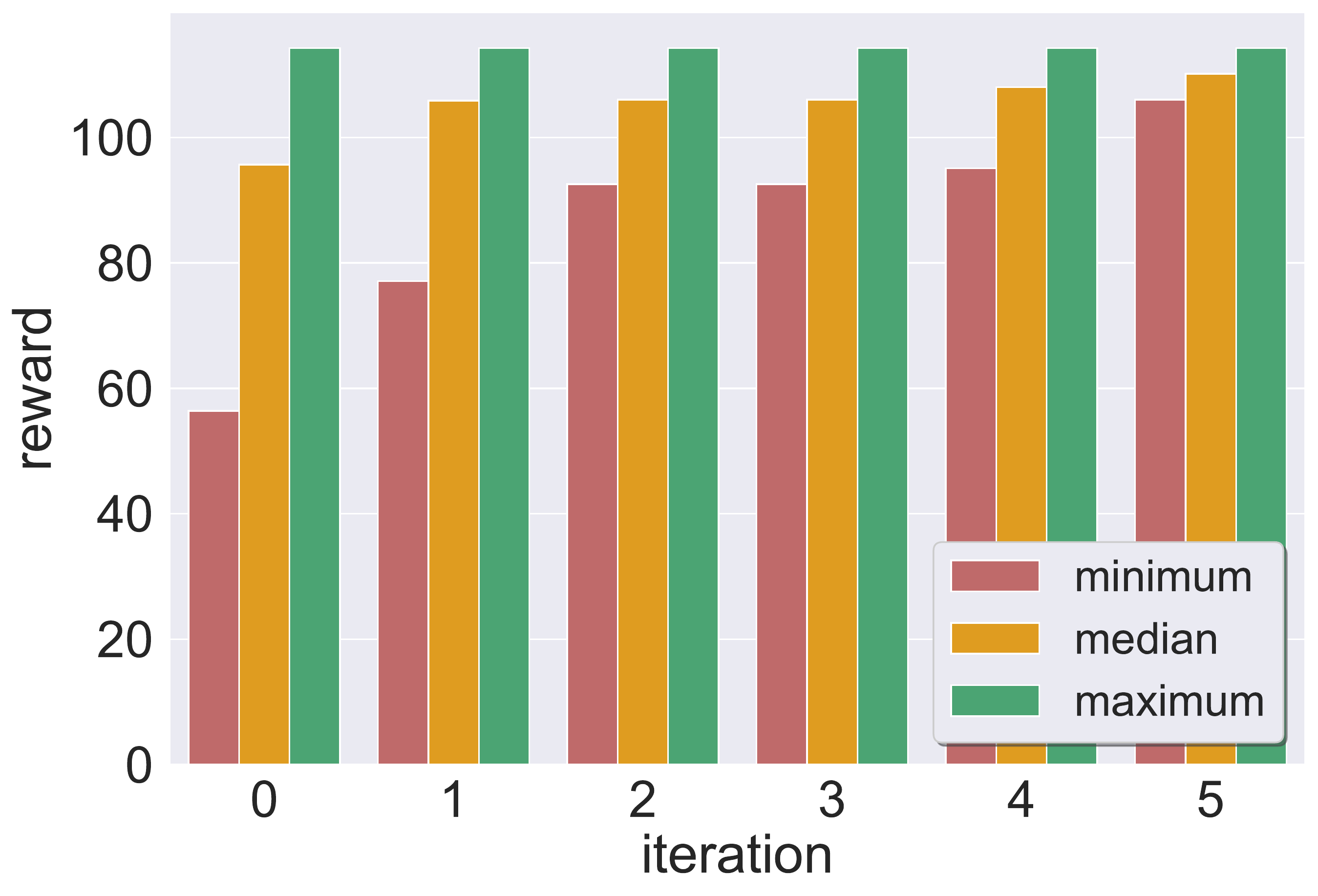}
         \caption{Rewards statistics}
        \label{}
     \end{subfigure}
     \hfill
     \begin{subfigure}[t]{0.49\linewidth}
        \includegraphics[width=\textwidth, height=0.67\textwidth]{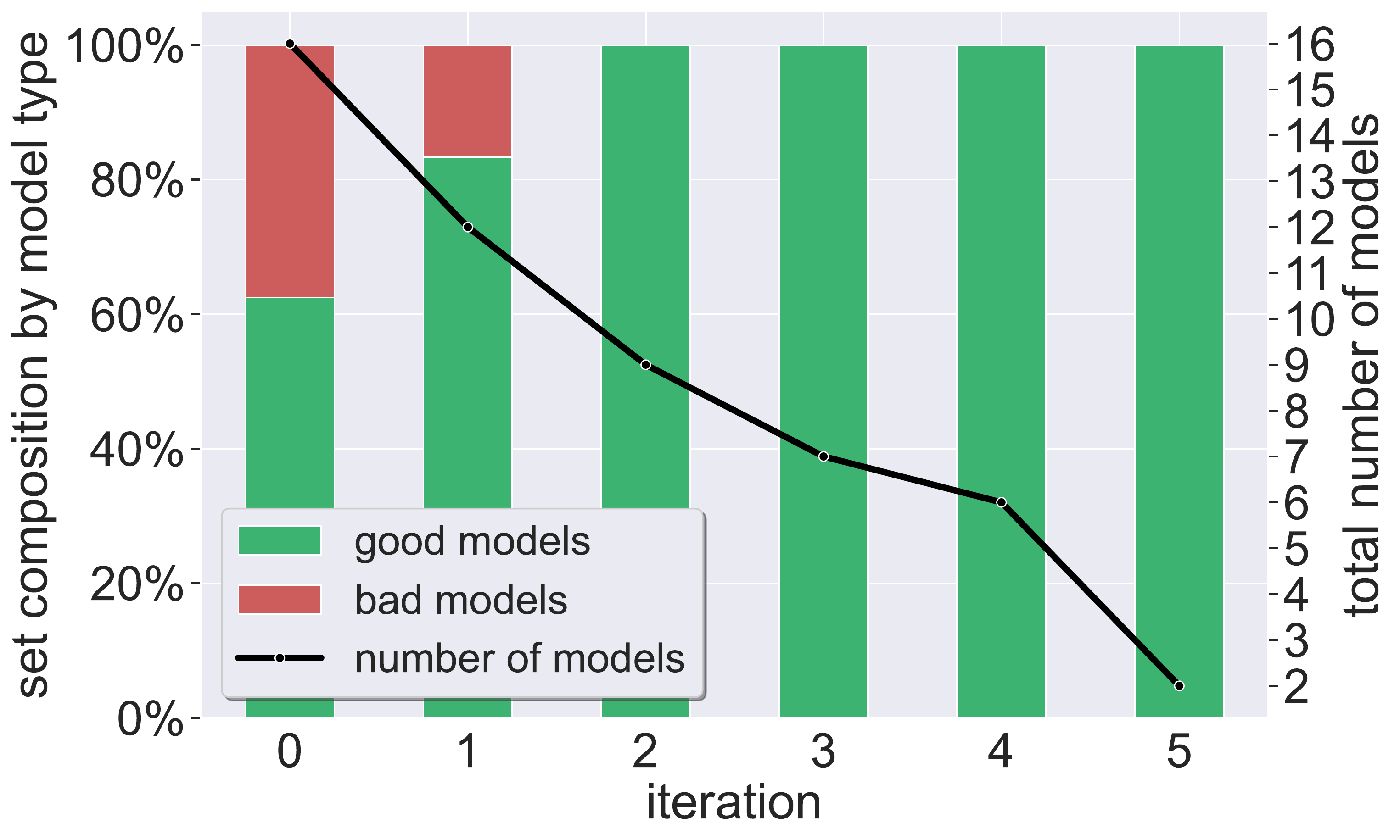}
         \caption{Remaining good and bad models ratio}
        \label{}
     \end{subfigure}
    \caption{Aurora Experiment~\ref{exp:auroraShort}: PDF 
    $\sim\mathcal{TN}(\mu_{low}, \sigma^{2}$): results using the 
    \conditionCombined filtering criterion.}
\end{figure}

\begin{figure}[h]
    \centering
    \captionsetup[subfigure]{justification=centering}
    \captionsetup{justification=centering} 
     \begin{subfigure}[t]{0.49\linewidth}
         \centering
    \includegraphics[width=\textwidth, height=0.67\textwidth]{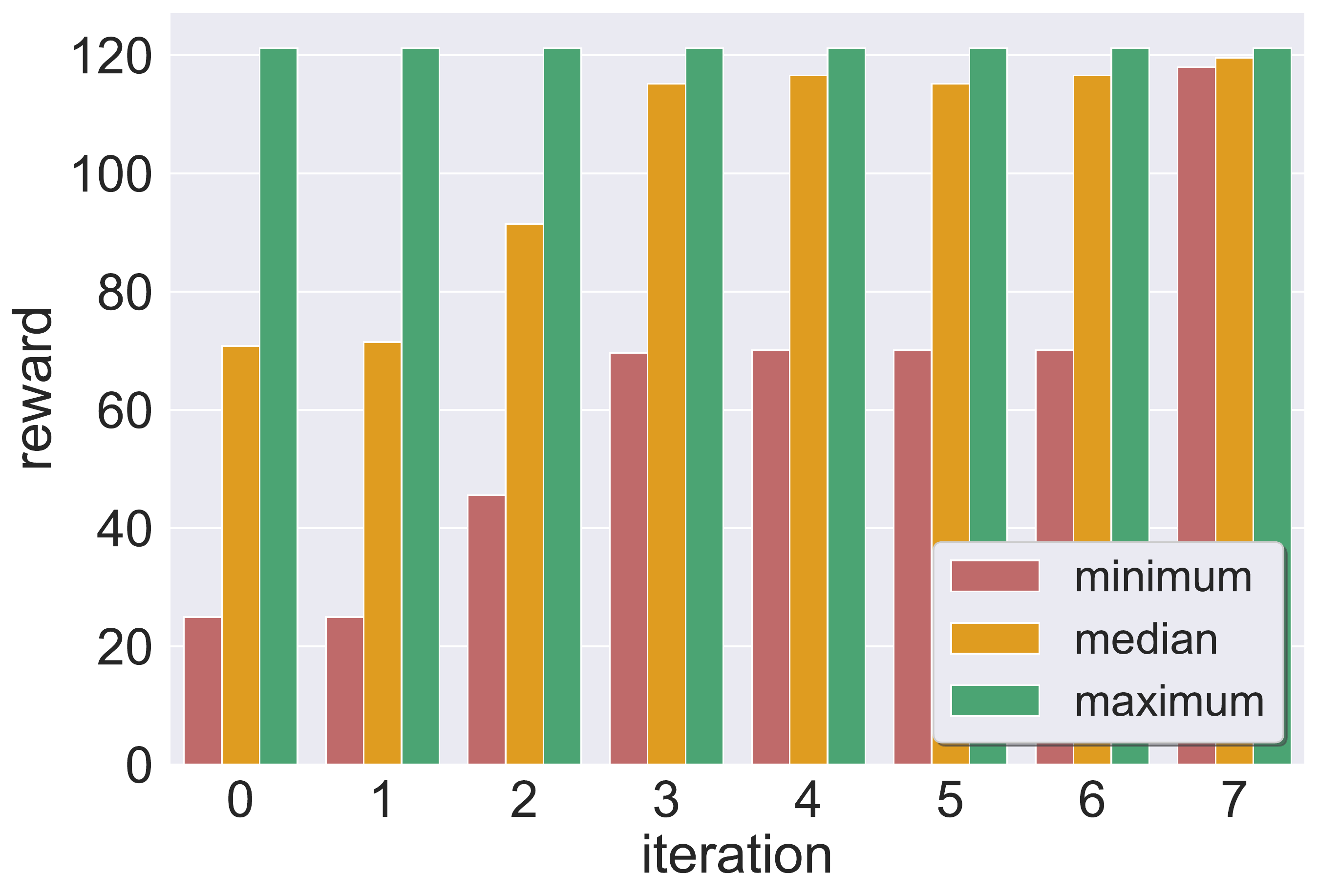}
         \caption{Rewards statistics}
        \label{}
     \end{subfigure}
     \hfill
     \begin{subfigure}[t]{0.49\linewidth}
        \includegraphics[width=\textwidth, height=0.67\textwidth]{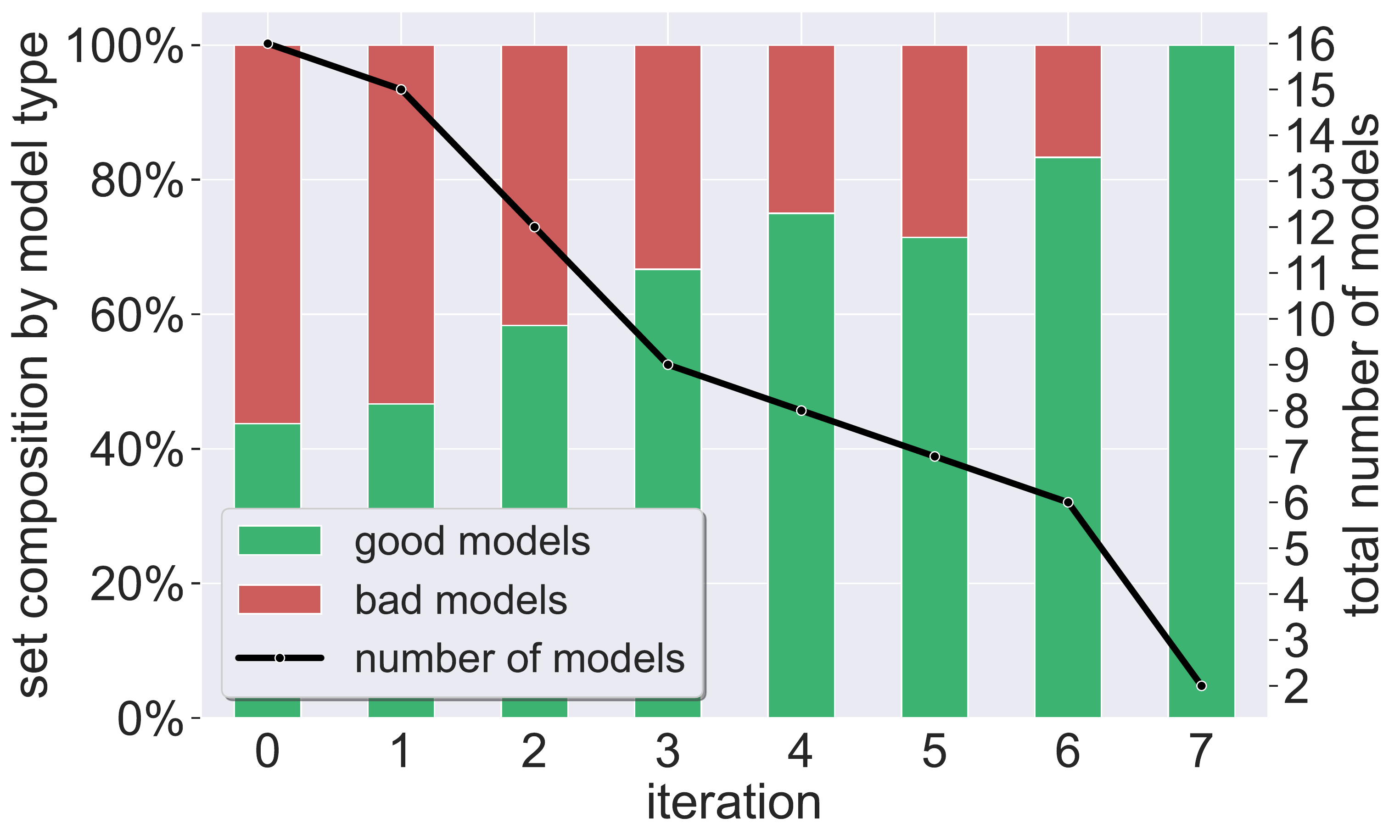}
         \caption{Remaining good and bad models ratio}
        \label{}
     \end{subfigure}
    \caption{Aurora Experiment~\ref{exp:auroraShort}: PDF 
    $\sim\mathcal{TN}(\mu_{high}, \sigma^{2}$): results using the \maxAgg 
    filtering criterion.}
\end{figure}

\begin{figure}[h]
    \centering
    \captionsetup[subfigure]{justification=centering}
    \captionsetup{justification=centering} 
     \begin{subfigure}[t]{0.49\linewidth}
         \centering
    \includegraphics[width=\textwidth, height=0.67\textwidth]{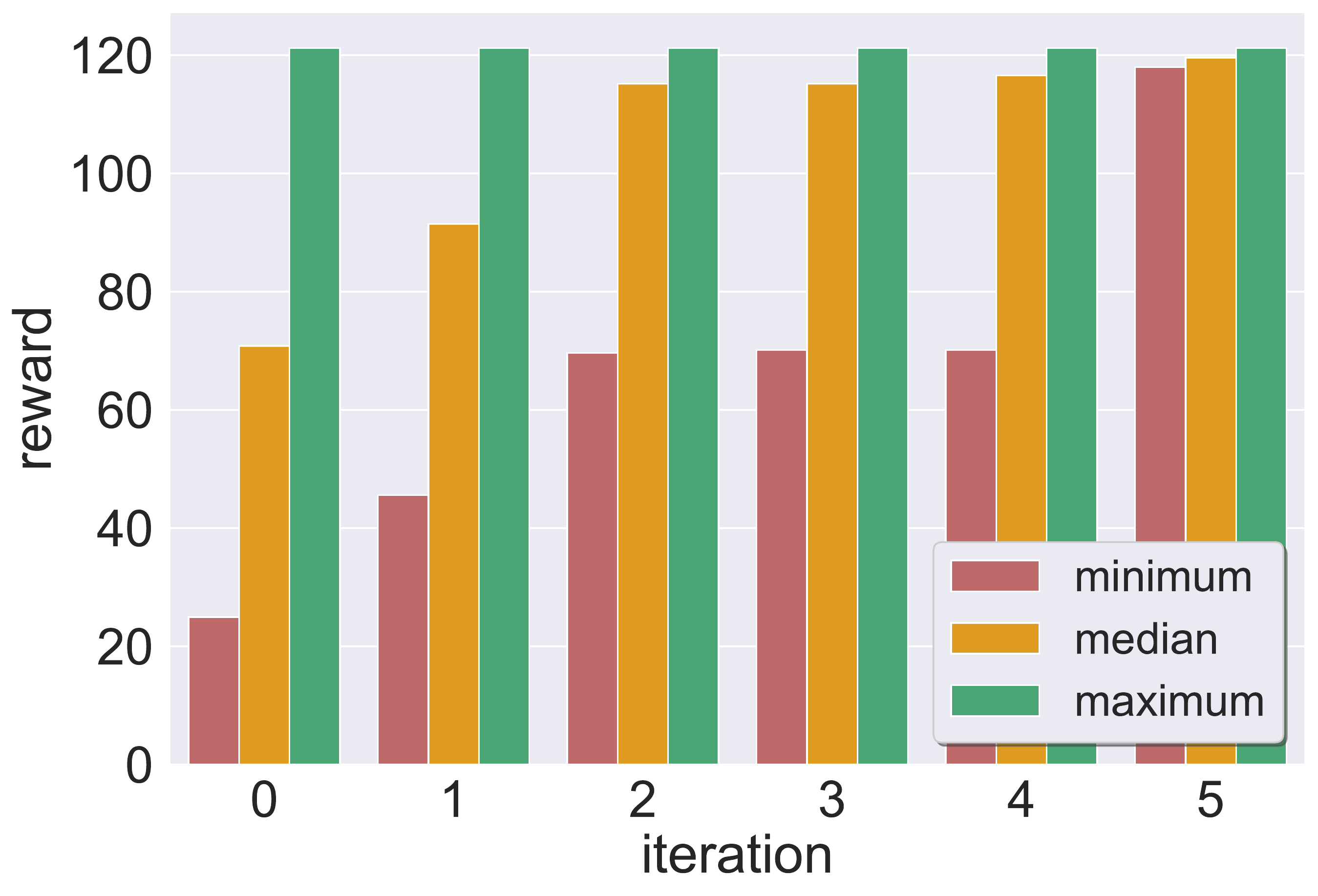}
         \caption{Rewards statistics}
        \label{}
     \end{subfigure}
     \hfill
     \begin{subfigure}[t]{0.49\linewidth}
        \includegraphics[width=\textwidth, height=0.67\textwidth]{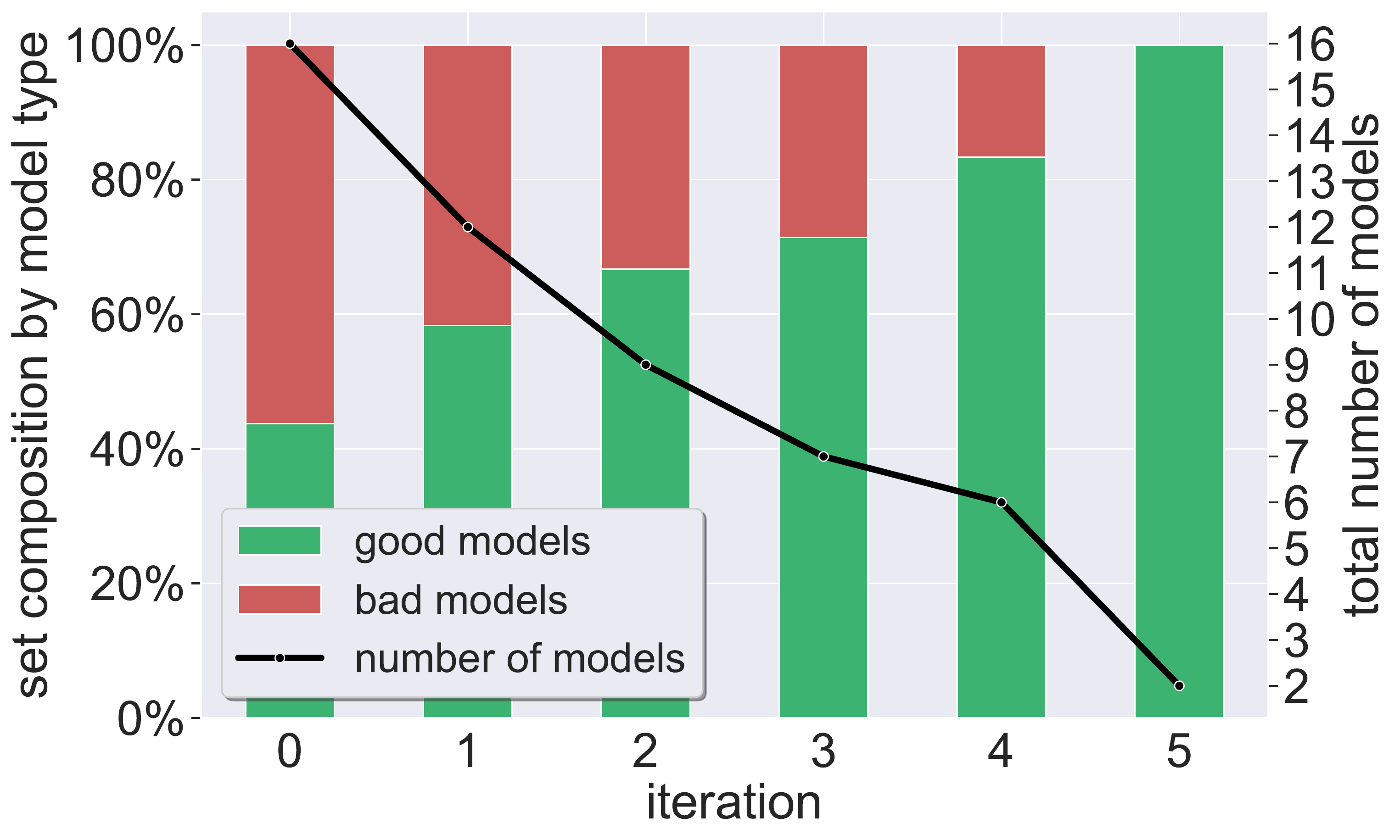}
         \caption{Remaining good and bad models ratio}
        \label{}
     \end{subfigure}
    \caption{Aurora Experiment~\ref{exp:auroraShort}: PDF 
    $\sim\mathcal{TN}(\mu_{high}, \sigma^{2}$): results using the 
    \conditionCombined filtering criterion.}
\end{figure}

\begin{figure}[h]
    \centering
    \captionsetup[subfigure]{justification=centering}
    \captionsetup{justification=centering} 
     \begin{subfigure}[t]{0.49\linewidth}
         \centering
    \includegraphics[width=\textwidth, height=0.67\textwidth]{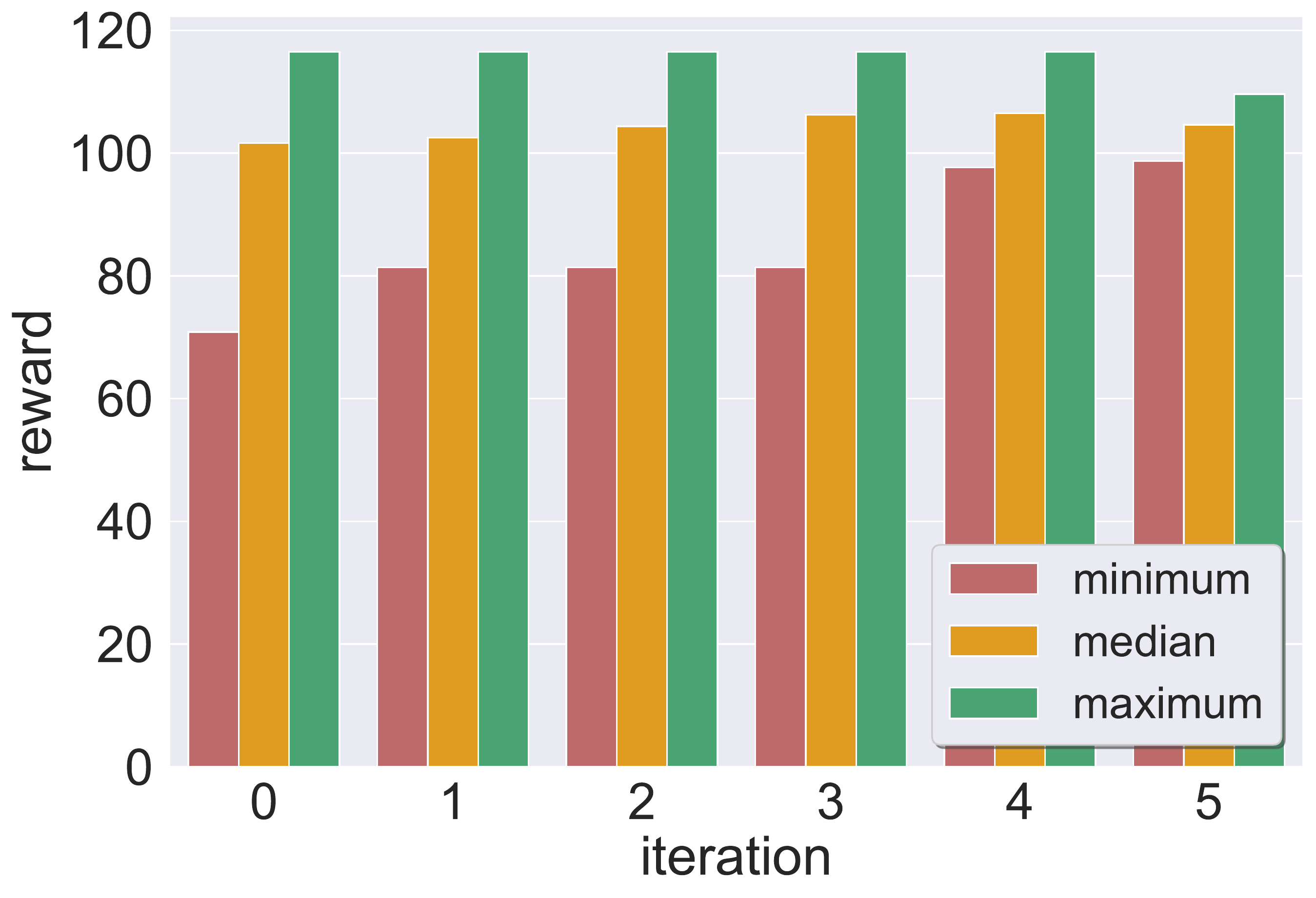}
         \caption{Rewards statistics}
        \label{}
     \end{subfigure}
     \hfill
     \begin{subfigure}[t]{0.49\linewidth}
        \includegraphics[width=\textwidth, height=0.67\textwidth]{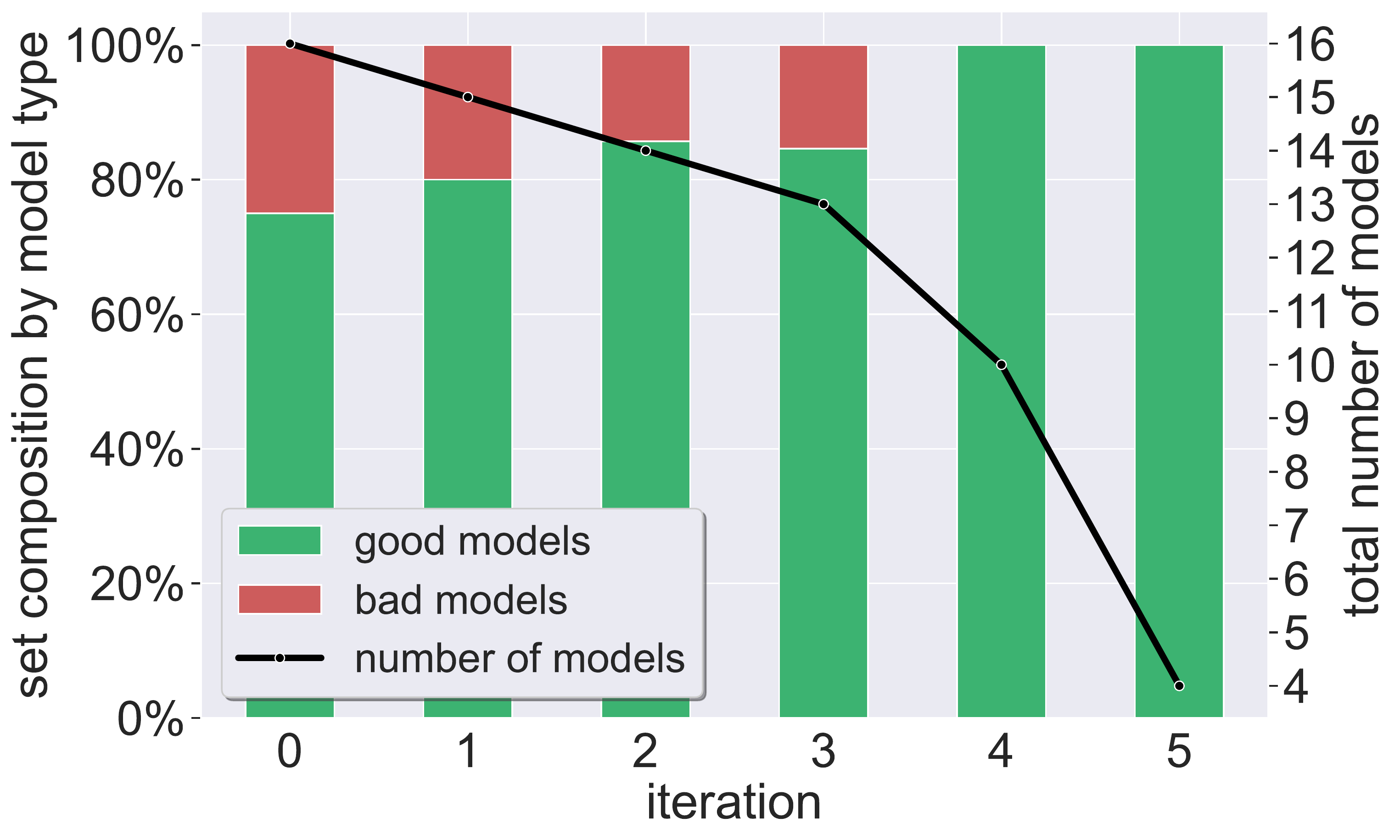}
         \caption{Remaining good and bad models ratio}
        \label{}
     \end{subfigure}
    \caption{Aurora Experiment~\ref{exp:auroraLong}:, PDF 
    $\sim\mathcal{TN}(\mu_{low}, \sigma^{2}$): results using the \maxAgg 
    filtering criterion.}
\end{figure}

\begin{figure}[h]
    \centering
    \captionsetup[subfigure]{justification=centering}
    \captionsetup{justification=centering} 
     \begin{subfigure}[t]{0.49\linewidth}
         \centering
    \includegraphics[width=\textwidth, height=0.67\textwidth]{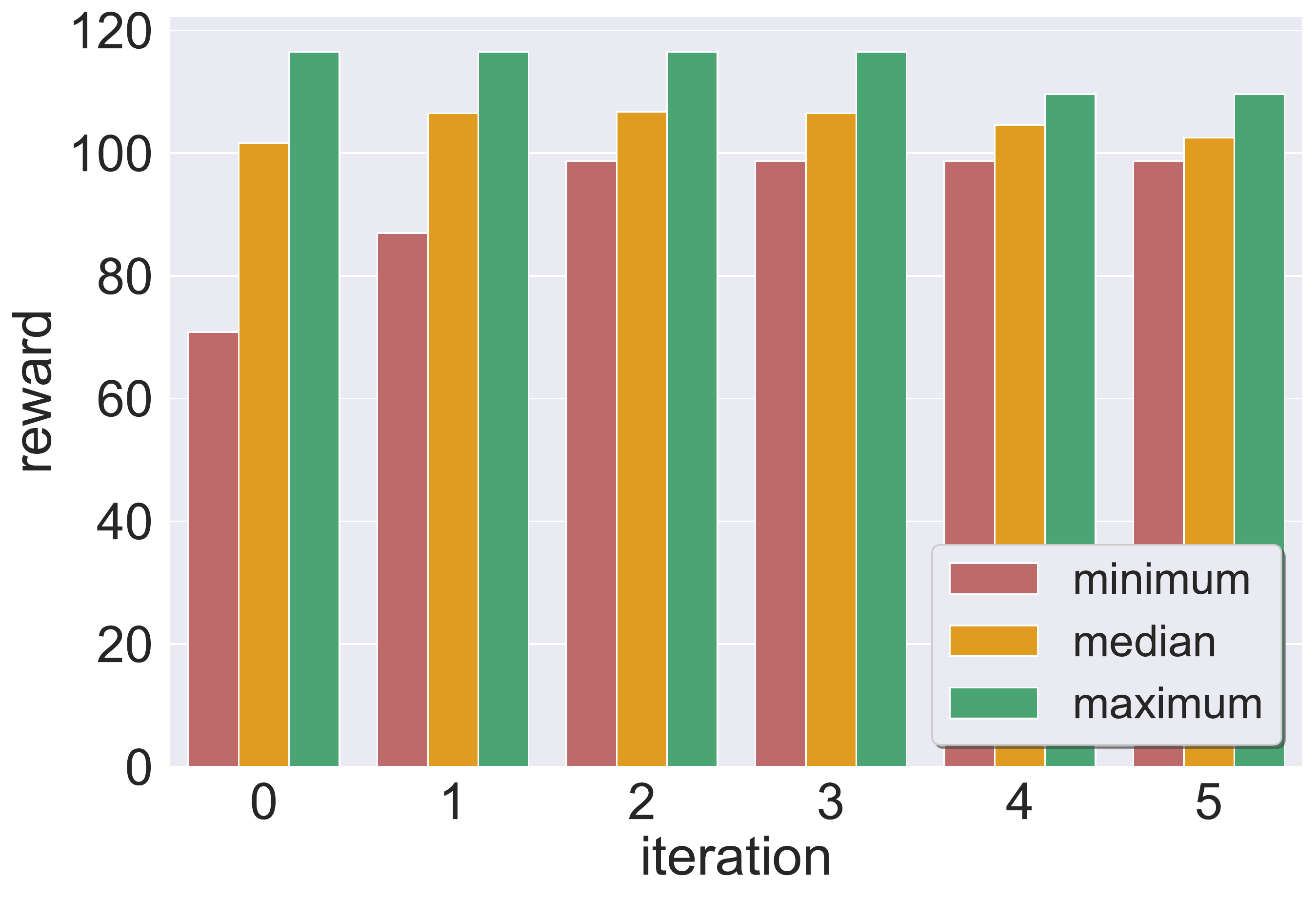}
         \caption{Rewards statistics}
        \label{}
     \end{subfigure}
     \hfill
     \begin{subfigure}[t]{0.49\linewidth}
        \includegraphics[width=\textwidth, height=0.67\textwidth]{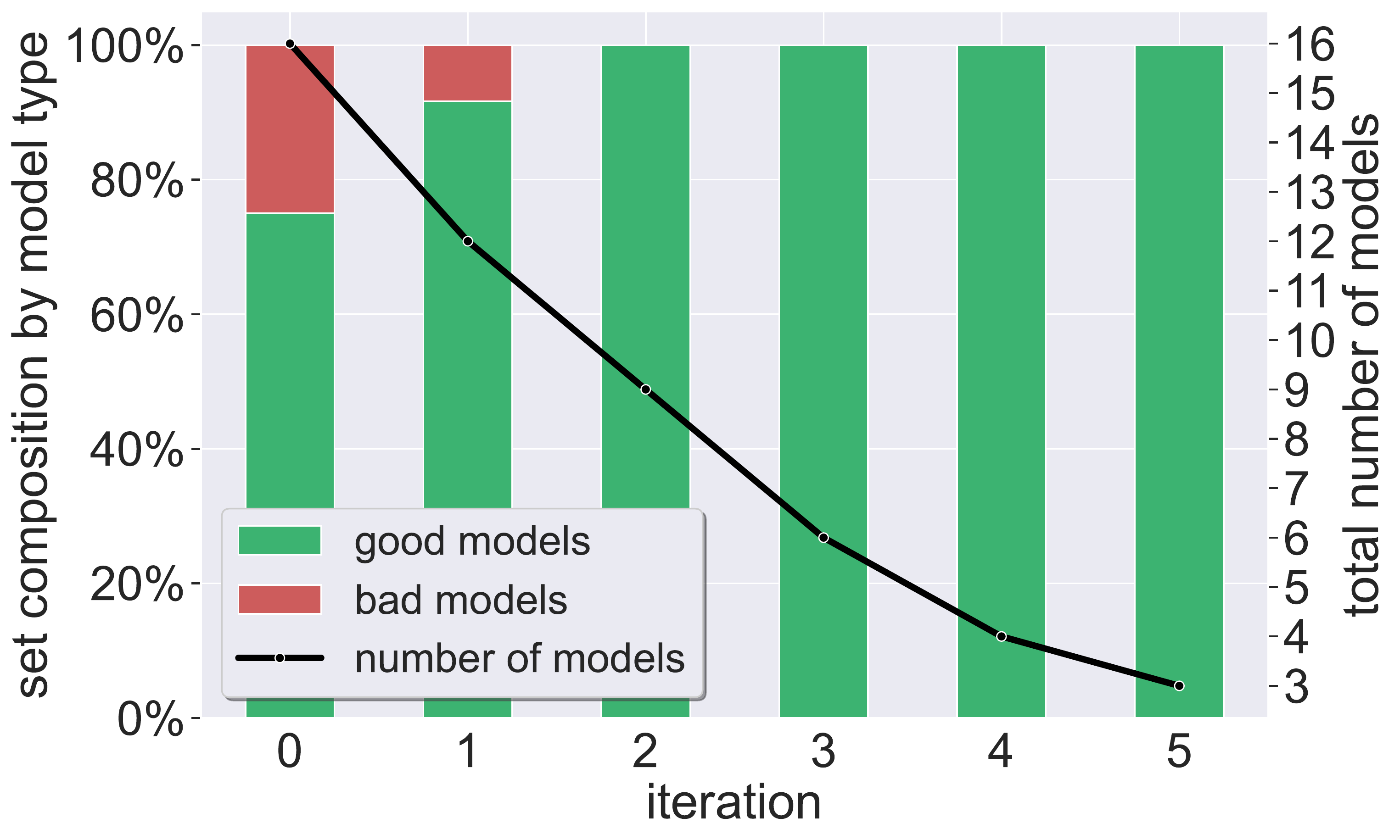}
         \caption{Remaining good and bad models ratio}
        \label{}
     \end{subfigure}
    \caption{Aurora Experiment~\ref{exp:auroraLong}: PDF 
    $\sim\mathcal{TN}(\mu_{low}, \sigma^{2}$): results using the 
    \conditionCombined filtering criterion.}
\end{figure}

\begin{figure}[h]
    \centering
    \captionsetup[subfigure]{justification=centering}
    \captionsetup{justification=centering} 
     \begin{subfigure}[t]{0.49\linewidth}
         \centering
    \includegraphics[width=\textwidth, height=0.67\textwidth]{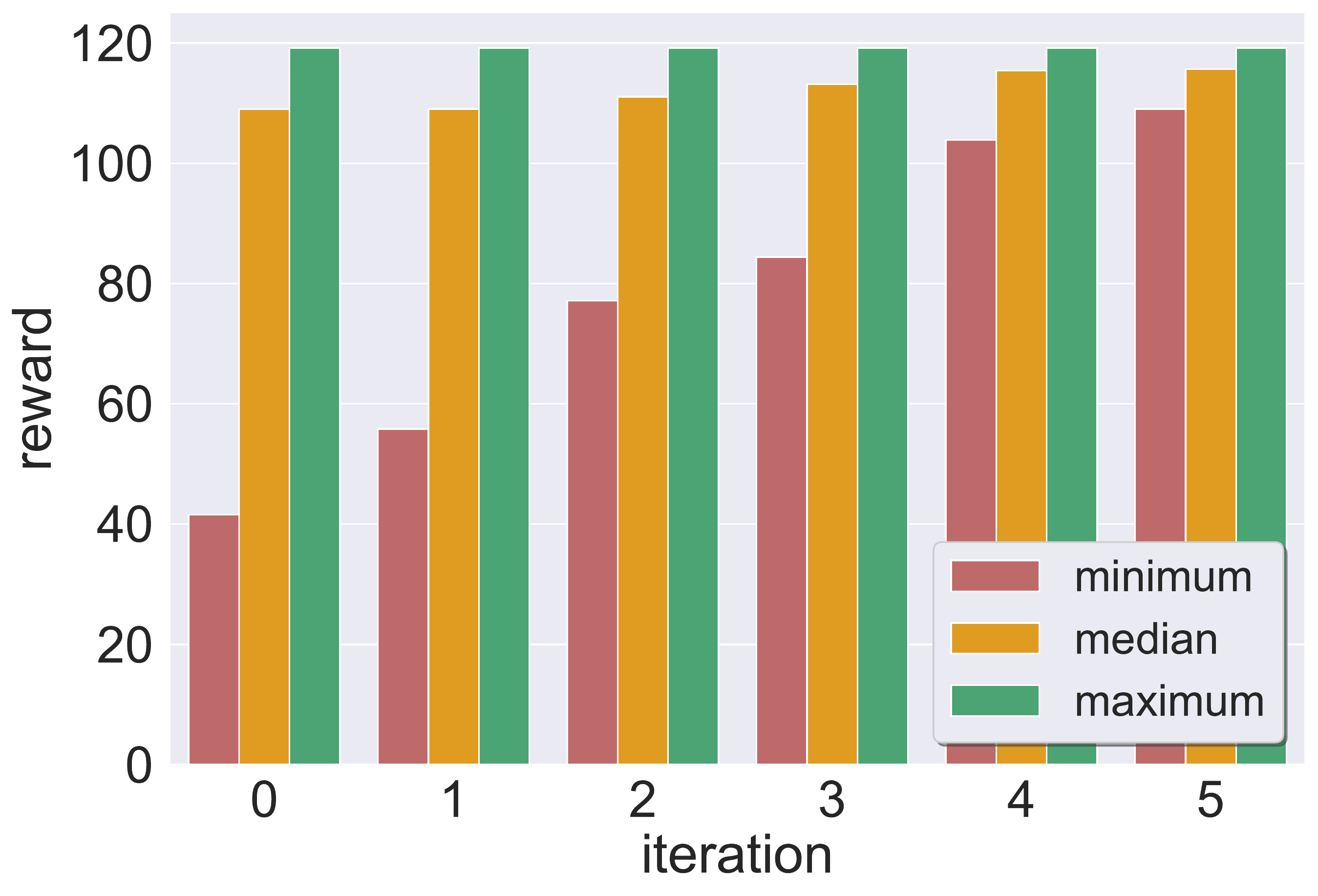}
         \caption{Rewards statistics}
        \label{}
     \end{subfigure}
     \hfill
     \begin{subfigure}[t]{0.49\linewidth}
        \includegraphics[width=\textwidth, height=0.67\textwidth]{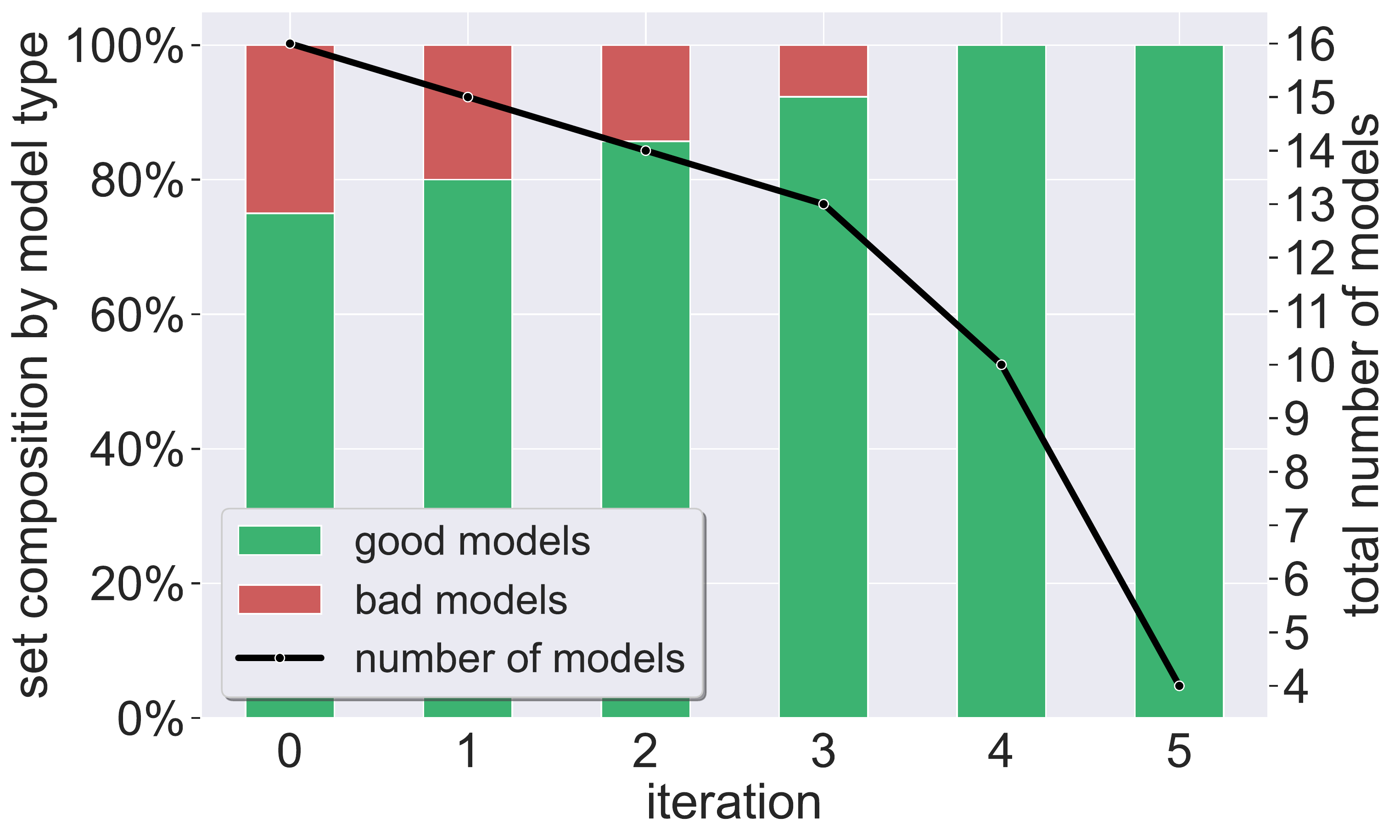}
         \caption{Remaining good and bad models ratio}
        \label{}
     \end{subfigure}
    \caption{Aurora Experiment~\ref{exp:auroraLong}: PDF 
    $\sim\mathcal{TN}(\mu_{high}, \sigma^{2}$): results using the \maxAgg 
    filtering criterion.}
\end{figure}

\begin{figure}[h]
    \centering
    \captionsetup[subfigure]{justification=centering}
    \captionsetup{justification=centering} 
     \begin{subfigure}[t]{0.49\linewidth}
         \centering
    \includegraphics[width=\textwidth, height=0.67\textwidth]{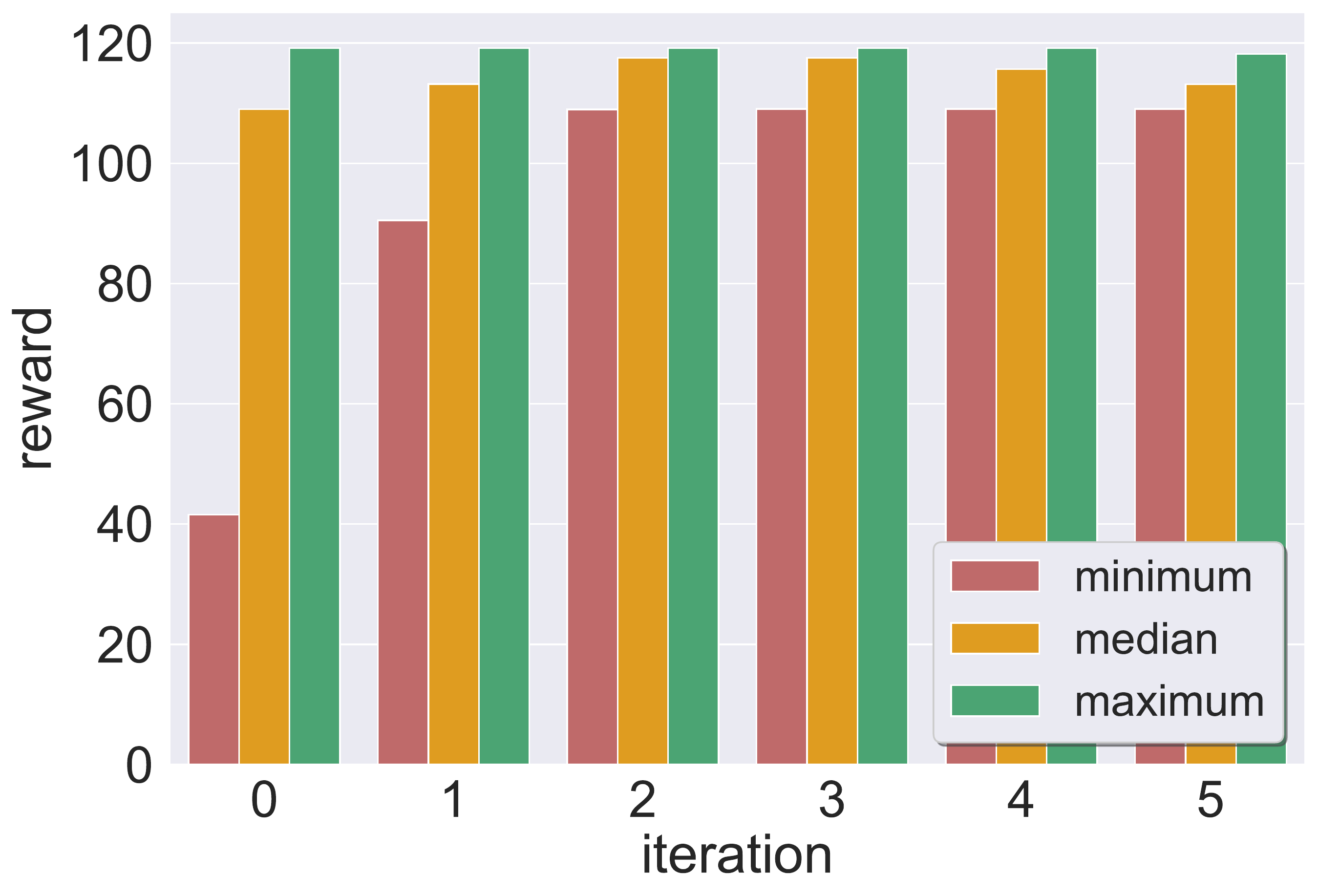}
         \caption{Rewards statistics}
        \label{}
     \end{subfigure}
     \hfill
     \begin{subfigure}[t]{0.49\linewidth}
        \includegraphics[width=\textwidth, height=0.67\textwidth]{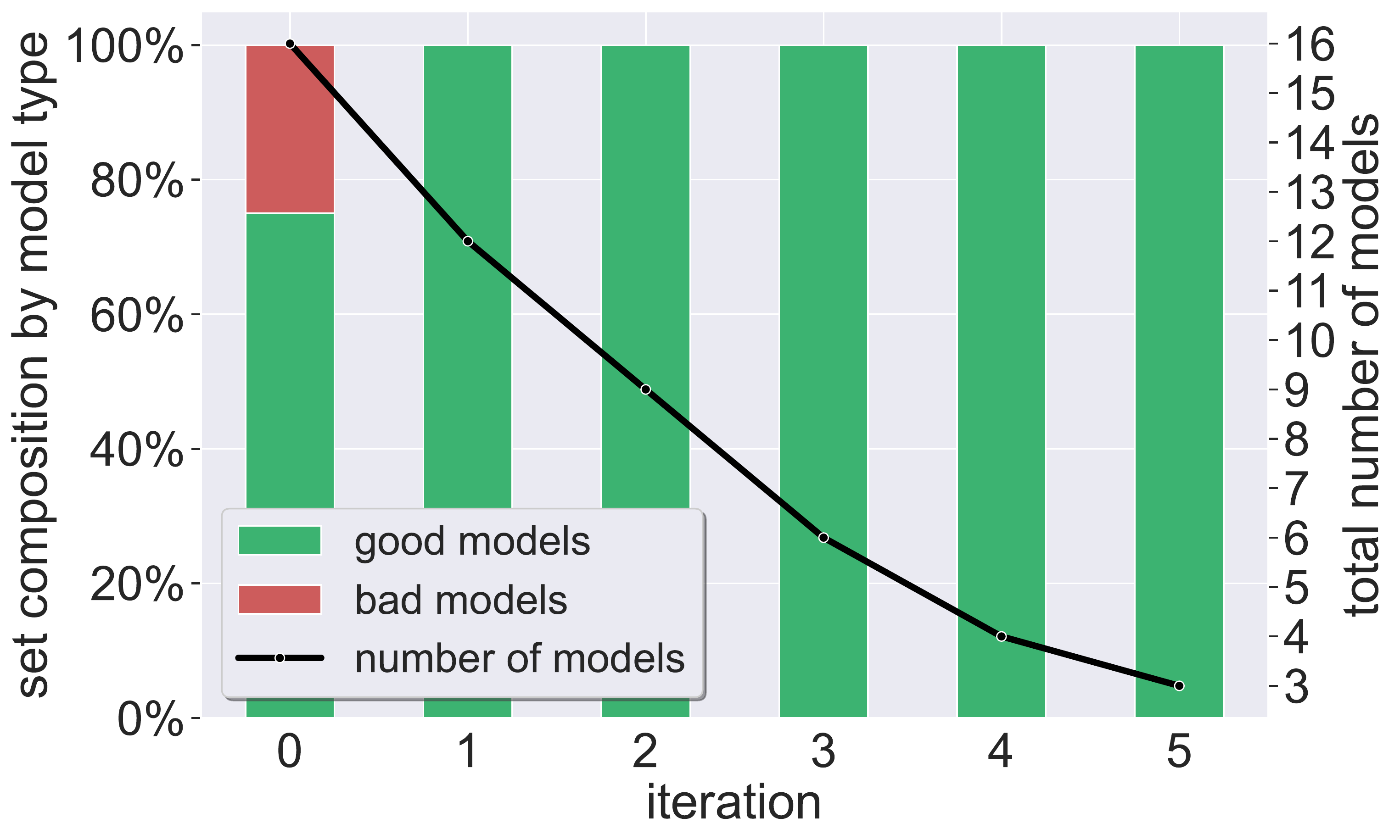}
         \caption{Remaining good and bad models ratio}
        \label{}
     \end{subfigure}
    \caption{Aurora Experiment~\ref{exp:auroraLong}: PDF 
    $\sim\mathcal{TN}(\mu_{high}, \sigma^{2}$): results using the 
    \conditionCombined filtering criterion.}
\end{figure}
\FloatBarrier

\clearpage

\section{Comparison to Gradient-Based Methods}
\label{sec:appendix:gradientAttacks}

\subsection{Introduction}
The methods presented in this paper effectively utilize DNN verification technology (Line~\ref{line:SMTsolverForPdt} in Alg.~\ref{alg:algorithmPairDisagreementScores}) in order to solve an optimization problem: \emph{given a pair of DNNs, a predefined inputs domain, and a distance function, what is the maximal distance between their outputs?} In other words, we use verification to find an input that maximizes the difference between the outputs of two neural networks, under certain constraints. Although verification is highly complex~\cite{KaBaDiJuKo21}, we demonstrate that it is crucial in our setting. Specifically, we show the results of our method when verification is replaced with other techniques. One such technique for optimizing non-convex functions is utilizing gradient-based algorithms (``attacks'') such as Gradient Descent~\cite{Ru16}, Projected Gradient Descent (PGD)~\cite{MaMaScTsVl17}, and others. These methods, however, are not directly suitable for optimizing non-trivial constraints, and hence require modification in order for them to succeed in our scenarios (see ``distance functions'' in Appendix~\ref{sec:appendix:VerificationQueries}). We generated three gradient attacks:
\begin{enumerate}
    \item \emph{gradient attack \# 1}: a non-iterative \textbf{Fast Gradient Sign Method (FGSM)}~\cite{HuPaGoDuAb17} attack, used when optimizing linear constraints (i.e., \emph{$L_{1}$ norm}; see Appendix~\ref{sec:appendix:VerificationQueries}), as in the case of Aurora.

    \item \emph{gradient attack \# 2}: an \textbf{Iterative PGD}~\cite{MaMaScTsVl17} attack, also used when optimizing linear constraints (i.e., \emph{$L_{1}$ norm}; see Appendix~\ref{sec:appendix:VerificationQueries}), as in the case of Aurora. We note that we used this attack in cases when the previous attack failed on linear constraints.

    \item \emph{gradient attack \# 3}: a \textbf{Constrained Iterative 
    PGD}~\cite{MaMaScTsVl17} attack, used in the case when encoding non-linear 
    constraints (i.e., \emph{c-distance} function; see 
    Appendix~\ref{sec:appendix:VerificationQueries}), as in the case of 
    Cartpole and Mountain Car. 
    
\end{enumerate}

Next, we formalize these constrained optimization problems. 

\subsection{Formulation}
Throughout our work, we focus on an output space $\outputSpace=\mathbb{R}$. For 
a shared input domain $\mathcal{D}$ and two neural networks $N_1: \mathcal{D} 
\to \mathbb{R}$ and $N_2: \mathcal{D} \to \mathbb{R}$, for which we wish to 
find an input $\boldsymbol{x}\in\mathcal{D}$ that maximizes the difference 
between the outputs of these neural networks. 

Formally, in the case of the $L_{1}$ norm, we wish to solve the following optimization problem:
\begin{alignat*}{2}
&\underset{x}{\textbf{max}} \quad &&|N_1(\boldsymbol{x}) - N_2(\boldsymbol{x})| \\
&\textbf{s.t.} &&\boldsymbol{x}\in\mathcal{D}
\end{alignat*}

\subsubsection{Gradient Attacks $1$ \& $2$.}
When only input constraints are present, a local maximum can be obtained via conventional gradient methods, maximizing the following objective function:
\[L(\boldsymbol{x}) = |N_1(\boldsymbol{x}) - N_2(\boldsymbol{x})|\]
by taking steps in the direction of its gradient, and projecting them into the domain $\mathcal{D}$, that is:
\begin{align*}
&\boldsymbol{x_0} \in \mathcal{D} \\
&\boldsymbol{x}_{t+1} = [\boldsymbol{x}_t + \epsilon \cdot \nabla_{\boldsymbol{x}} L(\boldsymbol{x}_t)]_{\mathcal{D}}
\end{align*}
Where $[\cdot]_\mathcal{D}: \mathbb{R}^n \to \mathcal{D}$  projects the result 
onto $\mathcal{D}$ and $\epsilon$ is the step size.
we note that $[\cdot]_\mathcal{D}$ may not be trivial to implement, however for $\mathcal{D} \equiv \{\boldsymbol{x}\ |\ x\in\mathbb{R}^n \,\, \forall i\in[n]: l_i \leq x_i \leq u_i \}$ it can be implemented by clipping every coordinate to its appropriate range and $\boldsymbol{x_0}$ can be obtained by taking $\boldsymbol{x_0} = \frac{\boldsymbol{l} + \boldsymbol{u}}{2}$.

For adversarial attacks on DNNs, i.e., maximizing a loss function for a pair of 
DNNs, relative to their input, the FGSM algorithm (\emph{gradient attack \# 1}) 
moves in a single step toward the direction of the gradient. This simple attack 
has been shown~\cite{HuPaGoDuAb17} to be quite efficient for causing 
misclassification. This (projected) FGSM is formalized as follows:
\begin{algorithm}[H]
	\caption{FGSM} 
	\hspace*{\algorithmicindent} \textbf{Input:} objective $L$, variables $\boldsymbol{x}$, input domain $\mathcal{D}$:(\init, \project), step size $\epsilon$ \\
	\hspace*{\algorithmicindent} \textbf{Output:} adversarial input $\boldsymbol{x}$
	\begin{algorithmic}[1]
	    \State $\boldsymbol{x_0} \gets \Call{Init}{\mathcal{D}}$
        \State $\boldsymbol{x}_{\textbf{adv}} \gets \Call{Project}{\boldsymbol{x}_0 + \epsilon \cdot \textbf{sign}(\nabla_{\boldsymbol{x}} L(\boldsymbol{x}_0))}$
		\State \Return{$\boldsymbol{x}_{\textbf{adv}}$}
	\end{algorithmic}
\end{algorithm}

In the context of our algorithms, we define $\mathcal{D}$ by two functions:  
\init which returns an initial value from $\mathcal{D}$ and \project which 
implements $[\cdot]_\mathcal{D}$. 

A more powerful extension of this attack is the PGD algorithm, which we refer 
to as \emph{gradient attack \# 2}. This attack iteratively moves in the 
direction of the gradient, often yielding superior results when compared to its 
single-step counterpart. The attack is formalized as follows:
\begin{algorithm} [H]
	\caption{PGD (maximize)}
	\hspace*{\algorithmicindent} \textbf{Input:} objective $L$, variables $\boldsymbol{x}$, input domain $\mathcal{D}$:(\init, \project), iterations $T$, step size $\epsilon$ \\
	\hspace*{\algorithmicindent} \textbf{Output:} adversarial input $\boldsymbol{x}$
	\begin{algorithmic}[1]
	    \State $\boldsymbol{x_0} \gets \init({\mathcal{D}})$
		\For{$t=0 \ldots T-1$}
            \State $\boldsymbol{x}_{t+1} \gets \project({\boldsymbol{x}_t + 
            \epsilon \cdot \nabla_{\boldsymbol{x}} L(\boldsymbol{x}_t)})$ 
            \Comment{$\textbf{sign}(\nabla_{\boldsymbol{x}} 
            L(\boldsymbol{x}_t))$ may also be used}
		\EndFor
		\State \Return{$\boldsymbol{x}_{T}$}
	\end{algorithmic}
\end{algorithm}
We note that the case for using PGD in order to \emph{minimize} the objective  
function is symmetric.

\subsubsection{Gradient Attack $3$.}
However, in some cases, the gradient attack needs to optimize a loss function 
that represents constraints on the \emph{outputs} of the DNN pairs as well. 
This is the case with the Cartpole and Mountain Car benchmarks, in which we 
used the c-distance function. Specifically, in this scenario, we may need to 
encode constraints of the form:

\begin{align*}
    &N_1(\boldsymbol{x}) \leq 0\\
    &N_2(\boldsymbol{x}) \leq 0
\end{align*}
resulting in the following \emph{constrained} optimization problem:
\begin{alignat*}{2}
& \underset{x}{\textbf{max}} \quad &&|N_1(\boldsymbol{x}) - N_2(\boldsymbol{x})| \\
& \textbf{s.t.} &&\boldsymbol{x}\in\mathcal{D} \\
& &&N_1(\boldsymbol{x}) \leq 0 \\
& &&N_2(\boldsymbol{x}) \leq 0
\end{alignat*}

Conventional gradient attacks are typically not used to solve such optimizations. Hence, we tailored an additional gradient attack (\emph{gradient attack \# 3}) that can efficiently do so by combining our Iterative PGD attack with  \emph{Lagrange Multipliers}~\cite{Ro93} $\boldsymbol{\lambda} \equiv (\lambda^{(1)}, \lambda^{(2)})$ to penalize solutions for which the constraints do not hold. To this end, we defined a novel objective function:
\[
L_{-}(\boldsymbol{x},\boldsymbol{\lambda}) = |N_1(\boldsymbol{x}) - 
N_2(\boldsymbol{x})| - \lambda^{(1)} \cdot \text{ReLU}(N_1(\boldsymbol{x})) - 
\lambda^{(2)} \cdot \text{ReLU}(N_2(\boldsymbol{x}))
\]
resulting in the following optimization problem:
\begin{alignat*}{2}
& \underset{x}{\textbf{max}}  \, \underset{\lambda}{\textbf{min}} \quad &&L_{-}(\boldsymbol{x},\boldsymbol{\lambda}) \\
& \textbf{s.t.} && \boldsymbol{x}\in\mathcal{D} \\
& &&\lambda^{(1)} \geq 0 \\
& &&\lambda^{(2)} \geq 0
\end{alignat*}
and implemented a Constrained Iterative PGD algorithm that approximates a solution:
\begin{algorithm} [H]
	\caption{Constrained Iterative PGD}
	\hspace*{\algorithmicindent} \textbf{Input:} objective $L$, input domain $\mathcal{D}$, constraints: $C_i(\boldsymbol{x})$, iterations: $T, T_x, T_\lambda$, step sizes: $\epsilon_x, \epsilon_\lambda$ \\
	\hspace*{\algorithmicindent} \textbf{Output:} adversarial input $\boldsymbol{x}$
	\begin{algorithmic}[H]
	    \State $\boldsymbol{x_0} \gets \Call{Init}{\mathcal{D}}$
        \State $L_{C}(\boldsymbol{x},\boldsymbol{\lambda}) \equiv 
        L(\boldsymbol{x}) - \sum^k_{i=0}\lambda^{(i)} \cdot 
        \text{ReLU}(C_i(\boldsymbol{x}))$ \Comment{the new objective}
		\For{$t=0 \ldots T-1$}
            \State $\boldsymbol{\lambda}_{t+1} \gets \Call{PGD\_min}{L_C, \boldsymbol{\lambda}, ( \boldsymbol{\lambda} \gets 0, \boldsymbol{\lambda} \geq 0), T_{\lambda}, \epsilon_{\lambda}}$ \Comment{minimize $L_C(\boldsymbol{x}_t, \boldsymbol{\lambda})$ with $\boldsymbol{x}_t$ as constant}
            \State $\boldsymbol{x}_{t+1} \gets \Call{PGD\_max}{L_C, \boldsymbol{x}, \mathcal{D}, T_{x}, \epsilon_{x}}$ \Comment{maximize $L_C(\boldsymbol{x}, \boldsymbol{\lambda}_t)$ with $\boldsymbol{\lambda}_t$ as constant}
		\EndFor
		\State \Return{$\boldsymbol{x}_{T}$}
	\end{algorithmic}
\end{algorithm}

\subsection{Results}
\begin{figure}[ht]
    \centering
    \captionsetup[subfigure]{justification=centering}
    \captionsetup{justification=centering} 
     \begin{subfigure}[t]{0.49\linewidth}
         \centering
    \includegraphics[width=\textwidth]{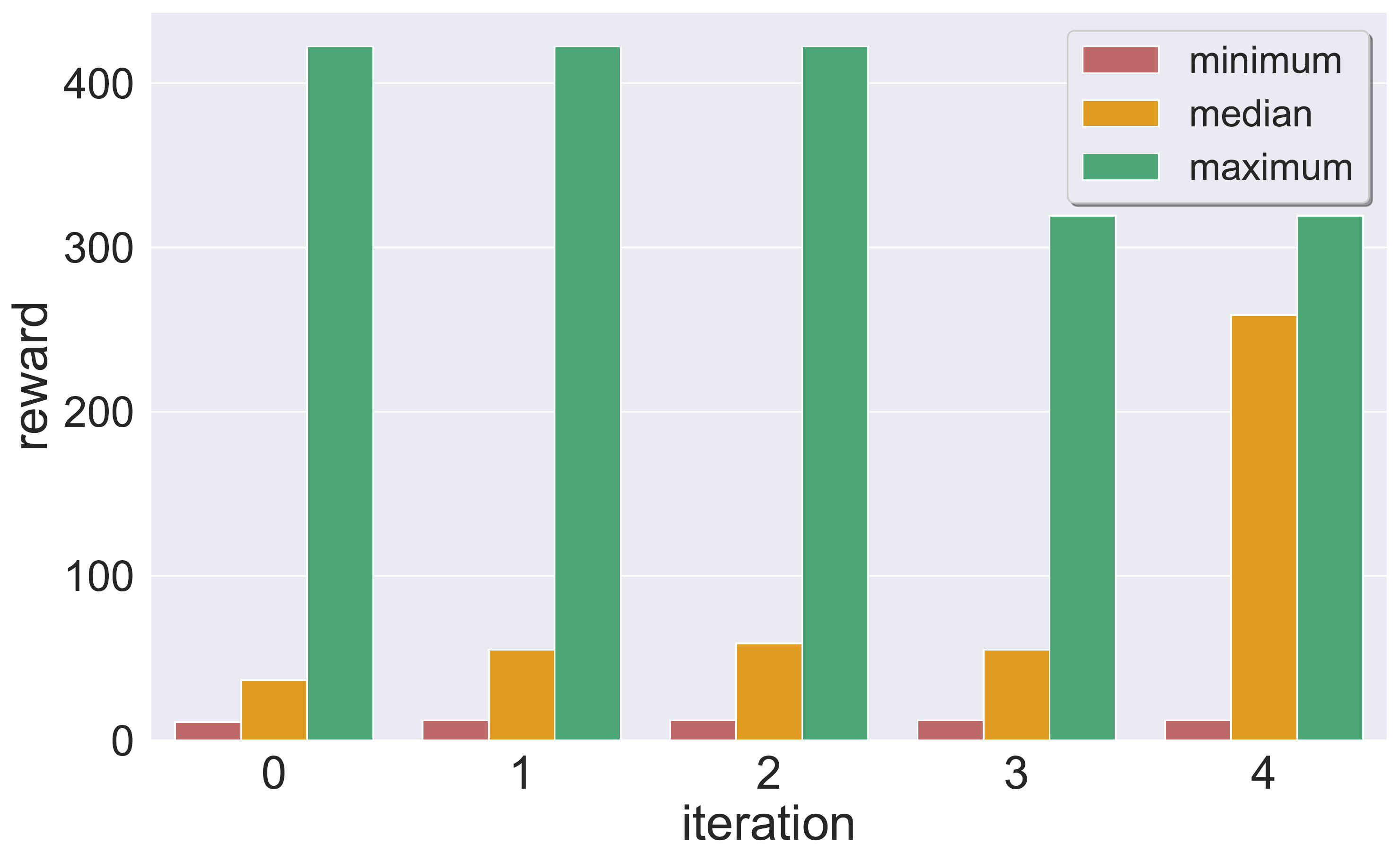}
     \end{subfigure}
     \hfill
     \begin{subfigure}[t]{0.49\linewidth}
        \includegraphics[width=\textwidth]{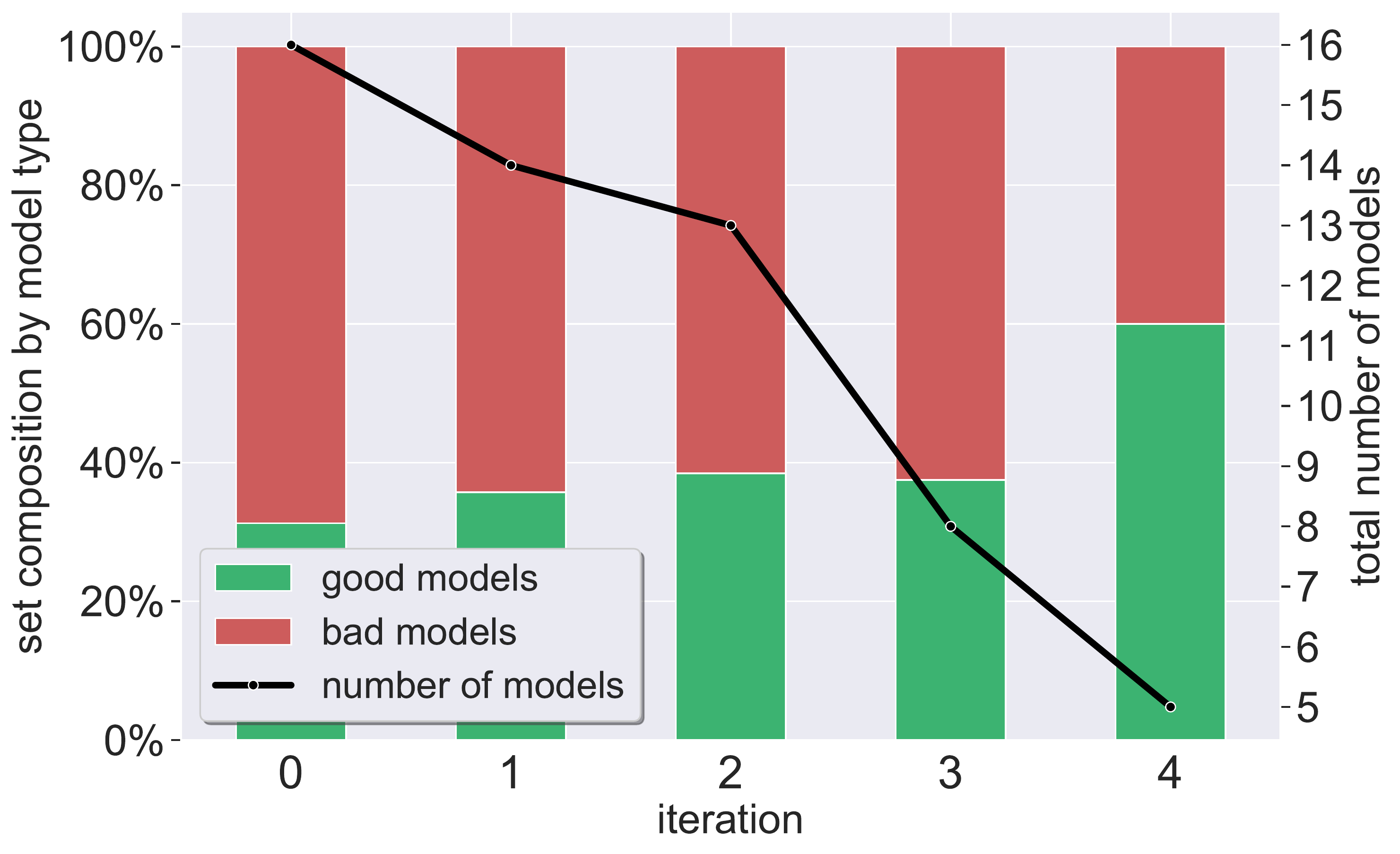}
     \end{subfigure}

     \begin{subfigure}[t]{0.49\linewidth}
         \centering
    \includegraphics[width=\textwidth]{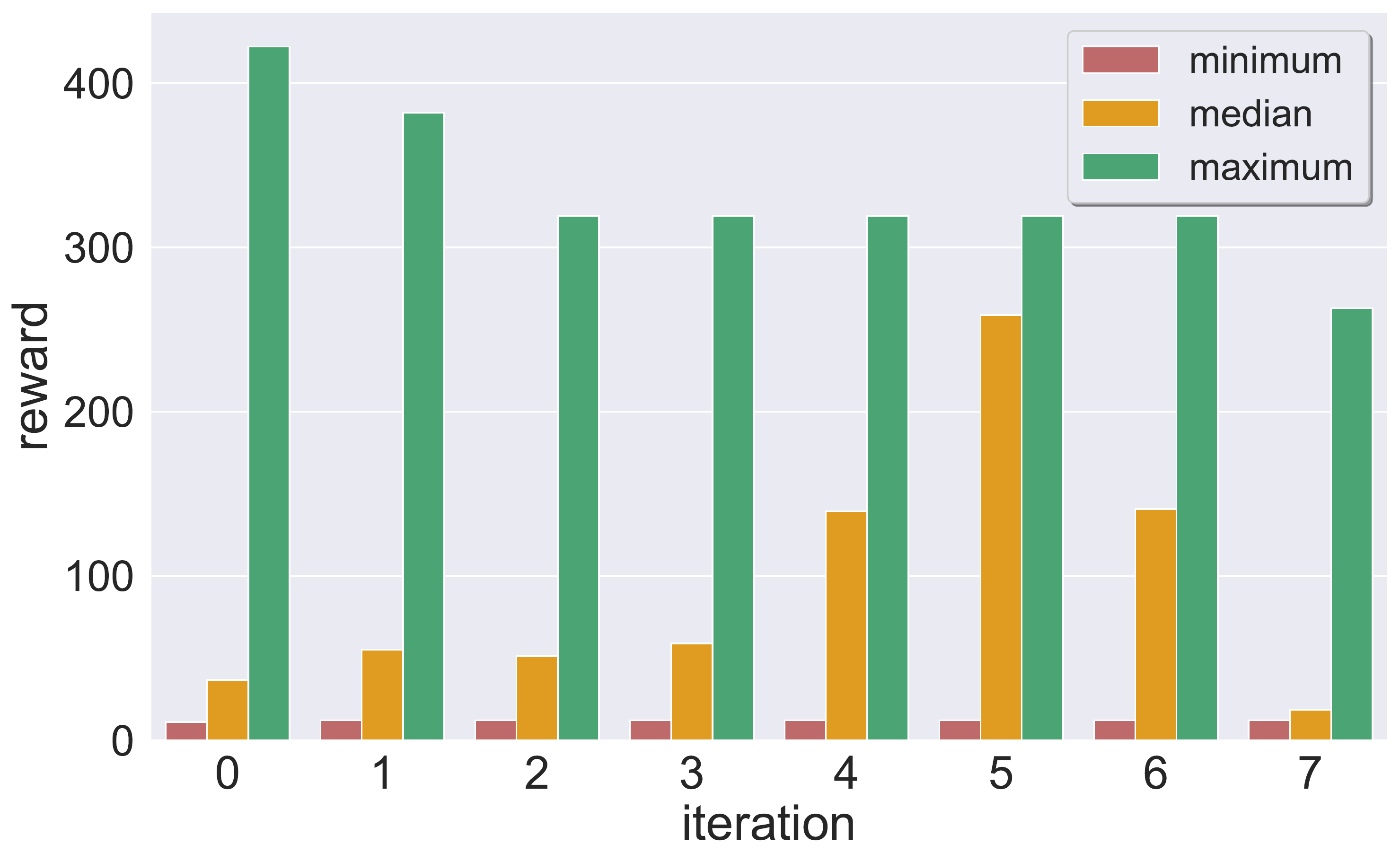}

     \end{subfigure}
     \hfill
     \begin{subfigure}[t]{0.49\linewidth}
        \includegraphics[width=\textwidth]{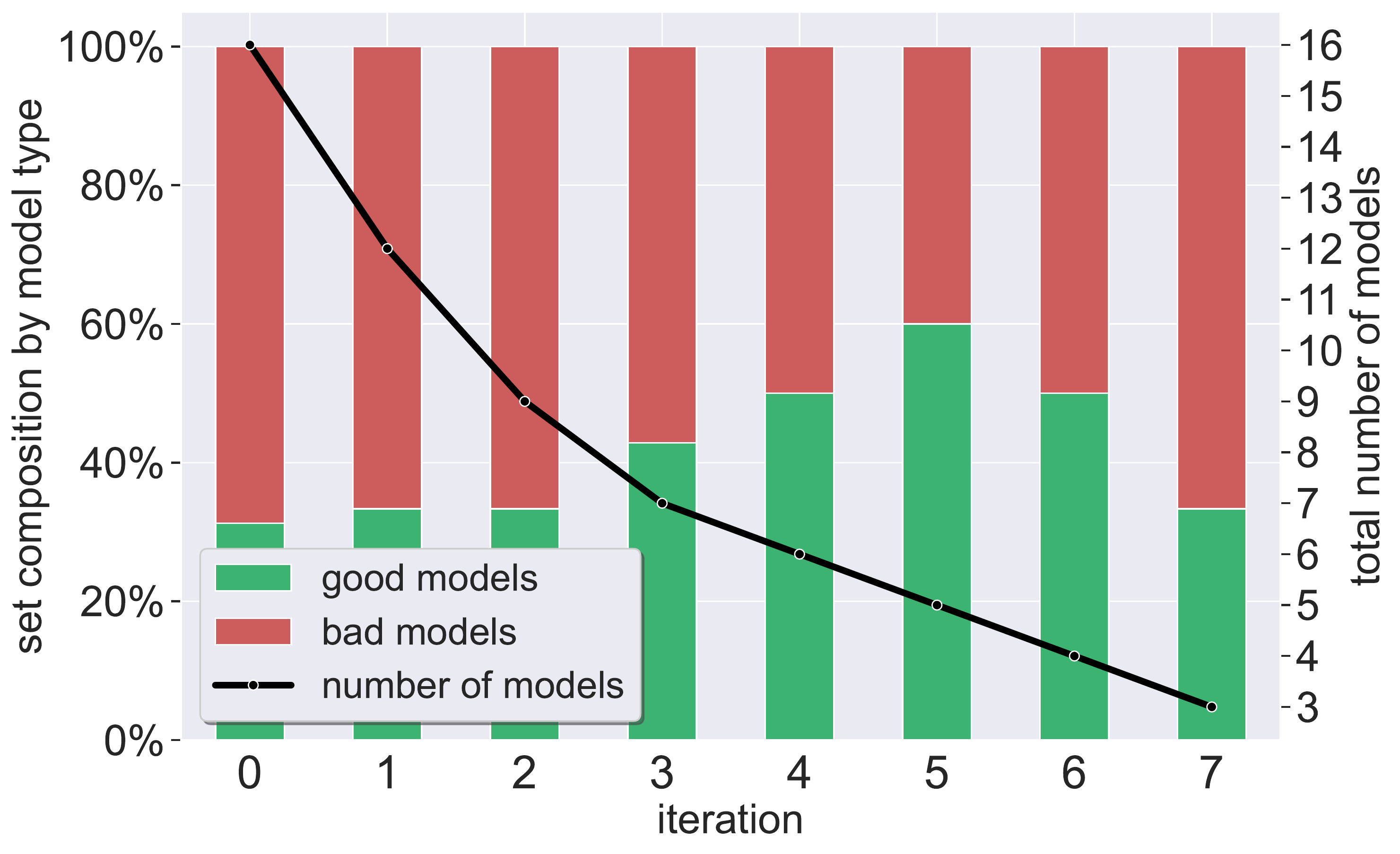}
     \end{subfigure}

    \begin{subfigure}[t]{0.49\linewidth}
         \centering
    \includegraphics[width=\textwidth]{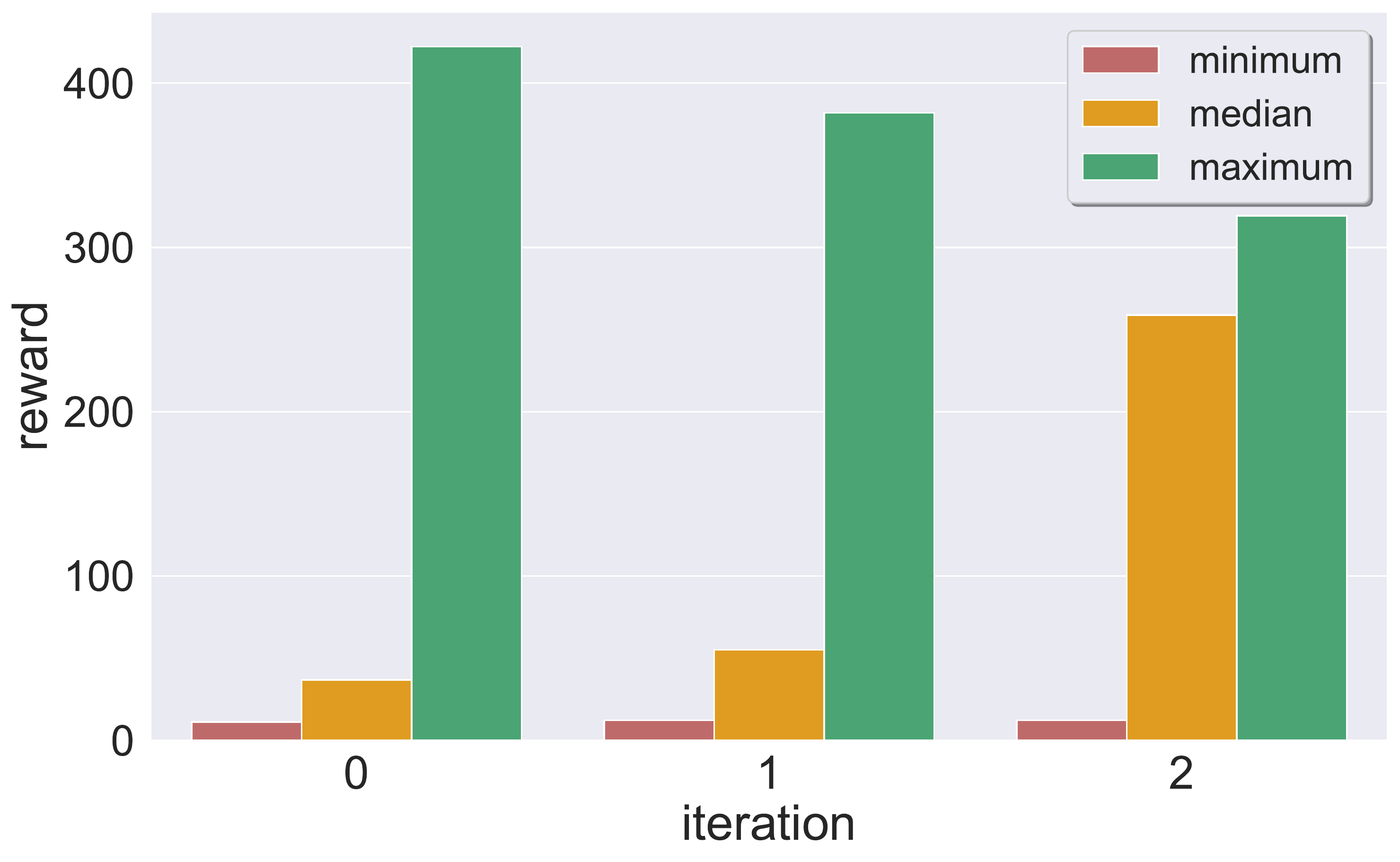}
    \caption{Rewards statistics}

     \end{subfigure}
     \hfill
     \begin{subfigure}[t]{0.49\linewidth}
        \includegraphics[width=\textwidth]{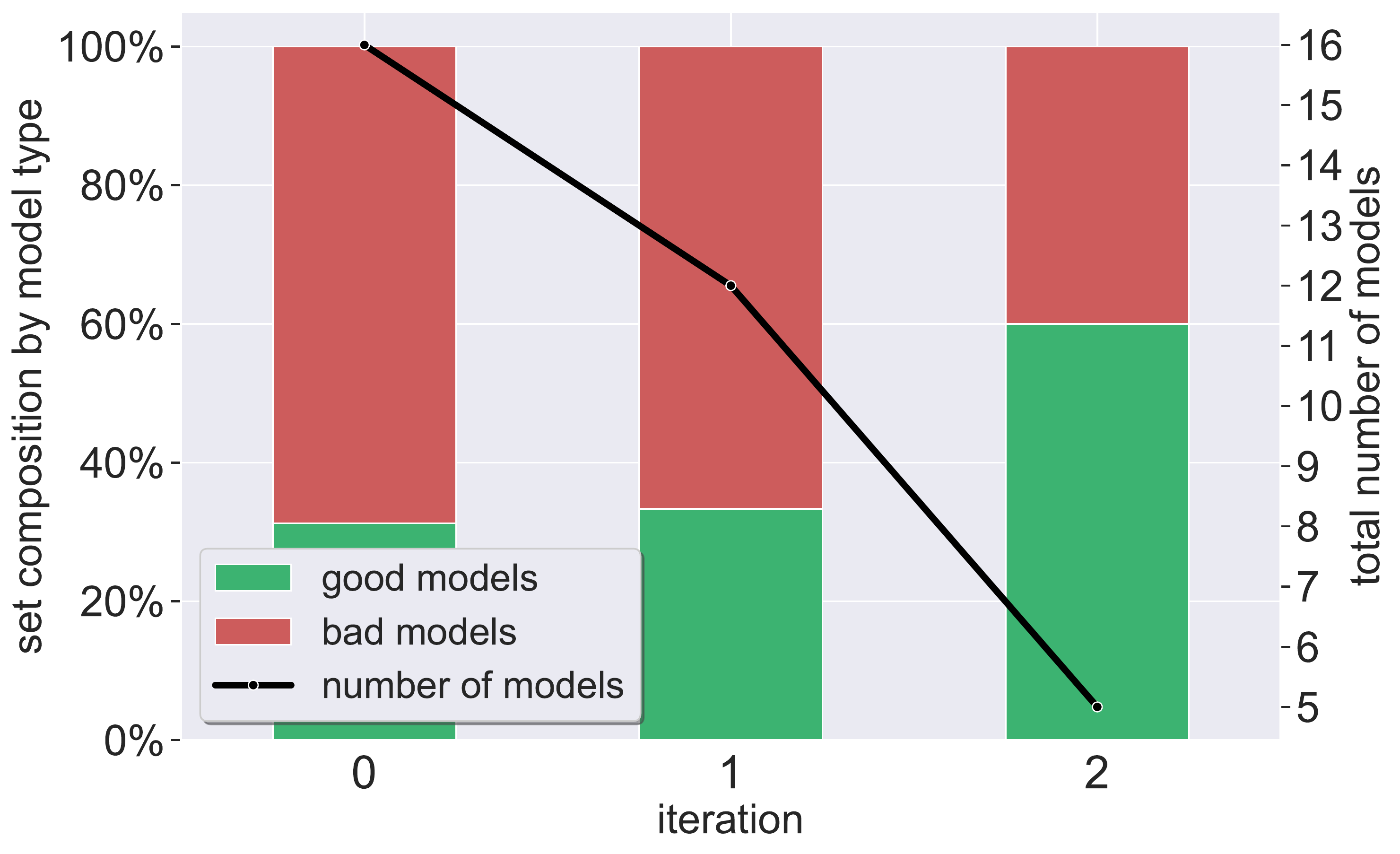}
         \caption{Remaining good and bad models ratio}
        \label{fig:gradientAttacks:cartPole:goodBadRatio}
     \end{subfigure}
    
    \caption{Cartpole: attack \# 3 (Constrained Iterative PGD): Results of 
    models filtered using \emph{PDT} scores approximated by \emph{gradient 
    attacks} (instead of a verification engine) on the Cartpole models.\\Each 
    row, from top to bottom, contains results using a different filtering 
    criterion (and terminating in advance if the disagreement scores are no 
    larger than $2$): \percentileAgg (compare 
    to Fig.~\ref{fig:cartpolePercentileGoodBadResults} and 
    Fig.~\ref{fig:cartpolePercentileMinMaxRewards}), \maxAgg (compare to 
    Fig.~\ref{fig:cartpole:maxResults}), and \conditionCombined (compare to 
    Fig.~\ref{fig:cartpole:CombinedResults}). \\ In all cases, the algorithm 
    returns at least one bad model (and usually more than one), resulting in 
    models with lower average rewards than our verification-based results.}
    \label{fig:cartPoleGradienAttack3}
\end{figure}

\begin{figure}[htpb]
    \centering
    \captionsetup[subfigure]{justification=centering}
    \captionsetup{justification=centering} 
     \begin{subfigure}[t]{0.49\linewidth}
         \centering
    \includegraphics[width=\textwidth, height=0.67\textwidth]{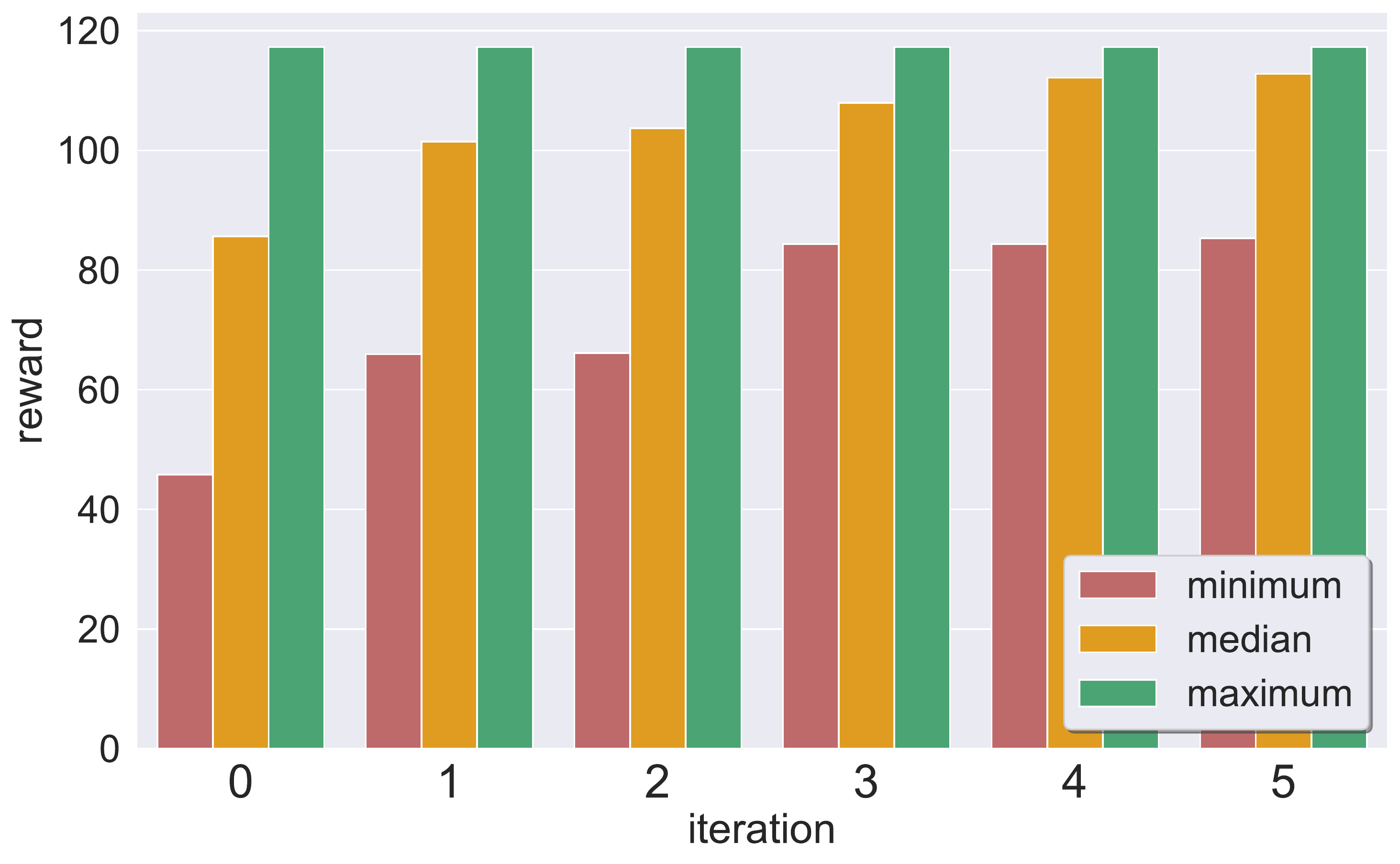}
    \caption{Rewards statistics}
    \label{fig:gradientAttacksSingleStep:aurora:rewardsStats}
     \end{subfigure}
     \hfill
     \begin{subfigure}[t]{0.49\linewidth}
        \includegraphics[width=\textwidth, height=0.67\textwidth]{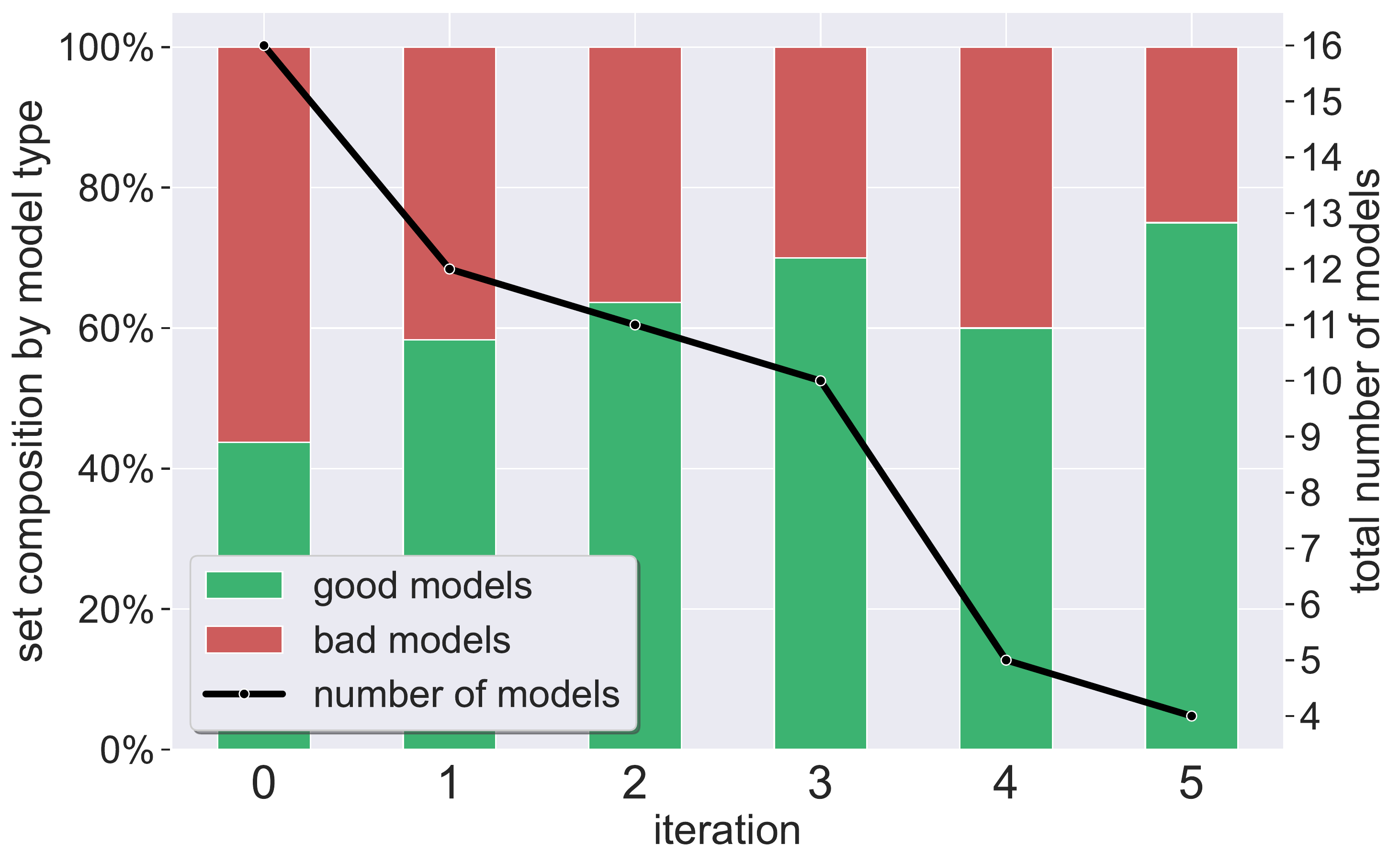}
         \caption{Remaining good and bad models ratio}
        \label{fig:gradientAttacksSingleStep:aurora:goodBadRatio}
     \end{subfigure}
    \caption{Aurora: attack \# 1 (single-step FGSM): Results of models filtered 
    using \emph{PDT} scores approximated by \emph{gradient attacks} (instead of 
    a verification engine) on short-trained Aurora models, using the \maxAgg 
    criterion (and terminating in advance if the disagreement scores are no 
    larger than $2$). In contrast to our verification-driven approach, the 
    final result contains a bad model. Compare to 
    Fig.~\ref{fig:auroraShortMaxFiltering}.}
\label{fig:auroraGradienAttack1}
\end{figure}

\begin{figure}[htpb]
    \centering
    \captionsetup[subfigure]{justification=centering}
    \captionsetup{justification=centering} 
     \begin{subfigure}[t]{0.49\linewidth}
         \centering
    \includegraphics[width=\textwidth, height=0.67\textwidth]{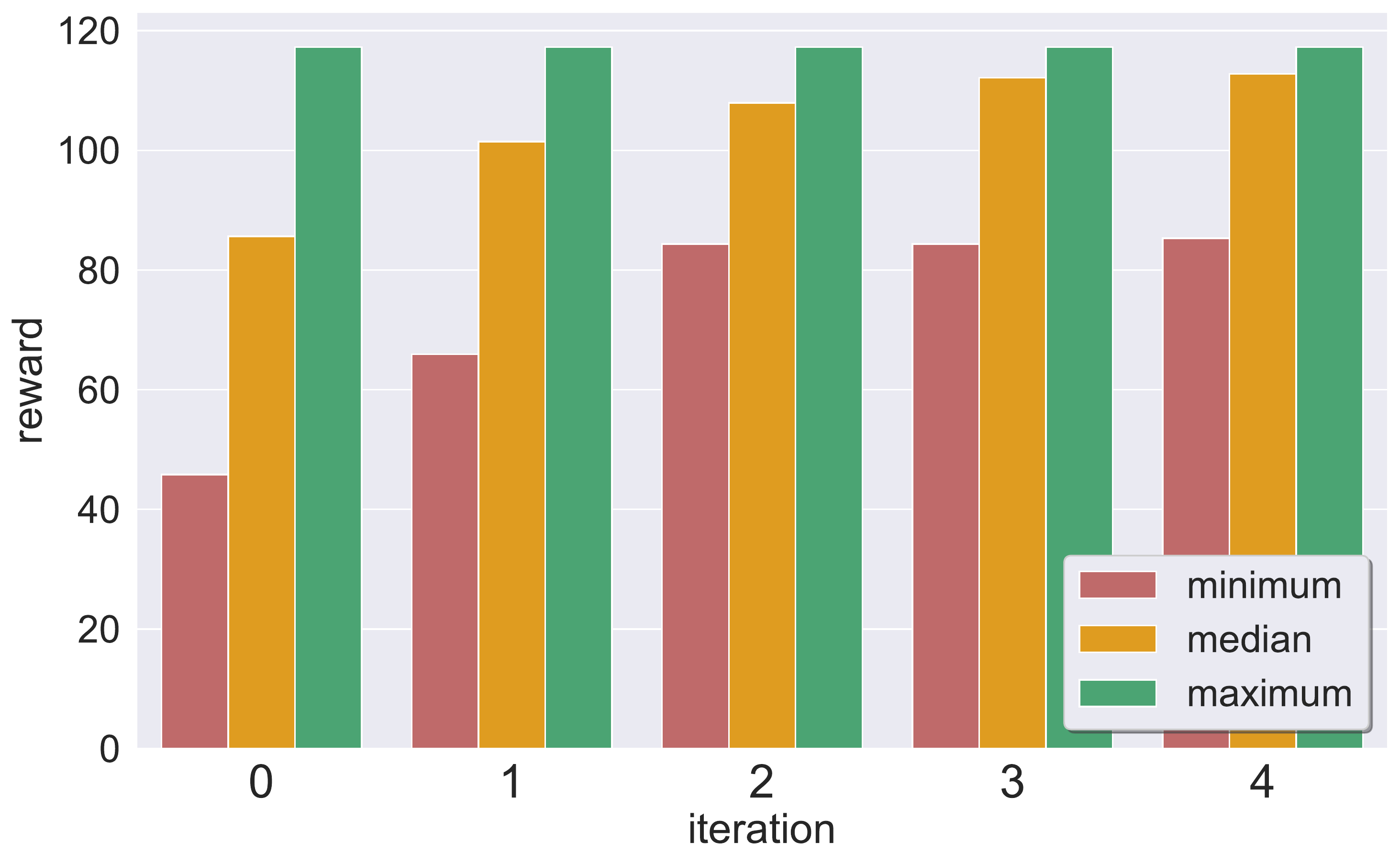}
    \caption{Rewards statistics}
    \label{fig:gradientAttacksMultiStep:aurora:rewardsStats}
     \end{subfigure}
     \hfill
     \begin{subfigure}[t]{0.49\linewidth}
        \includegraphics[width=\textwidth, height=0.67\textwidth]{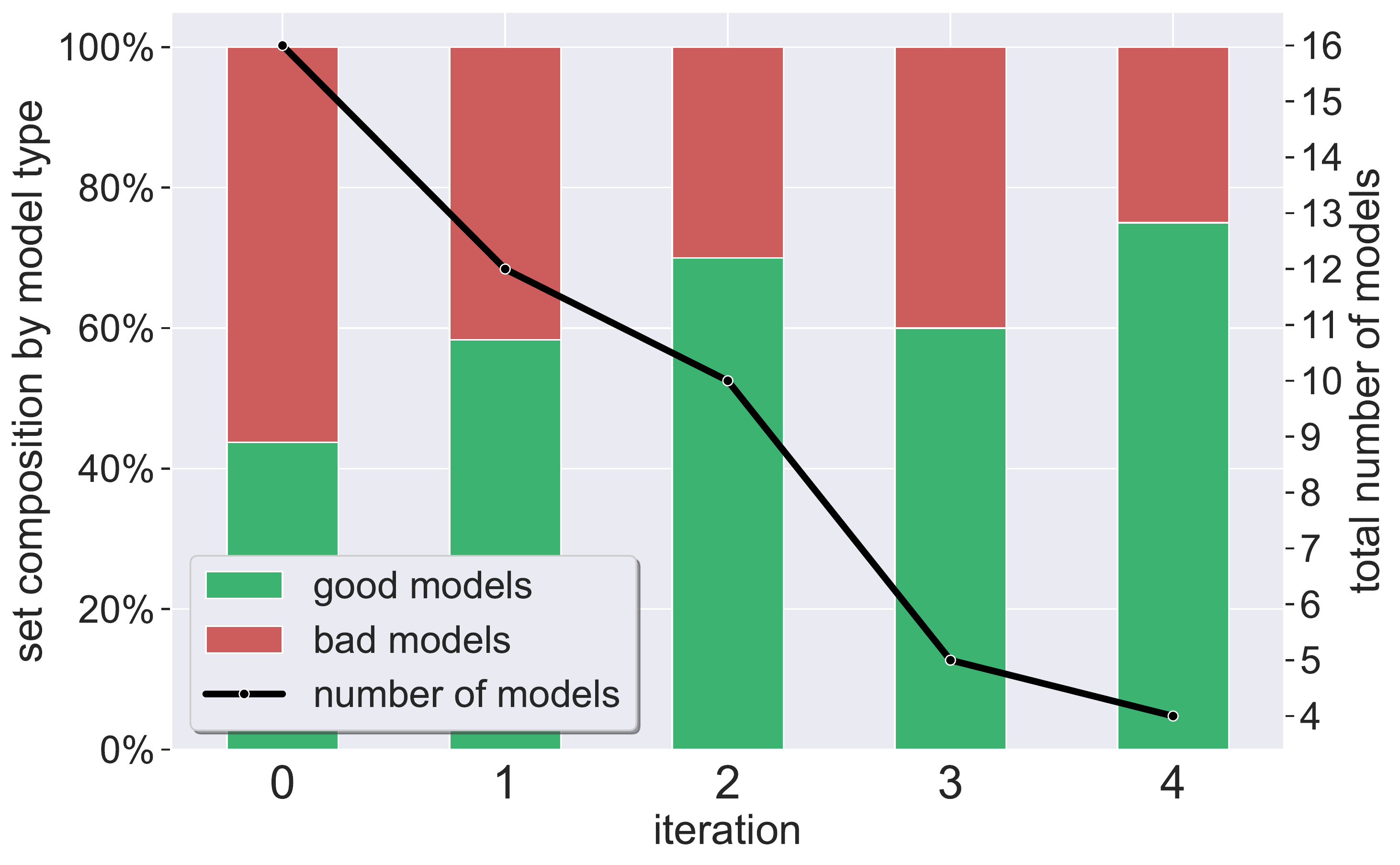}
         \caption{Remaining good and bad models ratio}
        \label{fig:gradientAttacksMultiStep:aurora:goodBadRatio}
     \end{subfigure}
    \caption{Aurora: attack \# 2 (Iterative PGD): Results of models filtered 
    using \emph{PDT} scores approximated by \emph{gradient attacks} (instead of 
    a verification engine) on short-trained Aurora models, using the \maxAgg 
    criterion (and terminating in advance if the disagreement scores are no 
    larger than $2$). In contrast to our verification-driven approach, the 
    final result contains a bad model. Compare to 
    Fig.~\ref{fig:auroraShortMaxFiltering}.}
\label{fig:auroraGradienAttack2}
\end{figure}

As elaborated in subsection~\ref{subsec:gradientAttackComparison}, we ran our 
algorithm on the original models, with the sole difference being the 
replacement of the backend verification engine (Line~\ref{line:SMTsolverForPdt} 
in Alg.~\ref{alg:algorithmPairDisagreementScores}) with the attacks described 
above. The first two attacks (i.e., FGSM and Iterative PGD) were used for both 
Aurora batches (``short'' and ``long'' training), and the third (Constrained 
Iterative PGD) was used in the case of Cartpole and Mountain Car. We note that 
in the case of Aurora, we ran the Iterative PGD attack only when the 
weaker attack failed (hence, only the models from 
Experiment~\ref{exp:auroraShort}). The results (summarized in 
Table~\ref{table:gradientAttackResultsSummary}) validate that in the majority 
of the cases, the attack either failed or mistakenly recognized bad models as 
good 
ones in the predefined input region. For further details regarding the cases in 
which the feasible gradient attacks failed, see 
Fig.~\ref{fig:cartPoleGradienAttack3}, Fig.~\ref{fig:auroraGradienAttack1},  
and Fig.~\ref{fig:auroraGradienAttack2}. These results show the importance of 
using verification for our method.


\begin{table}[ht]
	\centering
        \captionsetup{justification=centering}
	\begin{tabular}{|c|c|c|c|c|c|c|c|c|}
        \hline 
        \texttt{\textbf{ATTACK}} & \texttt{\textbf{BENCHMARK}} & \texttt{\textbf{FEASIBLE}} & \texttt{\textbf{$\#$ PAIRS}} & \texttt{\textbf{$\#$ ALIGNED}} & \texttt{\textbf{$\#$ UNTIGHTENED}} & \texttt{\textbf{$\#$ FAILED}} & 
        \texttt{\textbf{CRITERION}} & 
        \texttt{\textbf{SUCCESSFUL}} \\ \hline
        \multirow{3}{*}{1} & \multirow{3}{*}{Aurora (short)} & 
        \multirow{3}{*}{yes} & \multirow{3}{*}{120} & \multirow{3}{*}{70} & 
        \multirow{3}{*}{50} & \multirow{3}{*}{0} & \conditionMax & 
        \color{red}{no} \\
                \cline{8-9}
                & & & & & & & \conditionCombined & \color{forestGreen}{yes} \\ \cline{8-9}
                & & & & & & & \conditionPercentile & \color{forestGreen}{yes} \\ \hline

        \multirow{3}{*}{1} & 
        \multirow{3}{*}{Aurora (long)} & \multirow{3}{*}{yes} & 
        \multirow{3}{*}{120} & \multirow{3}{*}{111} & \multirow{3}{*}{9} & 
        \multirow{3}{*}{0} & \conditionMax & \color{forestGreen}{yes} \\ 
        \cline{8-9}
            & & & & & & & \conditionCombined & \color{forestGreen}{yes} \\ \cline{8-9}
            & & & & & & & \conditionPercentile & \color{forestGreen}{yes} \\ \hline
        
         \multirow{3}{*}{2} & \multirow{3}{*}{Aurora (short)} & 
         \multirow{3}{*}{yes} & \multirow{3}{*}{120} & \multirow{3}{*}{104} & 
         \multirow{3}{*}{16} & \multirow{3}{*}{0} & \conditionMax & 
         \color{red}{no} \\
                \cline{8-9}
                & & & & & & & \conditionCombined & \color{forestGreen}{yes} \\ \cline{8-9}
                & & & & & & & \conditionPercentile & \color{forestGreen}{yes} \\ \hline

         \multirow{3}{*}{3} & 
         \multirow{3}{*}{Mountain Car} & \multirow{3}{*}{no} & \multirow{3}{*}{120} & \multirow{3}{*}{38} & \multirow{3}{*}{35} & \multirow{3}{*}{47} & 
         \conditionMax & \color{red}{no} 
         \\ \cline{8-9}
                & & & & & & & \conditionCombined & \color{red}{no} \\ \cline{8-9}
                & & & & & & & \conditionPercentile & \color{red}{no} \\ \hline
        
        \multirow{3}{*}{3} & 
        \multirow{3}{*}{Cartpole} & \multirow{3}{*}{partially} & \multirow{3}{*}{120} & \multirow{3}{*}{56} & \multirow{3}{*}{61} & \multirow{3}{*}{3} & \conditionMax & \color{red}{no} \\
        	    \cline{8-9}
        	    & & & & & & & \percentileAgg & \color{red}{no} \\ \cline{8-9}
        	    & & & & & & & \percentileAgg  & \color{red}{no}  \\ \hline
  
	\end{tabular}
	
	\vspace{3mm}
    \caption{Summary of the gradient attack comparison. The first two columns 
    describe the attack chosen and the benchmark on which it was evaluated; the 
    third column states if the incomplete attack allowed a gradient-based 
    approximation of all PDT scores; the next four columns respectively 
    represent the total number of DNN pairs, the number of pairs in which the 
    attack returned a PDT score identical to the original one received by our 
    verification engine, the number of pairs in which the attack returned a 
    score that is less precise than the one returned by our verification 
    engine; and the number of cases in which that attack failed. The 
    second-to-last column shows the filtering criterion, and the last column 
    indicates whether using the attack-based scores resulted in good models 
    only (as was the case when verification was used).}

    \label{table:gradientAttackResultsSummary}
\end{table}

\end{document}